\documentclass[10pt,journal,compsoc]{IEEEtran}
\usepackage[utf8]{inputenc}
\usepackage{tabu}
\usepackage{times}
\usepackage[T1]{fontenc}
\usepackage{graphicx}
\graphicspath{{images/}}
\usepackage{adjustbox}
\usepackage{hyperref}

\usepackage{subcaption}
\usepackage{algorithmic}
\usepackage{stfloats}
\usepackage{float}
\fnbelowfloat

\usepackage{url}
\usepackage{amsmath,amssymb,amsfonts}
\usepackage{textcomp}
\usepackage{multirow}
\usepackage{booktabs}

\ifCLASSOPTIONcompsoc
  \usepackage[nocompress]{cite}
\else
  \usepackage{cite}
\fi
\ifCLASSINFOpdf
\else
\fi
\begin{document}
%
\title{S\textsuperscript{2}-FPN: Scale-ware Strip Attention Guided Feature Pyramid Network for Real-time Semantic Segmentation}

\author{Mohammed A. M. Elhassan\textsuperscript{1}, Chenhui Yang\textsuperscript{1,*}, Chenxi Huang\textsuperscript{1}
        , Tewodros Legesse Munea\textsuperscript{1},
        Xin Hong\textsuperscript{2}, Abuzar B. M. Adam\textsuperscript{3}, Amina Benabid\textsuperscript{4}\\
        \hspace{18cm}
        \textsuperscript{1} School of Informatics, Xiamen University,Xiamen, Fujian, China, 361005\\
       \textsuperscript{2} College of Computer Science and Technology, Huaqiao University, Xiamen, Fujian, China, 361021\\ 
       \textsuperscript{3} Centre for Security, Reliability and Trust (SnT), University of Luxembourg, 1855 Luxembourg City, Luxembourg\\
       \textsuperscript{4}	School of Computer Science and Technology, Zhejiang Normal University, Jinhua 321004, China
\IEEEcompsocitemizethanks{\IEEEcompsocthanksitem E-mail addresses: mohammedac29@stu.xmu.edu.cn (Mohammed A. M. Elhassan). 
\IEEEcompsocthanksitem E-mail addresses: chyang@xmu.edu.cn ( Chenhui Yang). \protect\\
\IEEEcompsocthanksitem E-mail addresses: supermonkeyxi@xmu.edu.cn ( Chenxi Huang)
\IEEEcompsocthanksitem E-mail addresses: teddylegessemunea@gmail.com ( Tewodros Legesse Munea)
\IEEEcompsocthanksitem E-mail addresses: xinhong@hqu.edu.cn ( Xin Hong)
\IEEEcompsocthanksitem E-mail addresses: abuzar.babikir@uni.lu(Abuzar B. M. Adam)
\IEEEcompsocthanksitem E-mail addresses: amina.benabid@zjnu.edu.cn(Amina Benabid)
\IEEEcompsocthanksitem \textsuperscript{*} Correspondence should be addressed to Chenhui Yang 
}
\thanks{Manuscript received April 19, 2005; revised August 26, 2015.}
}

\IEEEtitleabstractindextext{%
\begin{abstract}
Modern high-performance semantic segmentation methods employ a heavy backbone and dilated convolution to extract the relevant feature.  Although extracting features with both contextual and semantic information is critical for the segmentation tasks, it brings a memory footprint and high computation cost for real-time applications. This paper presents a new model to achieve a trade-off between accuracy/speed for real-time road scene semantic segmentation. Specifically, we proposed a lightweight model named Scale-aware Strip Attention Guided Feature Pyramid Network (S\textsuperscript{2}-FPN). Our network consists of three main modules: Attention Pyramid Fusion (APF) module, Scale-aware Strip Attention Module (SSAM), and Global Feature Upsample (GFU) module. APF adopts an attention mechanisms to learn discriminative multi-scale features and help close the semantic gap between different levels. APF uses the scale-aware attention to encode global context with vertical stripping operation and models the long-range dependencies, which helps relate pixels with similar semantic label. In addition, APF employs channel-wise reweighting block (CRB) to emphasize the channel features. Finally, the decoder of S\textsuperscript{2}-FPN then adopts GFU, which is used to fuse features from APF and the encoder. Extensive experiments have been conducted on two challenging semantic segmentation benchmarks, which demonstrate that our approach achieves better accuracy/speed trade-off with different model settings. The proposed models have achieved a results of 76.2\%mIoU/87.3FPS, 77.4\%mIoU/67FPS, and 77.8\%mIoU/30.5FPS on Cityscapes dataset, and 69.6\%mIoU,71.0\% mIoU,and 74.2\% mIoU on Camvid dataset.The code for this work will be made available at \url{https://github.com/mohamedac29/S2-FPN}
\end{abstract}

\begin{IEEEkeywords}
Semantic segmentation, Scale-aware strip attention,deep convolutional neural networks, real-time semantic segmentation.
\end{IEEEkeywords}}

\maketitle

\IEEEdisplaynontitleabstractindextext
\IEEEpeerreviewmaketitle

\IEEEraisesectionheading{\section{Introduction}\label{sec:introduction}}

Semantic segmentation is an essential high-level topic in computer vision and has been widely used in various challenging problems such as, medical diagnosis, \cite{ronneberger2015u,saha2018her2net}, autonomous vehicles  \cite{siam2018comparative}, and scene analysis \cite{zhao2018psanet,zhao2017pyramid}. Unlike image classification, which aims to classify the whole image, semantic segmentation predicts the per-pixel class for each image content. The current frontiers in semantic segmentation methods are driven by the success of deep convolution neural networks, specifically the fully convolution network (FCN) Framework \cite{long2015fully}. In the original FCN approach to obtain larger receptive fields, we increase the depth of the network with more convolutional and downsampling layers. However, simply increasing the number of downsampling operations leads to reduced feature resolution, which causes spatial information loss. On the other hand, the increased number of convolutional layers adds more challenges to network optimization. 

\begin{figure}
    \centering
    \includegraphics[width=0.9\linewidth]{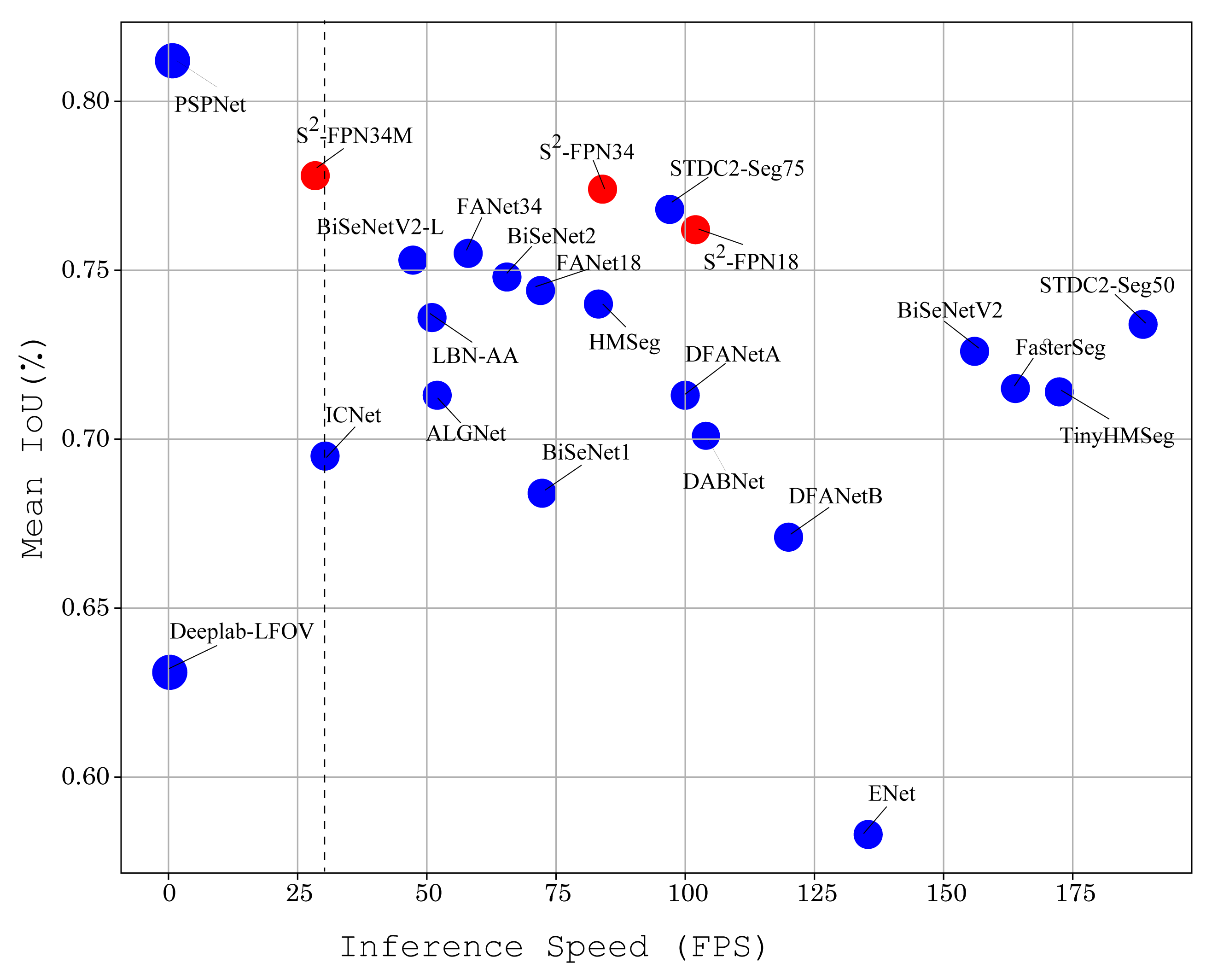}
    \caption{Accuracy/Speed performance comparison on the Cityscapes test set. Our methods are presented in red dots while other methods are presented in blue dots. Our approaches achieve state-of-the-art speed-accuracy trade-off}
    \label{fig:cityscapes_trade_off}
\end{figure}

\begin{figure*}
	\centering
	\begin{subfigure}[b]{0.23\textwidth}
		\includegraphics[width=\textwidth]{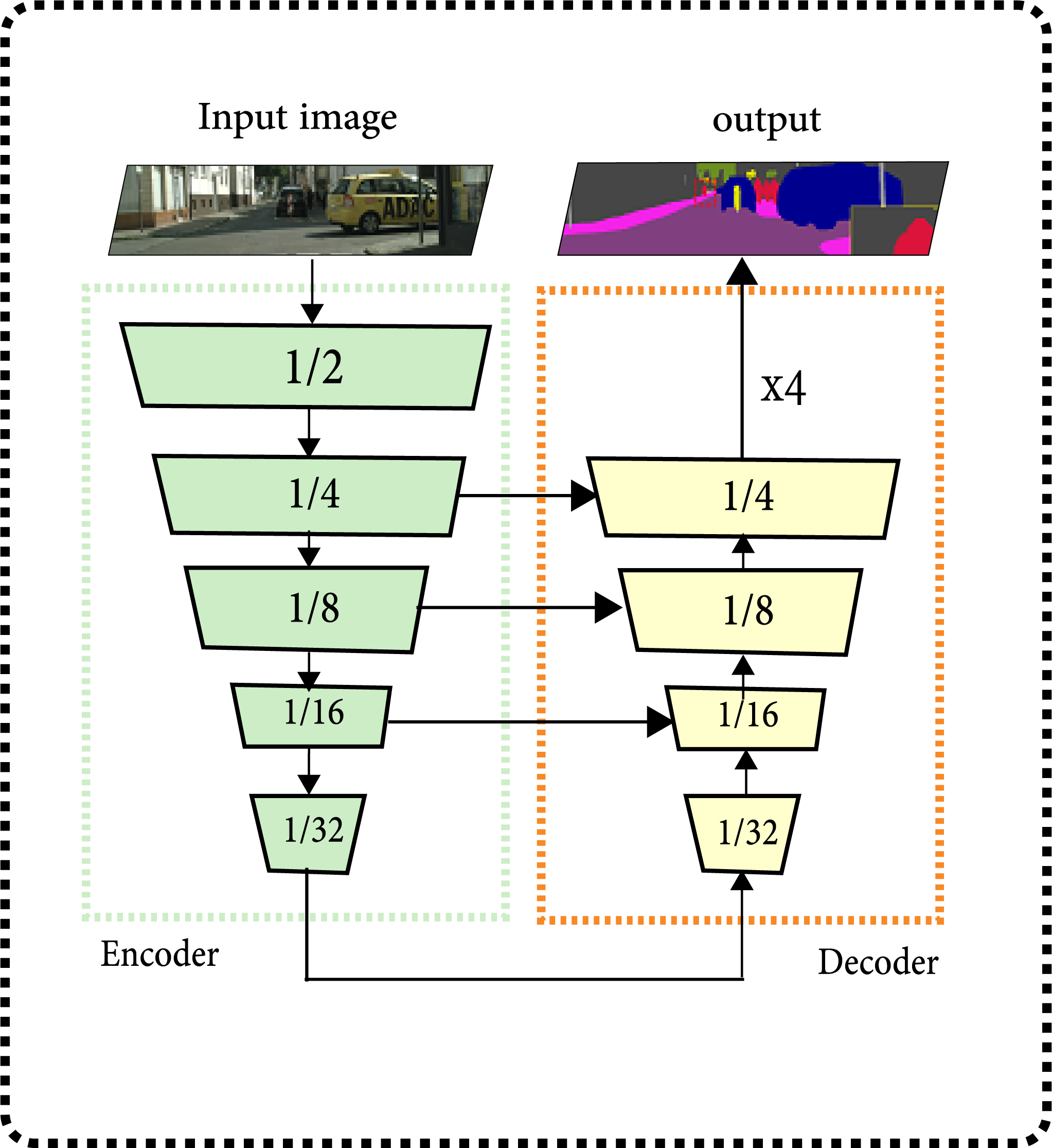}
		\caption{Encoder-decoder}
		\label{fig1:encoder}
	\end{subfigure} \hspace{1.5mm}
	\begin{subfigure}[b]{0.23\textwidth}
		\includegraphics[width=\textwidth]{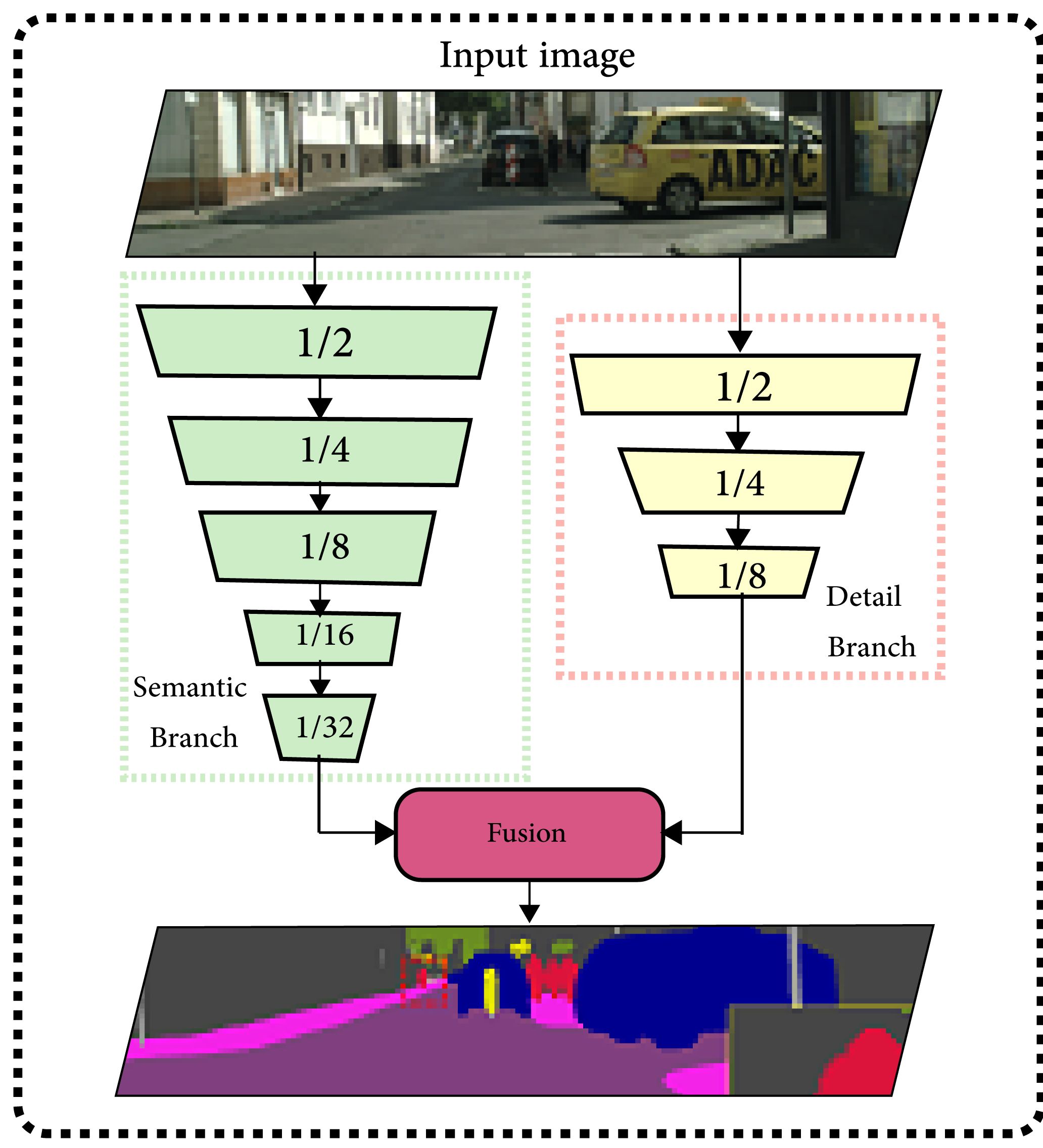}
		\caption{Two Pathway}
		\label{fig2:pathway}
	\end{subfigure}\hspace{1.5mm}
	\begin{subfigure}[b]{0.23\textwidth}
		\includegraphics[width=\textwidth]{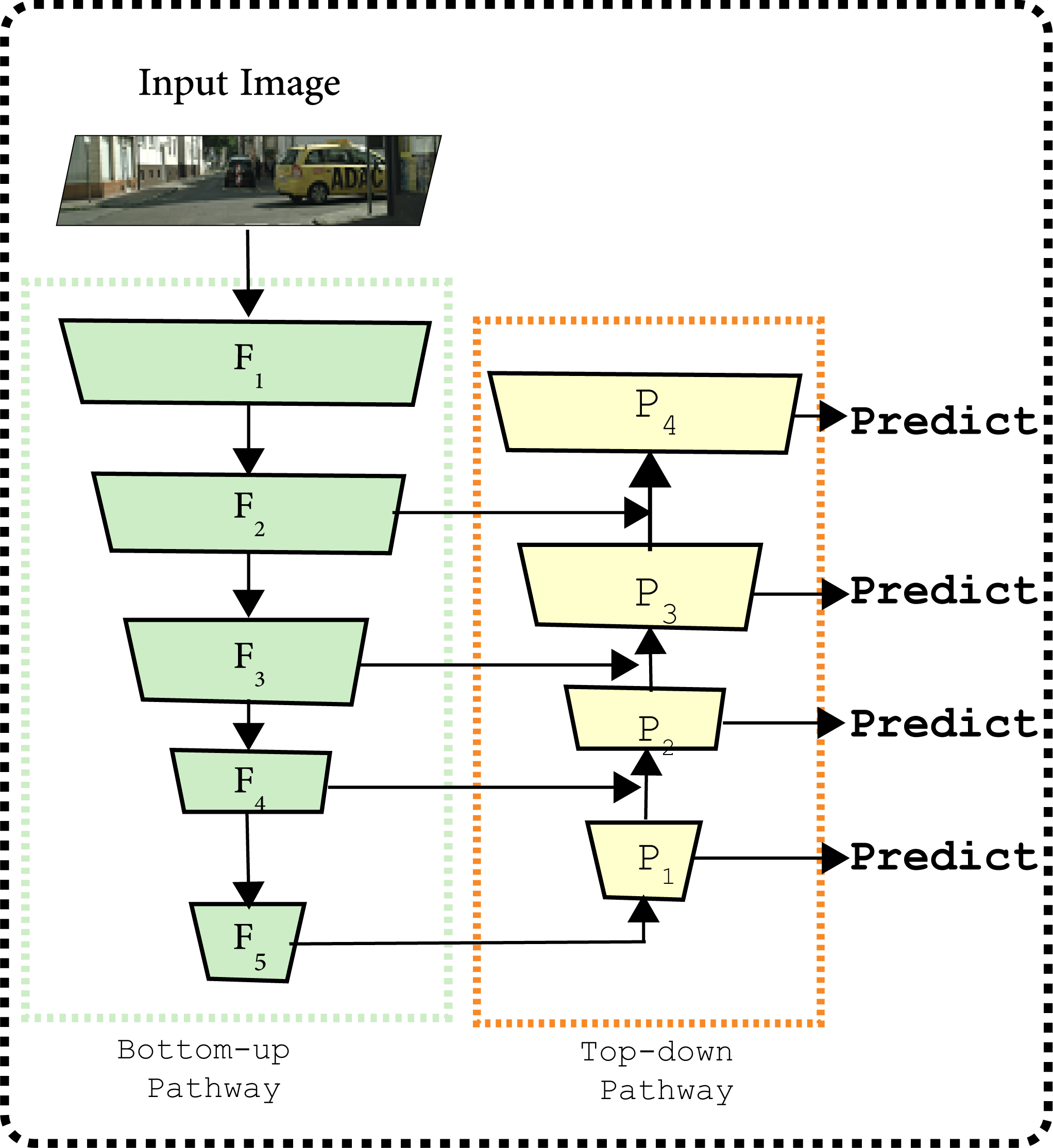}
		\caption{FPN}
		\label{fig3:fpn}
	\end{subfigure}\hspace{1.5mm}
	\begin{subfigure}[b]{0.23\textwidth}
		\includegraphics[width=\textwidth]{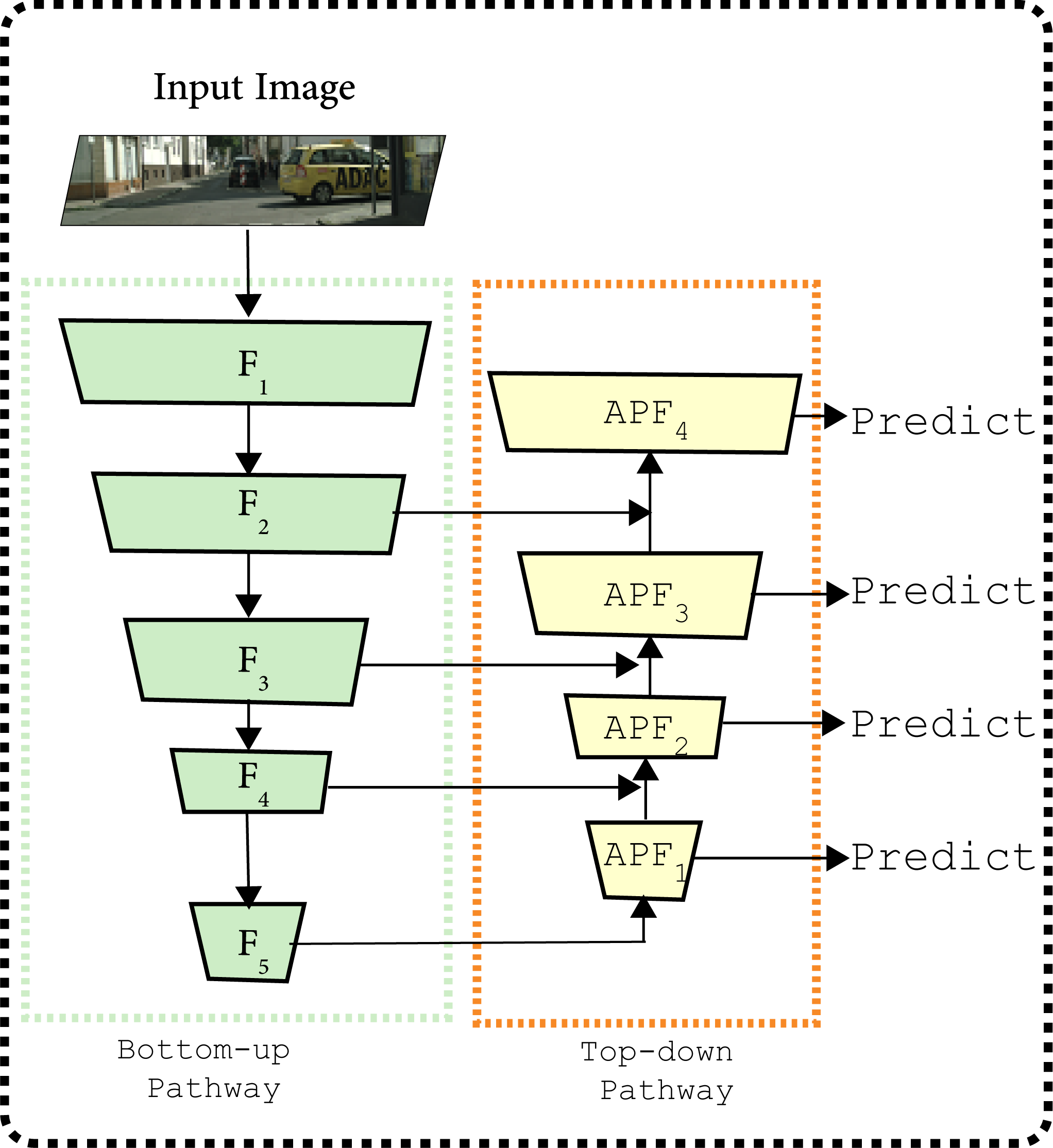}
		\caption{S\textsuperscript{2}\-FPN}
		\label{fig4:sfpn}
	\end{subfigure}\hspace{1.5mm}
	\caption{A comparison of an important semantic segmentation architectures,(a) encoder-decoder model,(b) two-pathway model,(c) feature pyramid model,(d) Scale-aware feature pyramid model (Ours).}
	\label{networks_comparison}
\end{figure*}

Over the years, various approaches have been proposed to tackle the drawbacks of the typical FCN-based architecture. Here we mention the following: (i) increasing the resolution of feature maps or maintaining a high-resolution feature across stages, e.g., through decoder network (Figure \ref{networks_comparison} (\subref{fig1:encoder})) \cite{ronneberger2015u,badrinarayanan2017segnet,zhou2018unet++}, dilated convolutions  \cite{yu2015multi,chen2017deeplab} or high resolution \cite{sun2019deep}. (ii) improve the segmentation performance with the subsequent approaches, e.g., PSPNet \cite{zhao2017pyramid}, DeepLab \cite{chen2017deeplab,chen2018encoder}, PAN \cite{li2018pyramid} DenseASPP \cite{yang2018denseaspp} , RefineNet \cite{lin2017refinenet}, PPANet \cite{elhassan2021ppanet}, and ParseNet \cite{liu2015parsenet}. These previous networks enlarge the receptive field to exploit the rich contextual information. Nevertheless, the latter solution is infeasible for real-time applications because it requires heavy networks and expensive computation. (iii) some networks such as  \cite{elhassan2021dsanet,yu2018bisenet} have added spatial encoding path to preserve the spatial details as shown in Figure \ref{networks_comparison}.\subref{fig2:pathway}.
To perform fast segmentation with light computation cost, small size and satisfactory segmentation accuracy, there are certain design philosophy that could be followed: (i) incorporate a pre-trained lightweight classification networks\cite{iandola2016squeezenet,howard2017mobilenets,ma2018shufflenet,chollet2017xception} as a backbone to construct segmentation architecture such as BiSeNet \cite{yu2018bisenet}, STDC-Seg \cite{fan2021rethinking}. (ii) design a building block that is suitable for low computation cost \cite{li2019dabnet,romera2017erfnet}. (iii) put into consideration the downsampling strategy and the depth of the network \cite{li2019dfanet}. (iv) combine the low-level and high-level features using the multi-path framework \cite{yu2018bisenet,yu2021bisenet,elhassan2021dsanet}. in addition to that, restricting the input image size plays an important part in increasing the inference speed.\\

Furthermore, many existing frameworks \cite{ghiasi2019fpn,liu2018path,tan2020efficientdet,kirillov2019panoptic,lin2017feature} have been developed based on FPN \cite{lin2017feature} to exploit the inherent multi-scale feature representation of deep convolutional networks. More specifically, FPN-based architectures combine large-receptive field features, low-resolution- with small-receptive-field features, high-resolution to capture objects at different scales. For instance, \cite{kirillov2019panoptic,lin2017feature} utilize the lateral path to fuse adjacent features in a top-down manner. This way FPN strengthen the learning of multi-scale representation.Until now, feature pyramid network based approaches have achieved state-of-arts performance in object detection. Nevertheless, the fusion of adjacent features from different stages ignores the semantic gap between these features, which degrades the multi-scale feature representation (FPN-like model is shown in Figure \ref{networks_comparison} (\subref{fig3:fpn}).
To alleviate this problem and achieve a better accuracy/speed trade-off, we propose Scale-aware Strip Attention guided Pyramid Network(S\textsuperscript{2}-FPN), which can extract multi-scale/multilevel representations while maintaining high computation efficiency. Fig. \ref{fig:cityscapes_trade_off} illustrates the accuracy (mIoU) and inference speed (FPS) obtained by several state-of-the-art methods, including PSPNet\cite{zhao2017pyramid}, DeepLab\cite{chen2017deeplab}, ENet\cite{zheng2015conditional},ICNet\cite{chen2014semantic}, DABNet\cite{long2015fully}, DFANet-A\cite{yu2016multi}, DFANet-B\cite{yu2016multi}, BiSeNet1\cite{yu2018bisenet}, BiSeNet2\cite{yu2018bisenet}, FasterSeg\cite{chen2019fasterseg}, TD4-Bise18\cite{hu2020temporally}, FANet-18\cite{hu2020real}, FANet-34\cite{hu2020real}, LBN-AA\cite{dong2020real}, AGLNet\cite{dong2020real}, BiSeNetV2\cite{yu2021bisenet}, BiSeNetV2-L\cite{yu2021bisenet}, HMSeg\cite{li2020humans}, TinyHMSeg\cite{li2020humans}, STDC2-Seg50\cite{fan2021rethinking}, STDC2-Seg50\cite{fan2021rethinking}, and our proposed method, on the Cityscapes test dataset.

The main contributions are summed as follows.
\begin{enumerate}
    \item We propose a Scale-ware strip attention module with low computation overhead.
    \item A new attention pyramid fusion(APF) module is proposed to address the semantic gap in the main FPN. Our APF aggregates multi-scale representation efficiently.
    \item 	Based on the two modules, we propose S\textsuperscript{2}-FPN
    architecture, which effectively improves the accuracy with high inference speed in complex environments.
    \item We report the results on a single-scale image on two datasets. S\textsuperscript{2}-FPN achieves 77.8\%, 77.4\% and 76.2\% mIoU on the Cityscapes dataset. 74.2\%, 71.0, and 69.6 with backbones ResNet34-modified, ResNet34 and ResNet18, respectively.
    \item For image resolution of 512x1024, S2-FPN(ResNet18) obtains 87.3 FPS on NVIDIA GTX1080Ti using PyTorch on the Cityscapes dataset, whereas S\textsuperscript{2}-FPN (ResNet34) and S\textsuperscript{2}-FPN(ResNet34-modified) achieve 67 FPS and 30.5 FPS, respectively. 
\end{enumerate}

\section{Related Work}
\subsection{Efficient Network Designs}
In recent literature, real-time model design has become an essential part of computer vision research development. Several lightweight architectures were introduced as a trade-off between accuracy and latency SqueezeNet \cite{iandola2016squeezenet}, MobileNet \cite{howard2017mobilenets,sandler2018mobilenetv2}, ShuffleNet \cite{zhang2018shufflenet,ma2018shufflenet},and Xception \cite{chollet2017xception}. SqueezeNet reduces the number of parameters through the squeeze and expansion module. MobileNetV1 implemented depth-wise separable convolutions to reduce the number of parameters. MobileNetV2 proposed an inverted residual block to alleviate the effect of depth-wise separable convolutions. ShuffleNet and Xception utilize group convolution to reduce FLOPs. ResNet \cite{he2016deep} adopted residual blocks to obtain outstanding performance. These models were designed primarily for image classification tasks. Later on, incorporated to work as backbones for semantic segmentation networks to achieve real-time semantic segmentation.
\subsection{Real-time Segmentation}
Real-time semantic segmentation methods aim to predict dense pixels accurately while maintaining high inference speed. Over the years there are different approaches have been proposed in this regard. ENet \cite{paszke2016enet}, ERFNet \cite{romera2017erfnet}, and ESPNet \cite{mehta2018espnet,mehta2019espnetv2} proposed compact semantic segmentation networks that utilize lightweight backbones. LEDNet \cite{wang2019lednet}, DABNet \cite{li2019dabnet} introduced a lightweight model to address the computation burden that limits the usage of deep convolutional neural networks in mobile devices. The former used asymmetric encoder-decoder-based architecture and implemented channel split and shuffle operations to reduce the computation cost. The latter structured a depth-wise symmetric building block to jointly extract local and global context information.  FDDWNet \cite{liu2020fddwnet} utilized factorized depth-wise separable convolutions to reduce the number of parameters and the dilation to learn multi-scale feature representations.
ICNet \cite{zhao2018icnet} proposes a compressed version of  PSPNet based on an image cascade network to increase the semantic segmentation speed. DFANet \cite{li2019dfanet} obtains the multi-scale feature propagation by utilizing a sub-stage and sub-network aggregation.Tow-pathway networks such as  BiSeNetV2 \cite{yu2021bisenet} and DSANet \cite{elhassan2021dsanet} Fast-SCNN \cite{poudel2019fast} introduce a shallow detailed branch to extract the spatial details and a deep semantic branch to extract the semantic context. This paper propose a lightweight model that utilizes (ResNet18 or ResNet34) as backbones. 

\subsection{Self-Attention Mechanism}
Recent studies have shown that attention mechanisms can be successfully applied to different tasks such as machine translation \cite{gehring2016convolutional,gehring2017convolutional}, image classification \cite{wang2017residual}, object detection \cite{carion2020end,zhu2020deformable} and semantic segmentation \cite{ye2019cross}. The attention mechanisms perform better than convolutional neural networks when used to model the long-range dependencies. SENet \cite{hu2018squeeze} proposed the squeeze and excitation module to re-calibrate the channel dependency dynamically. Motivated by self-attention, DANet \cite{fu2019dual} improves feature representation by applying channel attention and position attention, while OCNet \cite{yuan2018ocnet} uses self-attention to learn the object context. In this paper, to construct an efficient and effective approach for modelling the long-range dependencies, we introduce scale-aware attention with vertical striping operation to reduce the computation cost.
\subsection{Multi-scale Feature Fusion}
One of the main difficulties in semantic segmentation is effectively processing and representing multi-scale features. Earlier semantic segmentation methods perform multilevel feature representations fusion \cite{long2015fully,badrinarayanan2017segnet,ronneberger2015u}. Multi-branch architectures have been adopted to tackle the issue of multi-scale fusion, e.g., BiSiNetV1 \cite{yu2018bisenet}, BiSiNetV2 \cite{yu2021bisenet} and DSANet \cite{elhassan2021dsanet}, add a shallow branch to preserve the spatial details. ICNet fused branches with different input resolutions. SPFNet \cite{elhassan2022spfnet} proposed a method for learning separate spatial pyramid fusion for each feature subspace. Feature pyramid network \cite{lin2017feature} is widely adopted in the object detection task \cite{guo2020augfpn,ghiasi2019fpn} to address the problem of multi-scale feature representation. The feature pyramid network architecture introduces a top-down routing to fuse features. Based on FPN, NAS-FPN \cite{ghiasi2019fpn} utilizes the neural architecture search to obtain optimal topology. EfficientDet \cite{tan2020efficientdet} acquires more higher-level feature fusion using a repeated bidirectional path. PANet \cite{liu2018path} proposes a method that facilitates the information flow through a bottom-up path augmentation. This work proposes the Attention Pyramid Fusion module, which adopts a lightweight scale-aware attention strategy to bridge the semantic gap between the different levels more efficiently.

\begin{figure*}
	\begin{center}
		\includegraphics[width=\linewidth]{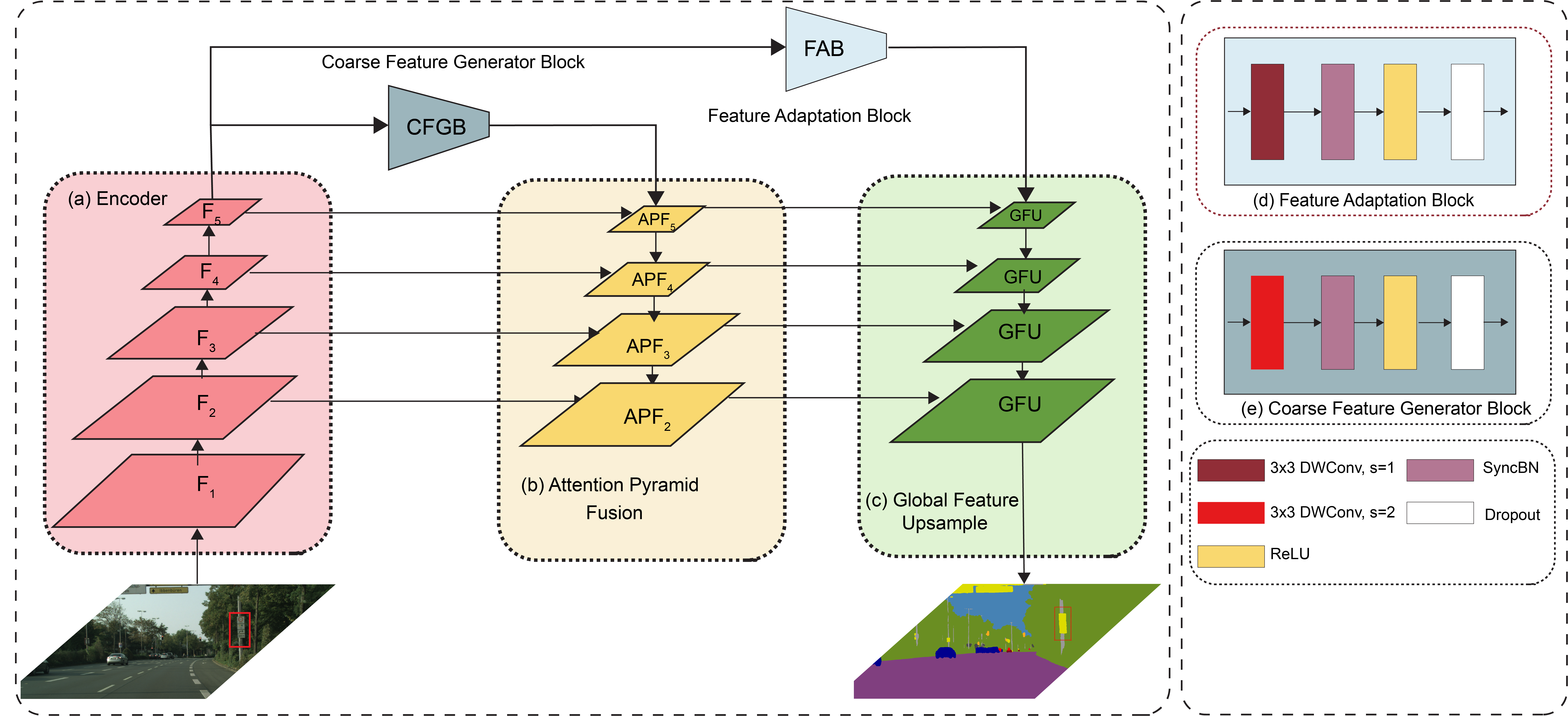}
		\caption{The detailed architecture of the proposed Scale-aware Strip Attention Guided Pyramid Fusion model (S\textsuperscript{2}-FPN). The model constructs from the following modules: (a) Encoder, which incorporates ResNet18 or ResNet34, (b) Attention Pyramid Fusion module, (c) illustrates the Global Feature Upsample(GFU) module, (e) Feature adaptation block (FAB), and (e) Components of Coarse Feature Generator block. }
		\label{main_architecture}
	\end{center}
\end{figure*}

\section{Methodology}
\label{sec3:method}
This section describes the proposed Scale-aware Strip Attention Guided Feature Pyramid Network (S\textsuperscript{2}-FPN). We first give an overview of S\textsuperscript{2}-FPN in Section \ref{method:overview}. Then, we introduce the different components of S\textsuperscript{2}-FPN from Section \ref{mehtod:scale_strip_module} to Section \ref{mehtod:global_feature_up_module}. Finally, we present the experimental results and analysis in Section \ref{experimental_results}. 

\subsection{Overview}
\label{method:overview}
Figure \ref{main_architecture}. illustrates the proposed S\textsuperscript{2}-FPN, which follows the feature pyramid network and encoder-decoder architectures. S\textsuperscript{2}-FPN consists of three main parts: Feature Extractor or Encoder, Attention Pyramid Fusion (APF), and Global Feature Upsample (GFU). In particular, Coarse Feature Generator Block (CFGB), Feature Adaptation Block (FAB), and Scale-aware Strip Attention Module (SSAM). 
S\textsuperscript{2}-FPN takes ResNet18 and ResNet34 \cite{he2016deep} as base models for feature extraction while removing the global average pooling and the softmax. We utilize $\{F_{1}, F_{2}, F_{3}, F_{4}, F_{5}\}$ as a feature hierarchy from the base model with strides $\{2,4,8,16,32\}$. Then, we attached a coarse feature generator block and feature adaptation block in parallel after $F_{5}$. CFGB is constructed from a depth-wise convolution layer with stride 2 to generate coarse features for the attention pyramid fusion, while FAB is produced through a depth-wise convolutional layer with stride 1 to project the feature maps of the encoder into the global feature upsample module. Note that the attention pyramid features have strides of  $\{4,8,16,23\}$ pixel w.r.t the input image. $\{APF_{2}, APF_{3}, APF_{4}, APF_{5}\}$ are the top-down features generated using the Attention Feature Pyramid (APF) module (Section \ref{mehtod:attention_pyramid_module}). 

\subsection{Scale-aware Strip Attention Module}
\label{mehtod:scale_strip_module}
\begin{figure*}
	\begin{center}
		\includegraphics[width=0.8\linewidth]{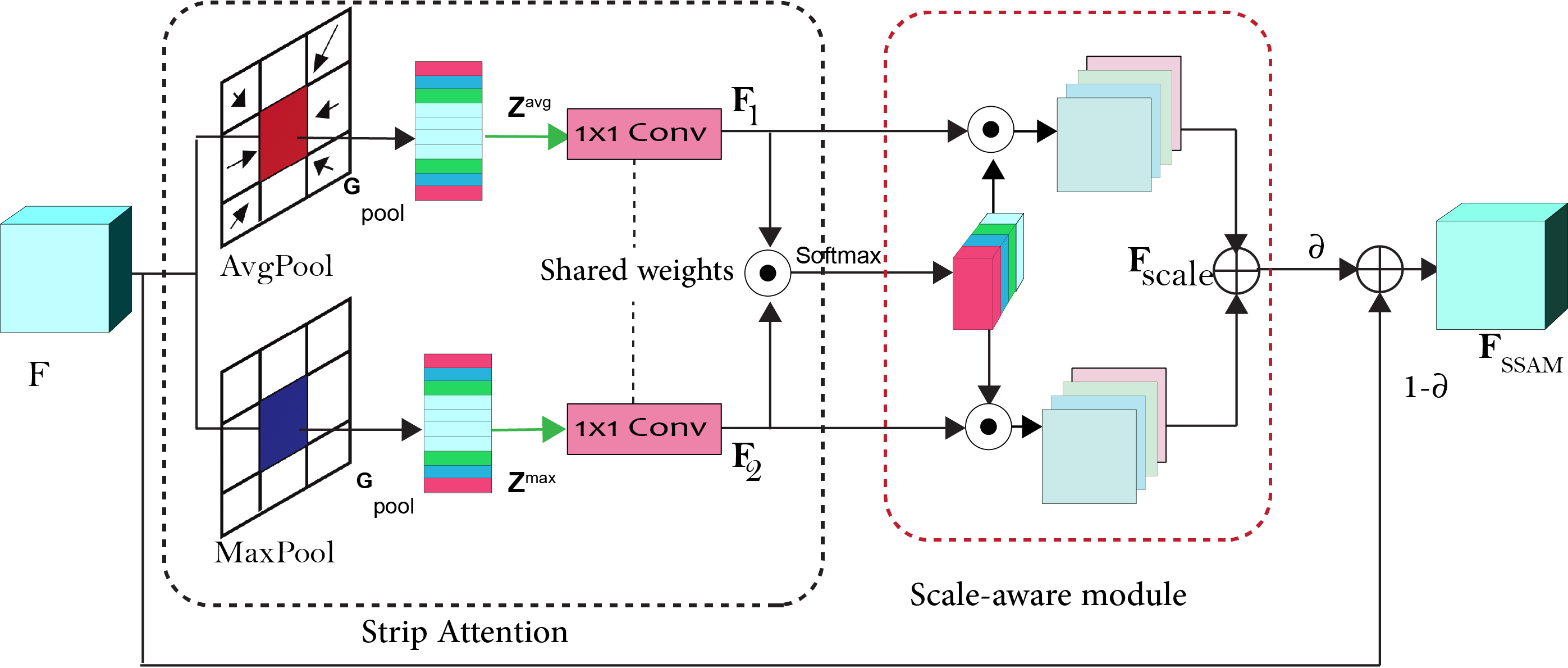}
		\caption{The illustration of Scale-Aware Strip Attention  module SSAM.}
		\label{scale_aware_strip_attention_}
	\end{center}
\end{figure*}

In order to capture the long-range dependencies and reduce the computation cost, we introduce the Scale-aware Strip Attention module. In particular, inspired by the results in \cite{hou2020strip,fu2019dual}; and to maintain the contextual consistency, we apply striping operation to collect long-range context along the vertical axis. Figure \ref{scale_aware_strip_attention_} illustrates the details structure of Scale-aware Strip Attention module.\\
Let $\textbf{F}_{Refine}\in \mathbb{R}^{C\times H\times W}$ be an input feature maps ($\textbf{F}_{Refine}$ is the output of feature refinement block (FRB) in APF Figure \ref{fig:attention_pyramid_fusion}) where C is the number of channels, W and H are the spatial dimensions. SSAM first extracts height-wise contextual information from each row of the refined feature maps $\textbf{F}_{Refine}$ by aggregating the $C\times H\times W$ input representation into a $\textbf{Z}^{max} \in \mathbb{R}^{C\times H\times 1}$, $\textbf{Z}^{avg} \in \mathbb{R}^{C\times H\times 1}$ using Max Poling or Average Pooling operation in parallel. $\textbf{Z}^{avg}$ and  $\textbf{Z}^{max}$ are illustrated in Equation \ref{vec1} and \ref{vec2}, respectively.

\begin{equation}\label{vec1}
    \textbf{Z}^{avg} = \textbf{G}_{pool}(\textbf{F}_{R})
\end{equation}

\begin{equation}\label{vec2}
    \textbf{Z}^{max} = \textbf{G}_{pool}(\textbf{F}_{R})
\end{equation}

Where $\textbf{G}_{pool}(.)$ represents max or average pooling operations.\\
Then SSAM applies $1\times1$ convolution layer with shared weights to $\textbf{Z}^{max}$ and $\textbf{Z}^{avg}$ (Equations \ref{f1} and \ref{f1}) to transfer the height-wise context information of every feature within $\textbf{F}_{R}$, and employs a softmax function to generate an attention map that highlights the importance of corresponding feature maps in $\textbf{F}_{R}$ (Equation \ref{att}). 

\begin{equation}\label{f1}
	\textbf{F}_{1} = f^{1s}(\textbf{Z}^{avg})
\end{equation}

\begin{equation}\label{f2}
	\textbf{F}_{2} = f^{1s}(\textbf{Z}^{max})
\end{equation}

\begin{equation}\label{att}
	\textbf{A} = \sigma(\textbf{F}_{1}\odot \textbf{F}_{2})
\end{equation}

Where $f^{1s}(.)$ represents a convolutional layer with shared weights.
The attention mechanism can select the appropriate scale feature dynamically and fuse feature of different scale by self-learning. Next, we have performed element-wise multiplication operations ($\odot$) between attention maps and $\textbf{F}_{1}$,$\textbf{F}_{2}$ to generate the scaled feature map $\textbf{F}_{scale}$. Finally, an element-wise sum operation between $\textbf{F}_{scale}$ and the input feature map \textbf{F} is employed with learnable parameters $\alpha$ to obtain the final output $\textbf{F}_{SSAM}$ as in Equation \ref{fout}

\begin{equation}\label{fscale}
	\textbf{F}_{scale} = \textbf{A}\odot \textbf{F}_{2} + \textbf{A}\odot \textbf{F}_{2}
\end{equation}

\begin{equation}\label{fout}
	\textbf{F}_{SSAM} = \alpha \textbf{F}_{scale} + (\alpha-1)\textbf{F}
\end{equation}

\subsection{Attention Pyramid Fusion (APF) Module}
\label{mehtod:attention_pyramid_module}
\begin{figure*}
	\begin{center}
		\includegraphics[width=0.9\linewidth]{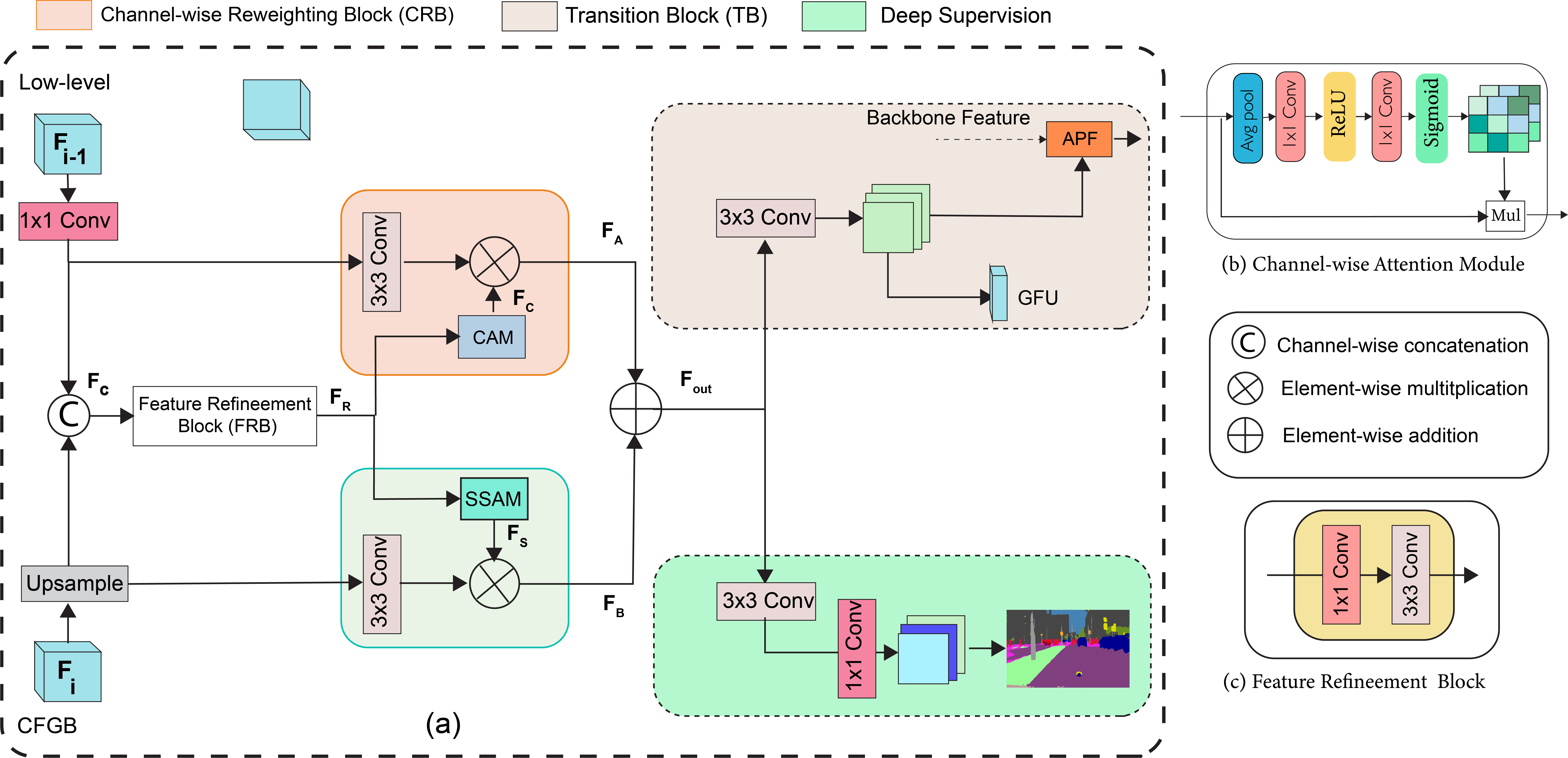}
		\caption{An overview of the Attention Pyramid Fusion Module. (a) APF module architecture. (b) Components of the channel attention module (CAM). (c) Components of the feature refinement block (FRB).}
		\label{fig:attention_pyramid_fusion}
	\end{center}
\end{figure*}

Maintaining detailed information of an object when passing an image through the encoder and decoder is a challenging problem. Feature pyramid network \cite{lin2017feature}, which was first introduced for object detection, has been used to leverage the multi-level feature from the backbone (see Figure \ref{networks_comparison} \subref{fig3:fpn}). However, inherent defects in FPN inhibit it from extracting sufficient discriminative features. For instance, the extracted feature by the lateral connection in FPN could suffer information loss in the high-level layers. Another issue is that using bilinear interpolation operation and simple element-wise addition of adjacent features ignores the semantic gaps of different depths.  Based on these observations, we propose the Attention Pyramid Fusion (APF) module, which can learn discriminative multi-scale features. The detailed structure of our Attention Pyramid Fusion block is shown in Figure \ref{fig:attention_pyramid_fusion}. In APF, the scale-aware strip attention guides the deeper stages to aggregate high-resolution features, while channel-wise attention employs channel reweighting. By combining these attention mechanisms, APF is able to extract more detailed information in the final segmentation map.\\
In order to combine the coarse $\textbf{F}_{i}$ and low-level $\textbf{F}_{i-1}$ adjacent feature maps, we first apply a 1x1 convolution on $\textbf{F}_{i-1}$ followed by batch normalization and ReLU operation. Meanwhile, we upsample the coarse feature map generated with CFGB to match the low-level feature in terms of channel and spatial dimensions. Then the coarse and the low-level feature maps are concatenated channel-wise to produce $\textbf{F}_{concat}$, and the output of this operation is fed into a feature refinement block (FRB) to reduce the aliasing effect. FRB is constructed by stacking two convolutional layers, $1\times1$ and $3\times3$ each, followed by batch normalization and ReLU.\\ 
Furthermore, we have developed two modules to reduce the gap between different levels and enhance semantic consistency. Specifically, We first apply a $3\times3$ convolution followed by batch normalization and ReLU on the low-level path, then multiplied by channel attention (Fig \ref{fig:attention_pyramid_fusion}.b ) to generate a feature reweighting (CRB) that filters the irrelevant information (see Fig \ref{fig:attention_pyramid_fusion}).
CAM employs global average pooling to extract global context and learn the weights for the channel sub-module. 
The final output of feature reweighting branch $\textbf{X}_{A}$ is given in Equation \ref{x1}.

\begin{equation}\label{x1}
    \textbf{X}_{A} = f^{3}(f^{1}(\textbf{F}_{i-1})) \odot \textbf{F}_{C}(\textbf{F}_{R})
\end{equation}

where $f^{3}(\cdot)$ represents the convolution layer whose kernel size is $3\times3$, $f^{1}(\cdot)$ represents the convolution layer whose kernel size is $1\times1$. 
Secondly, we introduce scale-aware attention to aggregate discriminative contextual information. The output of SSAM ($\textbf{F}_{S}$) is multiplied by the output of the CFGB branch. The output of feature reweighting branch $\textbf{X}_{B}$ is given in Equation \ref{x2}.

\begin{equation}\label{x2}
    \textbf{X}_{B} = f^{3}(\mathbb{UP}(\textbf{F}_{i})) \odot \textbf{F}_{S}(\textbf{F}_{R})
\end{equation}

where $\mathbb{UP}(\cdot)$ represents the bilinear upsample operation.
The two branches $\textbf{X}_{A}$ and $\textbf{X}_{B}$ are fused with an element-wise summation to produce $\textbf{F}_{out}$ . Finally, the output feature is refined using a $3\times3$ convolution, dropout, and projection layer to produce the deep supervision in the four-level hierarchy. Similarly, the output is passed into a feature refinement block consisting of $3\times3$ convolution to generate the output for attention pyramid fusion of the next stage.

\begin{equation}
    \textbf{F}_{out} = \textbf{X}_{A} + \textbf{X}_{B}
\end{equation}
Where $\textbf{F}_{out}$ represents the fused coarse and low-level features.

\section{Global Feature Upsample (GFU)}
\label{mehtod:global_feature_up_module}
 In typical U-Net or encoder-decoder architectures, the extracted high-level feature maps are directly upsampled in the decoding part to restore the original resolution. However, if the high-level feature can be fully propagated, it will obtain an accurate segmentation map. We applied a depthwise convolution with stride to $F_{5}$ to obtain enough information to produce a coarse semantic segmentation. On the other hand, the features from the attention-based feature pyramid network are rich in details information about different objects, which are beneficial in recognizing the boundaries between objects. We can get an accurate semantic map by combining the details and coarse semantic information. The proposed GFU module is shown in Fig.6. For the high-level features, the proposed global feature upsampling module firstly performs a bilinear upsampling on the input feature to sample it to the same dimensions of AFPN features. We then applied $1\times1$ convolution and non-linearity followed by global average pooling to generate a global context and sample it to the same channels of AFPN. In the meantime, we performed $1\times1$ convolution with batch normalization and relu on the feature from AFPN and added high-level features. Finally, the output feature is obtained by applying $1\times1$ convolution with batch normalization and relu. The proposed GFU can process and integrate features maps from different stages more efficiently. The output of this block $ \textbf{X}_{GFU}$ can be summarized as follows:
\begin{equation}
    \textbf{X}_{GFU} = f^{1}(\textbf{G}_{pool}(f^{1}(ReLU(\mathbb{UP}(\textbf{X}_{F}))))+f^{1}(\textbf{X}_{P}))
\end{equation}
where ReLU is the Rectified Linear Unit, $\textbf{X}_{F}$ denotes the feature map from the feature adaptation block (FAB), $\textbf{X}_{P}$ represents the feature from the attention pyramid fusion (APF) module. 

\begin{table*}
	\caption{\MakeUppercase{The Per-class, class, and category IoU  evaluation on the Cityscapes Test set."}}
	\label{per_class_cityscapes}
    	\vspace{1ex}
    	\begin{center}
    	\renewcommand\arraystretch{1.5}
    	\resizebox{1\textwidth}{!}{
    	\small
			\begin{tabu}{l|l|l|l|l|l|l|l|l|l|l|l|l|l|l|l|l|l|l|l|l}
				\hline
				\bf{Method}&\bf{Road}&\bf{S.Walk}&\bf{Build}&\bf{Wall}&\bf{Fence}&\bf{Pole}&\bf{T-Light}&\bf{T-Sign}&\bf{Veg}&\bf{Terrain}&\bf{Sky}&\bf{Person}&\bf{Rider}&\bf{Car}&\bf{Truck}&\bf{Bus}&\bf{Tra}&\bf{Motor}&\bf{Bic}&\bf{mIoU}\\
				\hline
				\hline
				CRF-RNN\cite{zheng2015conditional}&96.3&73.9&88.2&47.6&41.3&35.2&49.5&59.7&90.6&66.1&93.5&70.4&34.7&90.1&39.2&57.5&55.4&43.9&54.6&62.5\\
				FCN\cite{long2015fully}&97.4&78.4&89.2&34.9&44.2&47.4&60.1.5&65.0&91.4&69.3&93.9&77.1&51.4&92.6&35.3&48.6&46.5&51.6&66.8&65.3\\
				DeepLabv2\cite{chen2017deeplab}&-&-&-&-&-&-&-&-&-& -&-&-&-&-&-&-&-&-&-&70.4\\
				Dilation10\cite{yu2016multi}&97.6&79.2&89.9&37.3&47.6&53.2&58.6&65.2&91.8&69.4&93.7&78.9&55.0&3.3&45.5&53.4&47.7&52.2&66.0&67.1\\
				AGLNet\cite{zhou2020aglnet}&97.8& 80.1& 91.0&51.3&50.6&58.3&63.0&68.5&92.3&71.3&94.2&80.1&59.6&93.8&48.4&68.1&42.1&52.4& 67.8&70.1\\
				BiSeNetV2\slash BiSeNetV2\_L\cite{yu2021bisenet}&-&-&-&-&-&-&-&-&-& -&-&-&-&-&-&-&-&-&-&73.2\\
				LBN-AA\cite{dong2020real}&98.2&84.0&91.6&50.7&49.5&60.9&69.0&73.6&92.6& 70.3&94.4&83.0&65.7&94.9&62.0&70.9&53.3&62.5&71.8-&73.6\\
				\hline
				S\textsuperscript{2}-FPN18&98.2&84.7&92.0&50.2&54.9&62.9&71.3&75.6&93.0&70.0&94.7&84.6&67.4&95.2&65.1&79.4&68.7&65.1&73.9&76.2\\
				S\textsuperscript{2}-FPN34&98.4&85.5&92.7&56.2&57.6&64.7&72.7&76.2&93.3&70.8&94.9&85.2&69.0&95.52&66.7&80.8&69.0&66.4&74.3&77.4\\
				S\textsuperscript{2}-FPN34M&98.1&84.5&92.7&51.5&57.3&68.5&76.5&79.3&93.5&71.4&94.5&86.4&68.4&95.3&61.5&80.4&72.6&68.0&76.5&77.8\\
				\hline
			\end{tabu}}
	\end{center}
\end{table*}

\begin{figure}
	\begin{center}
		\includegraphics[width=0.9\linewidth]{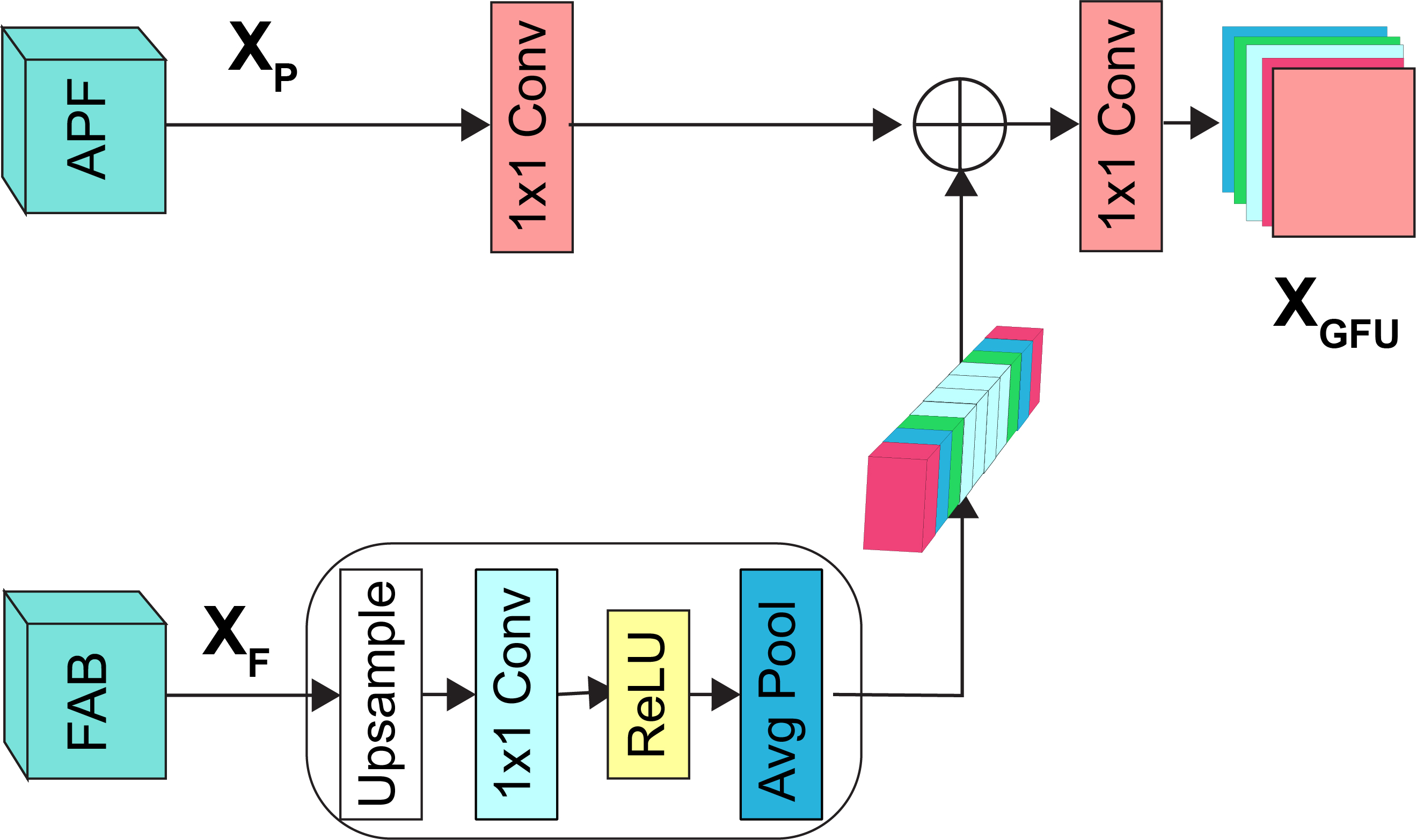}
		\caption{Illustration of global feature upsampling module.}
		\label{fig:stage_transions}
	\end{center}
\end{figure}

\begin{table*}
	\caption{\MakeUppercase{Comparison between the proposed method S\textsuperscript{2}-FPN and the other SOTA methods on the Cityscapes test dataset. We report the backbone, input resolution, GPU type, number of parameters (M), Flops (G), evaluation split (set), achieved accuracy (mIoU), and the inference speed (FPS)}}
	\label{miou_cityscapes}
	\begin{center}
		\begin{adjustbox}{width=1\textwidth}
			\small
			\begin{tabu}{l|l|l|l|l|l|l|l|l}
				\hline
				Method&Backbone&Resolution&GPU&Parameters(M)&Flops (G)&test set&mIoU&FPS\\ 
				\hline\hline
				PSPNet\cite{zhao2017pyramid}&ResNet101&713 $\times$713&-&250.8&412.2& \checkmark&81.2&0.78\\
				DeepLab\cite{chen2017deeplab}&VGG16&512$\times$1024&Titan X&262.1&457.8& \checkmark&63.1&0.25\\
				\hline
				ENet\cite{zheng2015conditional}&No&640 $\times$360&TitanX&0.4&3.8& \checkmark&58.3& 135.4\\
				ICNet\cite{chen2014semantic}&PSPNet50&1024$\times$2048&TitanX&26.5&28.3&\checkmark&69.5&30.3\\
				DABNet\cite{long2015fully}&No &512$\times$1024& GTX 1080Ti&0.76&10.4&\checkmark&70.1&104\\
				DFANet-A\cite{yu2016multi}&XceptionA&1024$\times$1024&Titan X&7.8&3.4&\checkmark &71.3&100\\
				DFANet-B\cite{yu2016multi}&XceptionB&1024$\times$1024&Titan X&4.8&2.1&\checkmark&67.1&120\\
				BiSeNet1\cite{yu2018bisenet}&Xception39&768$\times$1536&NVIDIA Titan X&5.8&14.8&\checkmark &68.4&105.8\\
				BiSeNet2\cite{yu2018bisenet}&ResNet18 &768$\times$1536&NVIDIA Titan X&49&54.02&\checkmark &74.8&65.5\\
				FasterSeg\cite{chen2019fasterseg}&No &1024$\times$2048&GTX 1080Ti&-&&\checkmark &71.5&163.9\\
				TD4-Bise18\cite{hu2020temporally}&BiseNet18&1024$\times$2048&Titan Xp&-&&\checkmark &74.9&-\\
				FANet-18\cite{hu2020real}&ResNet18&1024$\times$2048&Titan X&-&49&\checkmark &74.4&72\\
				FANet-34\cite{hu2020real}&ResNet34&1024$\times$2048&Titan X&-&65&\checkmark &75.5&58\\
				LBN-AA\cite{dong2020real}&LBN-AA+MobileNetV2&488$\times$896&Titan X&6.2&49.5&\checkmark &73.6&51.0\\
				AGLNet\cite{dong2020real}&No&512$\times$1024&GTX 1080Ti&1.12&13.88&\checkmark &71.3&52.0\\
				BiSeNetV2\cite{yu2021bisenet}&No&512$\times$1024&GTX 1080Ti&49.0&21.2&\checkmark &72.6&156\\
				BiSeNetV2-L\cite{yu2021bisenet}&No&512$\times$1024&GTX 1080Ti&&118.5&\checkmark &75.3&47.3\\
				HMSeg\cite{li2020humans}&No&768$\times$1536&GTX 1080Ti&2.3&&\checkmark &74.3 &83.2\\
				TinyHMSeg\cite{li2020humans}&No&768$\times$1536&GTX 1080Ti&0.7&&\checkmark &71.4&172.4\\
				STDC1-Seg50\cite{fan2021rethinking}&STDC1&512$\times$1924&GTX 1080Ti &8.4&-&\checkmark &71.9&250.4\\
				STDC2-Seg50\cite{fan2021rethinking}&STDC2&512$\times$1024&GTX 1080Ti &12.5&-&\checkmark &73.4&188.6\\
				STDC1-Seg75\cite{fan2021rethinking}&STDC1&768$\times$1536&GTX 1080Ti &8.4&-&\checkmark &75.3&126.7\\
				STDC2-Seg75\cite{fan2021rethinking}&STDC2&768$\times$1536&GTX 1080Ti&12.5&-&\checkmark &76.8&97.0\\
				\hline 
				S\textsuperscript{2}-FPN18&ResNet18&512$\times$1024&GTX 1080Ti&17.8&29.1&\checkmark &76.2&87.3\\
				S\textsuperscript{2}-FPN34&ResNet34&512$\times$1024&GTX 1080Ti&27.9&48.4&\checkmark &77.4&67\\
				S\textsuperscript{2}-FPN34M&ResNet34&512$\times$1024&GTX 1080Ti&27.9&190&\checkmark &77.8&30.5\\
				\hline
			\end{tabu}
		\end{adjustbox}
	\end{center}
\end{table*}

\section{Experimental Results}
\label{experimental_results}
We evaluate the effectiveness of our proposed segmentation network on two datasets: Cityscapes \cite{cordts2016cityscapes} and Camvid \cite{brostow2009semantic}. We first introduce the training details. Then, we provide our full report of S\textsuperscript{2}-FPN compared to other state-of-the-arts methods on the two benchmarks. The following criteria are used for the performance evaluation, i.e. accuracy,speed,backbone,input resolution and parameters in Cityscapes. Experimental results demonstrate that S\textsuperscript{2}-FPN obtains a trade-off between accuracy and speed on both Camvid and Cityscapes datasets. In the following subsection, we provide further details.

\subsection{Training setting details}
All the experiments are implemented with one Nvidia GTX 1080Ti GPU, on the PyTorch platform \cite{paszke2016enet}.
We use ResNet with weights pre-trained on ImageNet \cite{russakovsky2015imagenet} as our backbone. The Adam optimizer \cite{kingma2014adam} with initial learning rate of $3e-4$ is used to train the model on both Camvid and Cityscapes datasets. We use weight decay of 5e-6. Following the previous methods settings \cite{zhao2017pyramid,chen2017deeplab}, we decay the learning rate using polynomial learning rate scheduling using equation $(1-\frac{iter}{max_iter})^{power}$. The model has been trained for 500 epochs on Cityscapes dataset . For Camvid dataset we have train for 180 epochs when the model configured with ResNet18 and 150 when configured modified ResNet34. All BatchNorm layers in our architecture replace by InPlaceABN-Sync\cite{rotabulo2017place}. To avoid over-fitting, We apply a number of data augmentations techniques as follows: random resize with scale range [0.75,1.0,1.25,1.5,1.75, 2.0], random horizontal flipping  and random cropping images to a resolution of $512\times1024$ for Cityscapes and $360\times360$ for Camvid.\\
We adopt a well-known metrics to evaluate our model. Frame per-second (FPS) and mean Intersection over Union (mIoU), which measure the semantic segmentation latency and accuracy, respectively. Moreover, we evaluate the computational complexity and memory consumption based on floating point operations (FLOPs) and model parameters (Params), respectively.
\subsection{Cityscapes Dataset}
\subsubsection{Ablation Study}

In this subsection, we conduct a series of experiments to explore the effect of each component (GFU, APF, SSAM, Supervision) in the proposed method. SS-FPN is implemented with a different backbone setting (ResNet18, ResNet34) and we have tested the model with the different settings. 
\begin{table}
	\caption{\MakeUppercase{Comparison between S\textsuperscript{2}-FPN16, S\textsuperscript{2}-FPN34, and S\textsuperscript{2}-FPN34M in terms of parameters, frame per second and accuracy on Cityscapes Validation set}}
	\label{compare_sspnet_settings}
	\centering
	\begin{adjustbox}{width=\linewidth}
		\small
		\begin{tabular}{c|c|c|c|c}
			\hline
			Method&Parameters(M)&Flops&FPS&mIoU\\ 
			\hline\hline
			S\textsuperscript{2}-FPN18&17.8&29.1&87.3&76.4\\
			S\textsuperscript{2}-FPN34&27.9&48.4&67&77.1\\
			S\textsuperscript{2}-FPN34M&27.9&190.0&30.5&77.5\\
			\hline
		\end{tabular}
	\end{adjustbox}
\end{table}

\subsubsection{Ablation Study With Modules}
In this section, we conduct an empirical evaluation to ensure the effectiveness of the different components in our architecture. We show the importance of the scale-aware strip attention module (SSAM), attention pyramid fusion module (APF), Global Feature (GFU) module, and Deep Supervision. The training set is used to train the model, while the validation set is used for evaluation. We first conduct the experiment with ResNet18 as a baseline model. As illustrated in Table\ref{cityscape_ablation_result}, the baseline network obtains 65.7\% mIoU . We further evaluate the effect of global feature upsample module, Fig \ref{main_architecture} by attaching it to the encoder (Fig \ref{main_architecture} (a). Essentially, the GFU is designed to attain category information without using complex decoder blocks. The result in Table \ref{cityscape_ablation_result} shows that adding GFM obtains 71.9 \% mIoU, which add  6.2\%mIoU improvement to the baseline. Moreover, we add APF without supervision or scale-aware strip attention module (replaced with $1\times1$ convolution), it brings 3.7 \%mIoU performance gain. Using deep supervision has added 0.2 \% mIoU improvement. To capture the long-range dependencies, We further evaluate the significance of the scale-aware attention module by adding it into APF with all the other modules. Table \ref{cityscape_ablation_result}, shows this module improves the performance by 0.5\%mIoU. The ablation study and analyses highlighted the importance of each component in our network. S\textsuperscript{2}-FPN has improved the baseline by 10.7 \% mIoU according to our training settings. Thus, it presents an important strategy to design a network with accuracy/speed trade-off.
\begin{figure*}[!h]
	\centering
	\begin{subfigure}[b]{0.19\textwidth}
		\includegraphics[width=\textwidth]{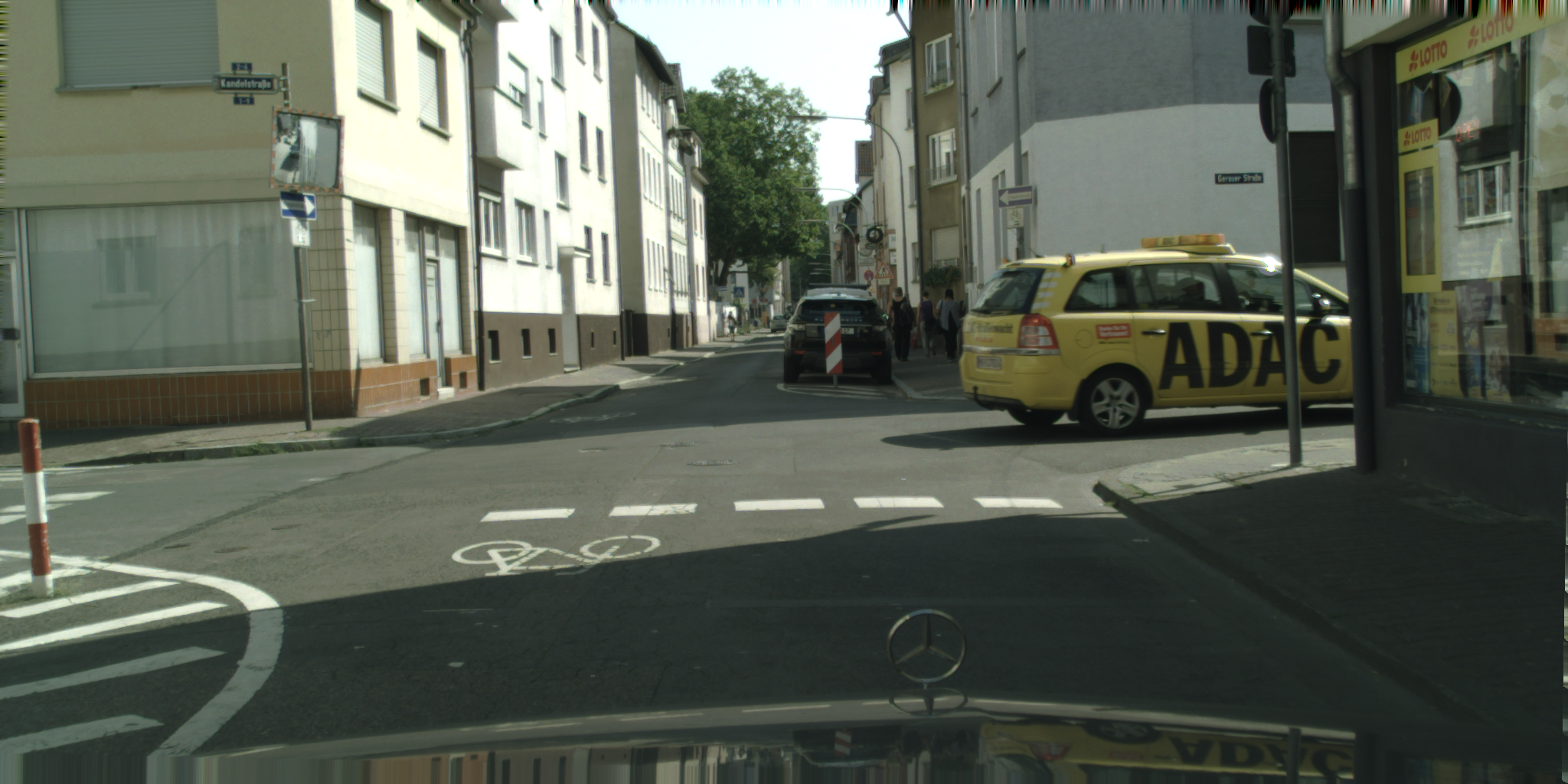}
	\end{subfigure}
	\begin{subfigure}[b]{0.19\textwidth}
		\includegraphics[width=\textwidth]{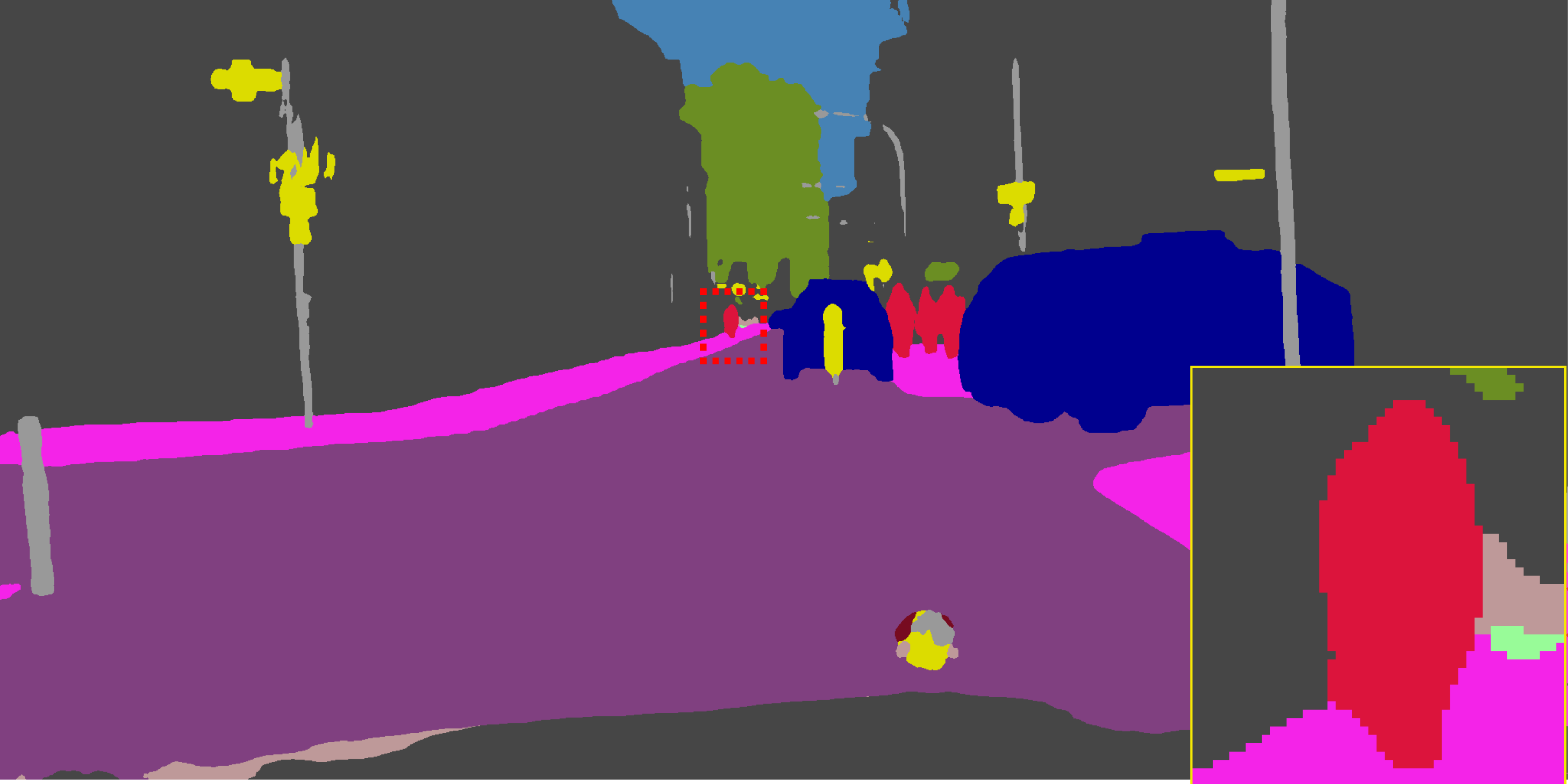}
	\end{subfigure}
	\begin{subfigure}[b]{0.19\textwidth}
		\includegraphics[width=\textwidth]{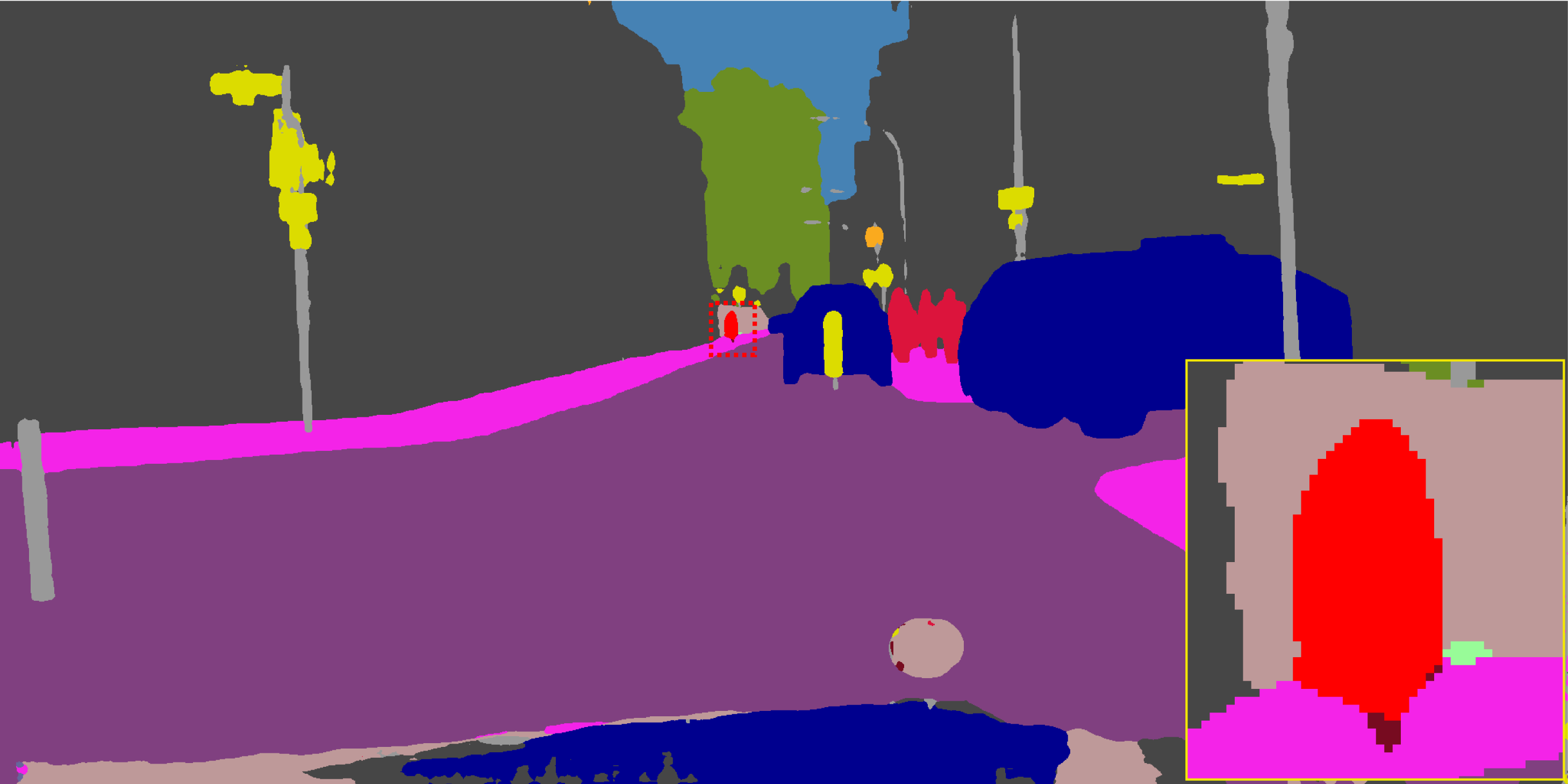}
	\end{subfigure}
	\begin{subfigure}[b]{0.19\textwidth}
		\includegraphics[width=\textwidth]{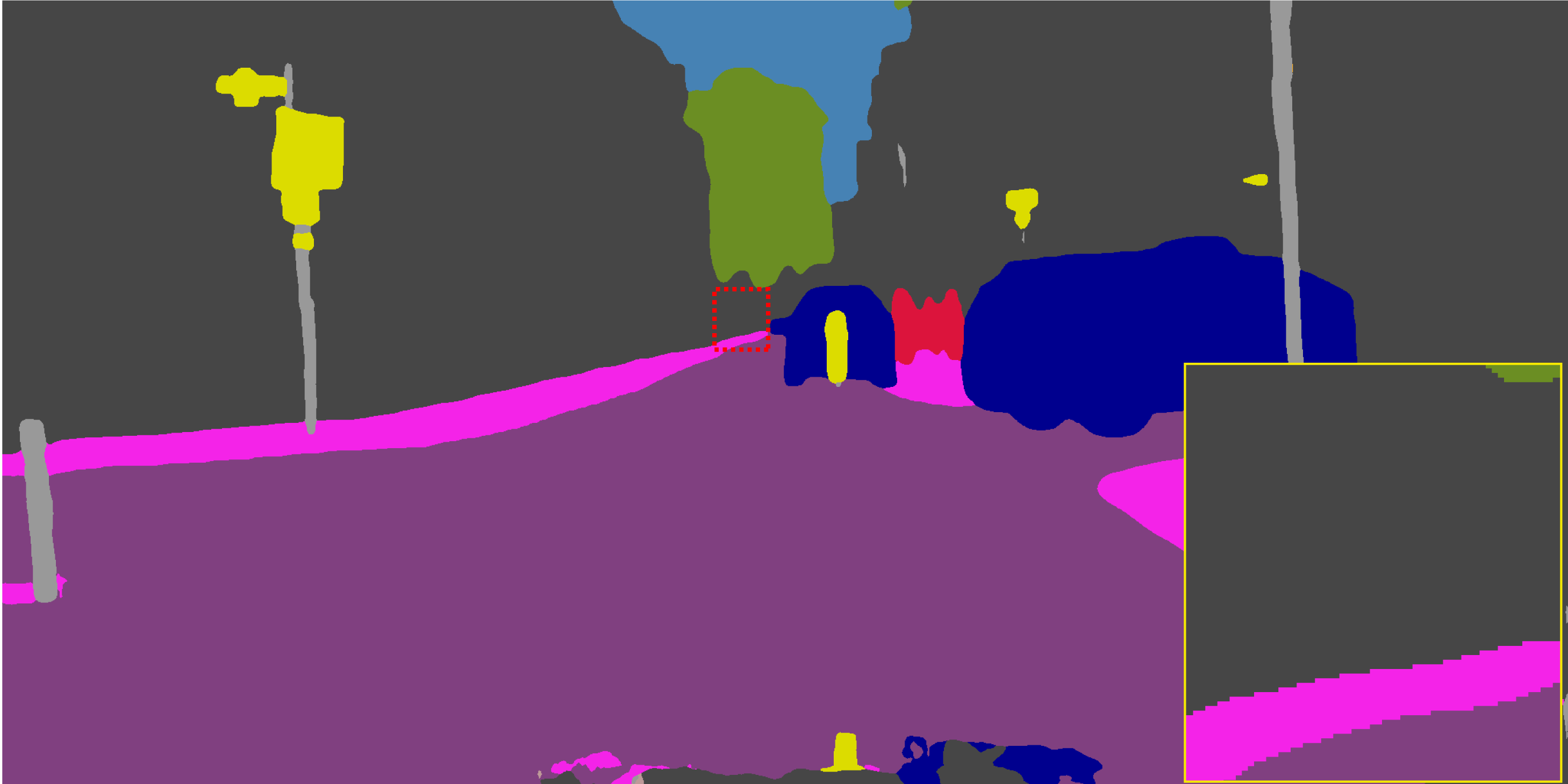}
	\end{subfigure}
	\begin{subfigure}[b]{0.19\textwidth}
		\includegraphics[width=\textwidth]{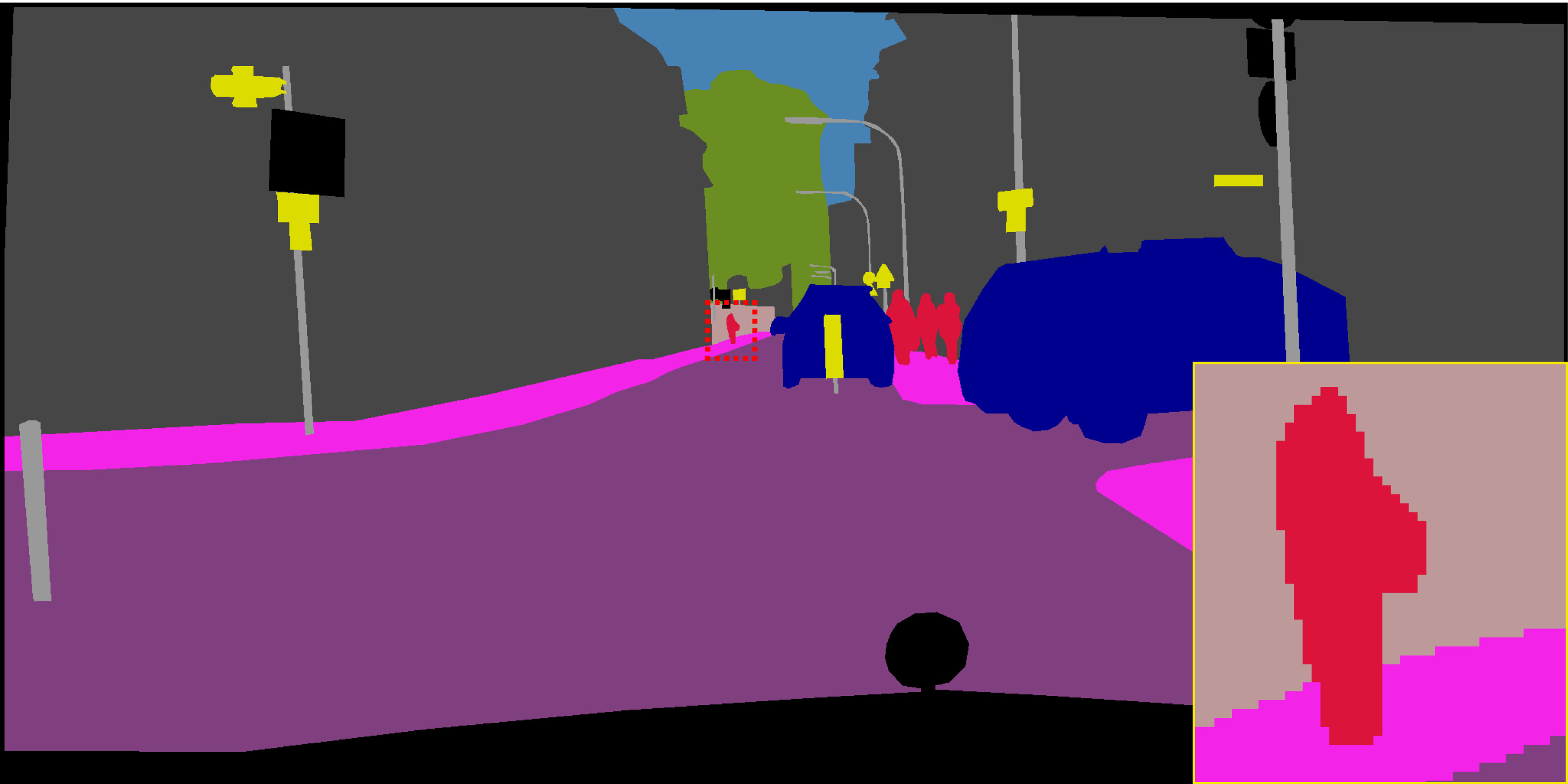}
	\end{subfigure}
	\\
	\begin{subfigure}[b]{0.19\textwidth}
		\centering
		\includegraphics[width=\textwidth]{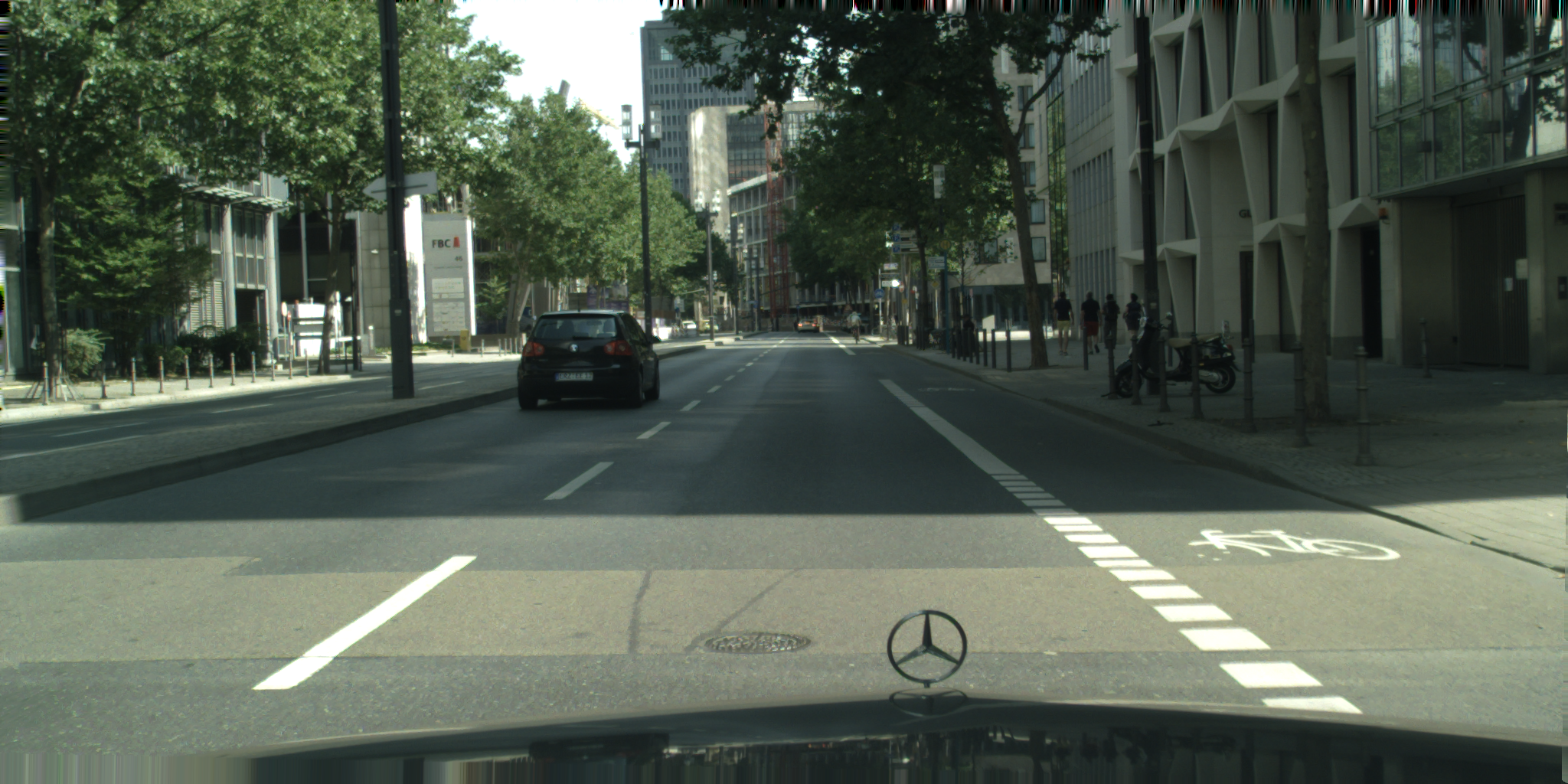}
	\end{subfigure}
	\begin{subfigure}[b]{0.19\textwidth}
		\centering
		\includegraphics[width=\textwidth]{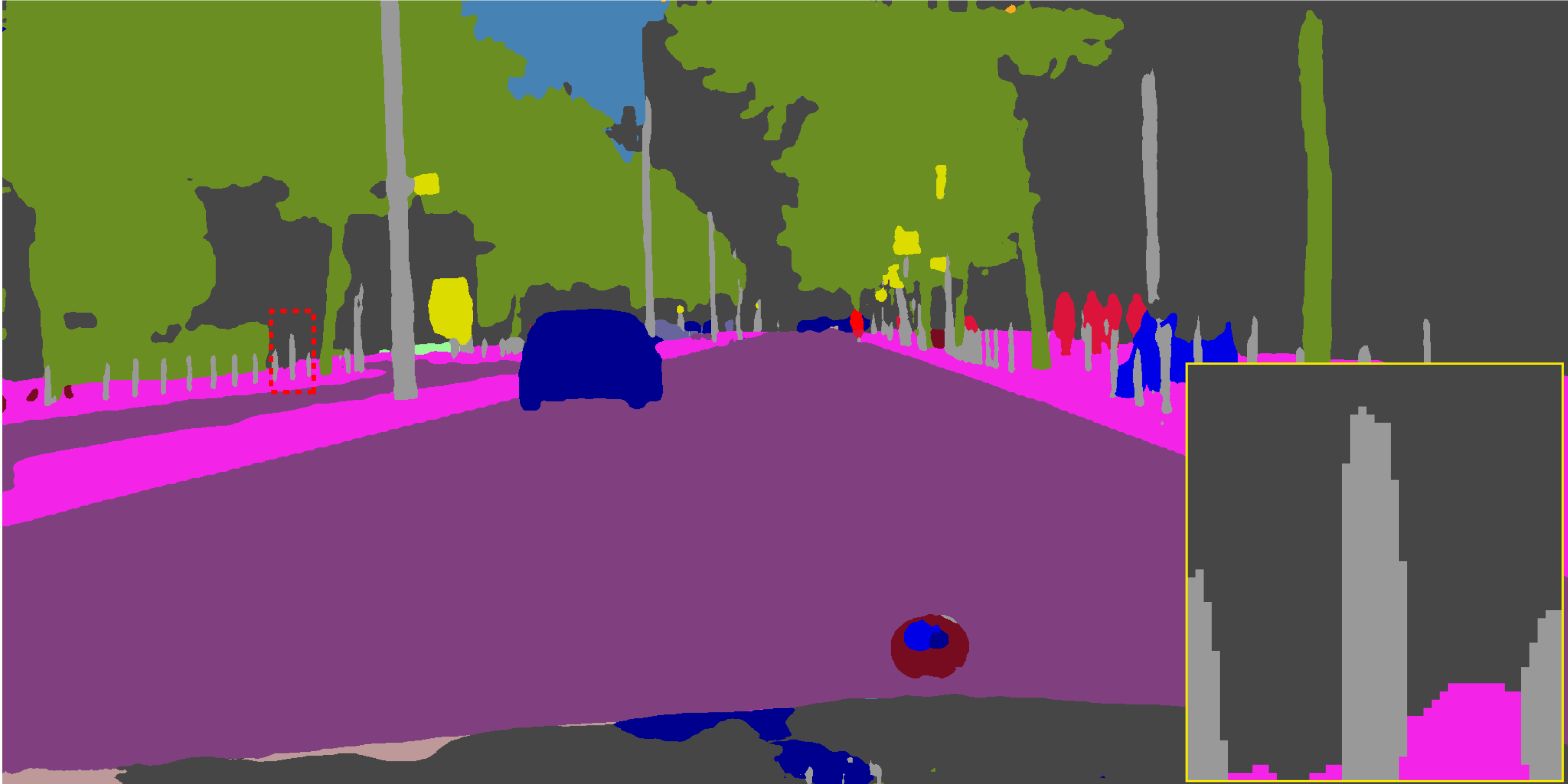}
	\end{subfigure}
	\begin{subfigure}[b]{0.19\textwidth}
		\centering
		\includegraphics[width=\textwidth]{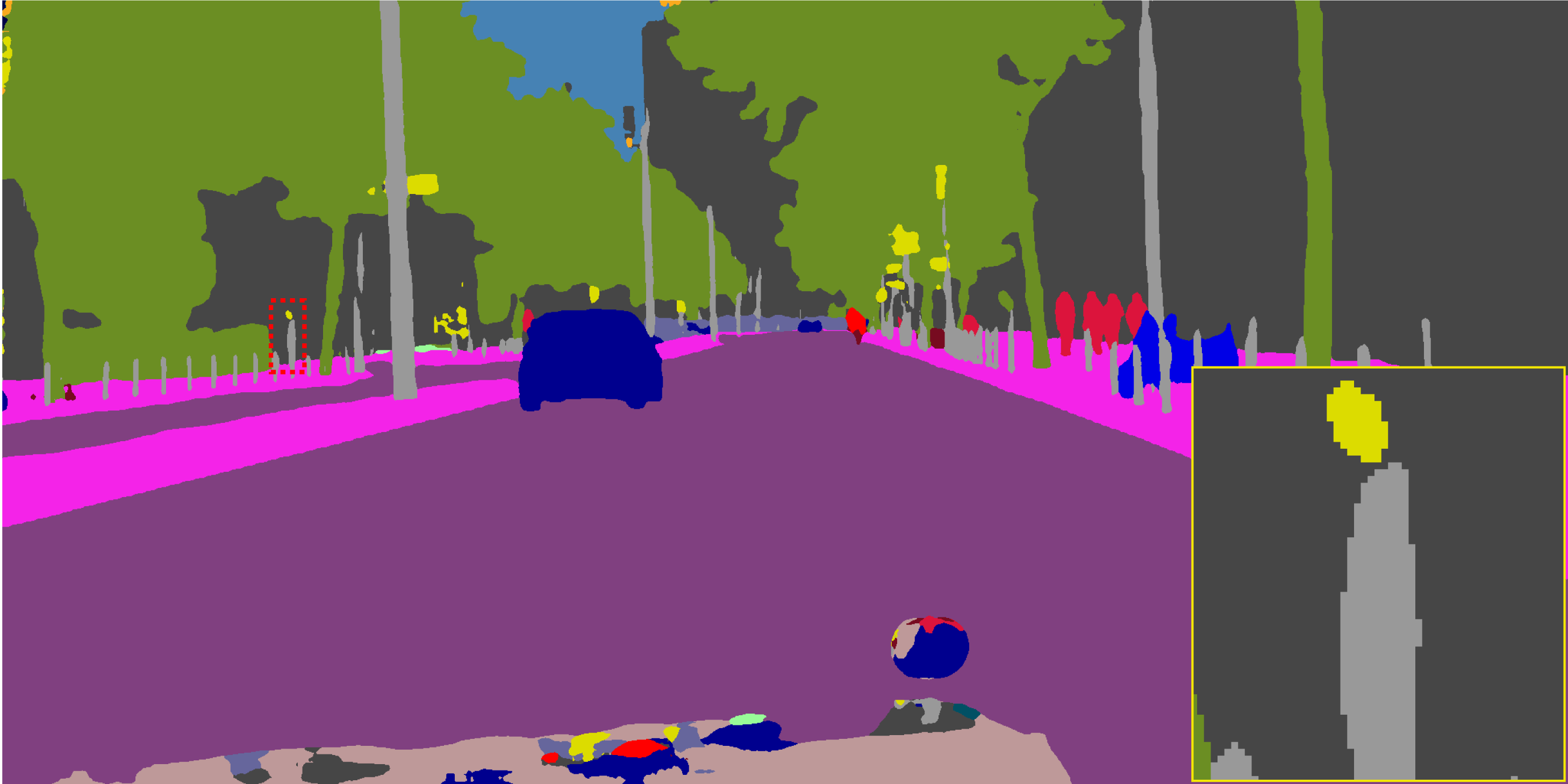}
	\end{subfigure}
	\begin{subfigure}[b]{0.19\textwidth}
		\centering
		\includegraphics[width=\textwidth]{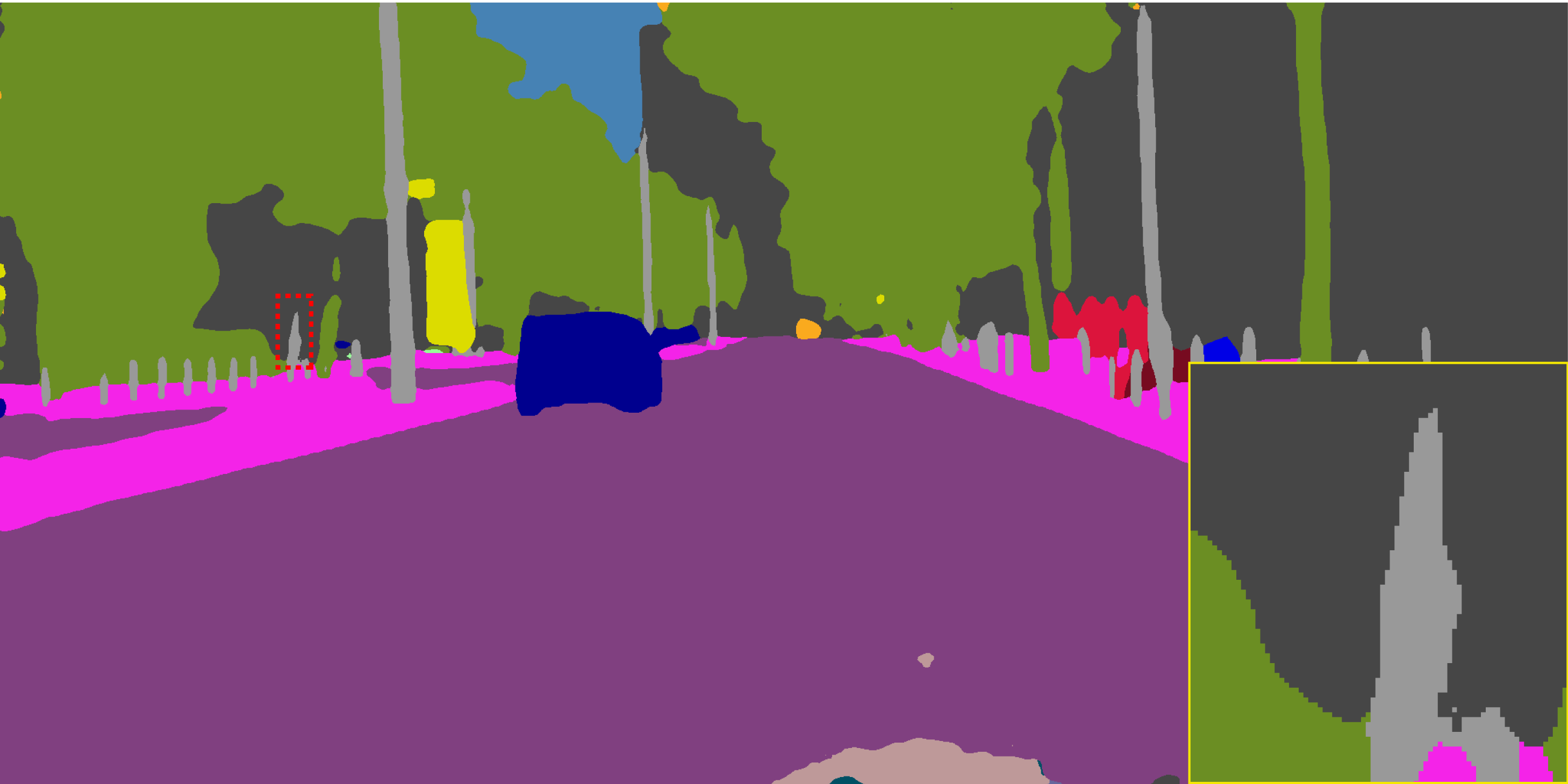}
	\end{subfigure}
	\begin{subfigure}[b]{0.19\textwidth}
		\centering
		\includegraphics[width=\textwidth]{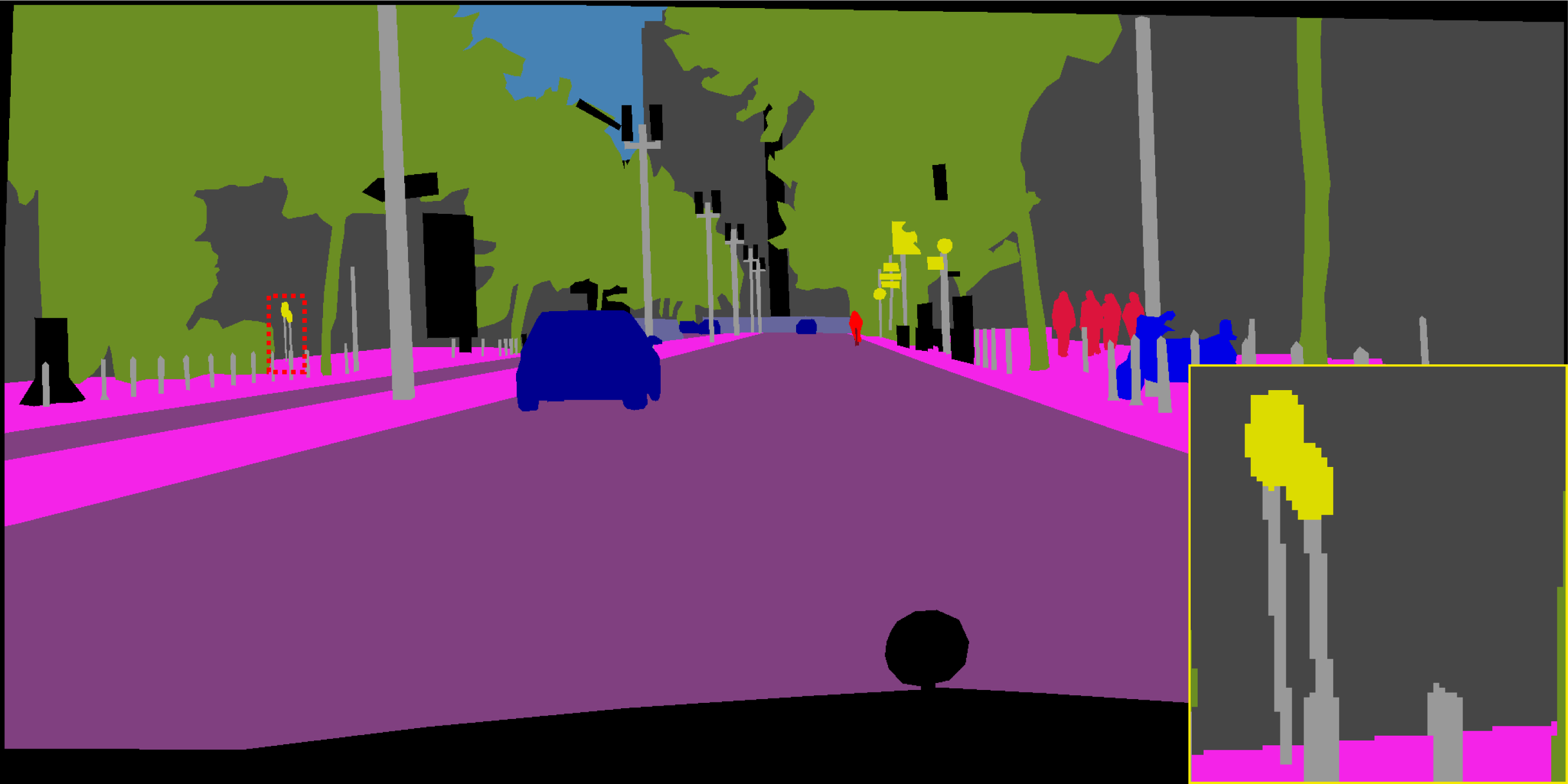}
	\end{subfigure}
	\begin{subfigure}[b]{0.19\textwidth}
		\centering
		\includegraphics[width=\textwidth]{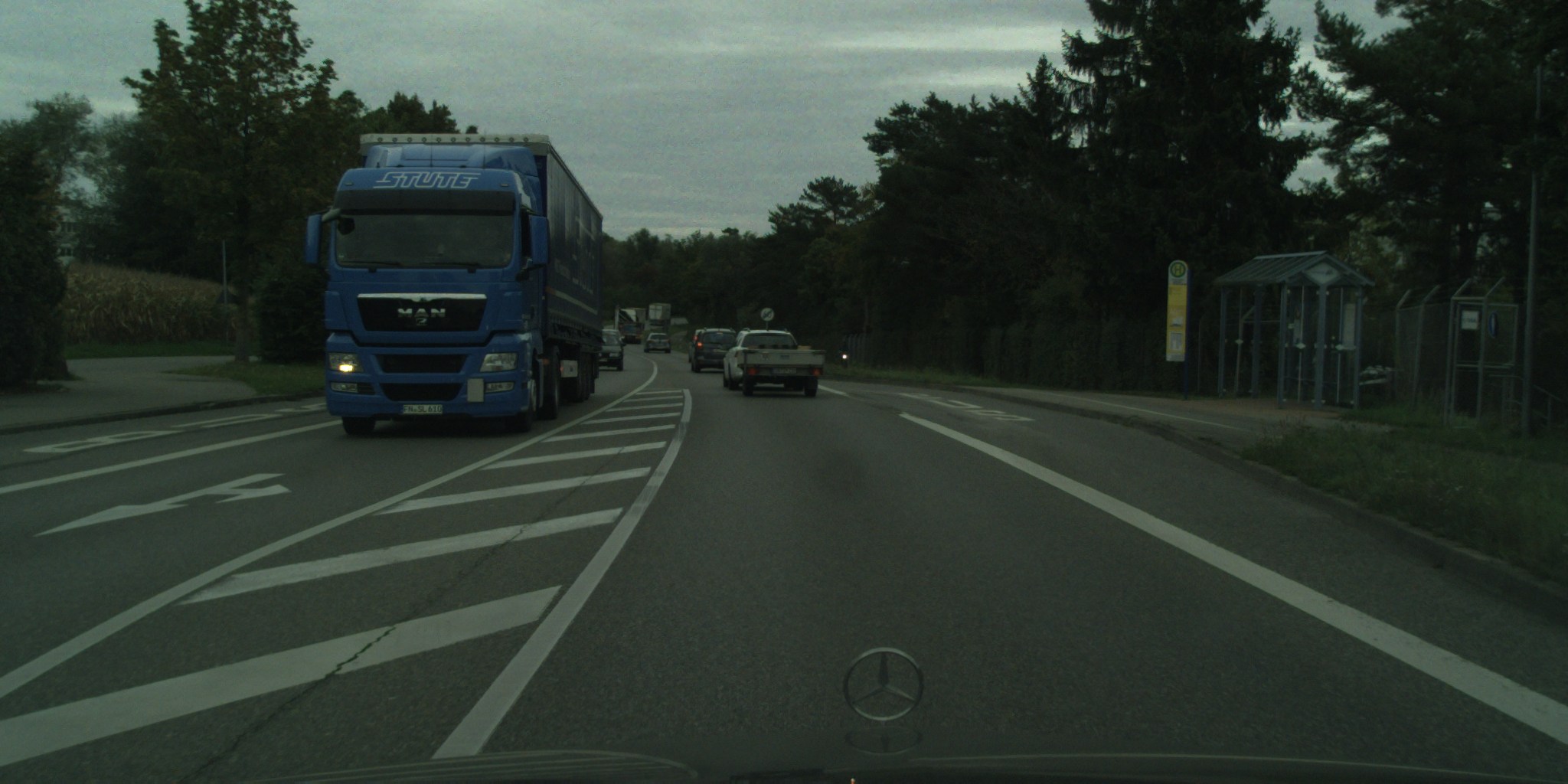}
	\end{subfigure}
	\begin{subfigure}[b]{0.19\textwidth}
		\centering
		\includegraphics[width=\textwidth]{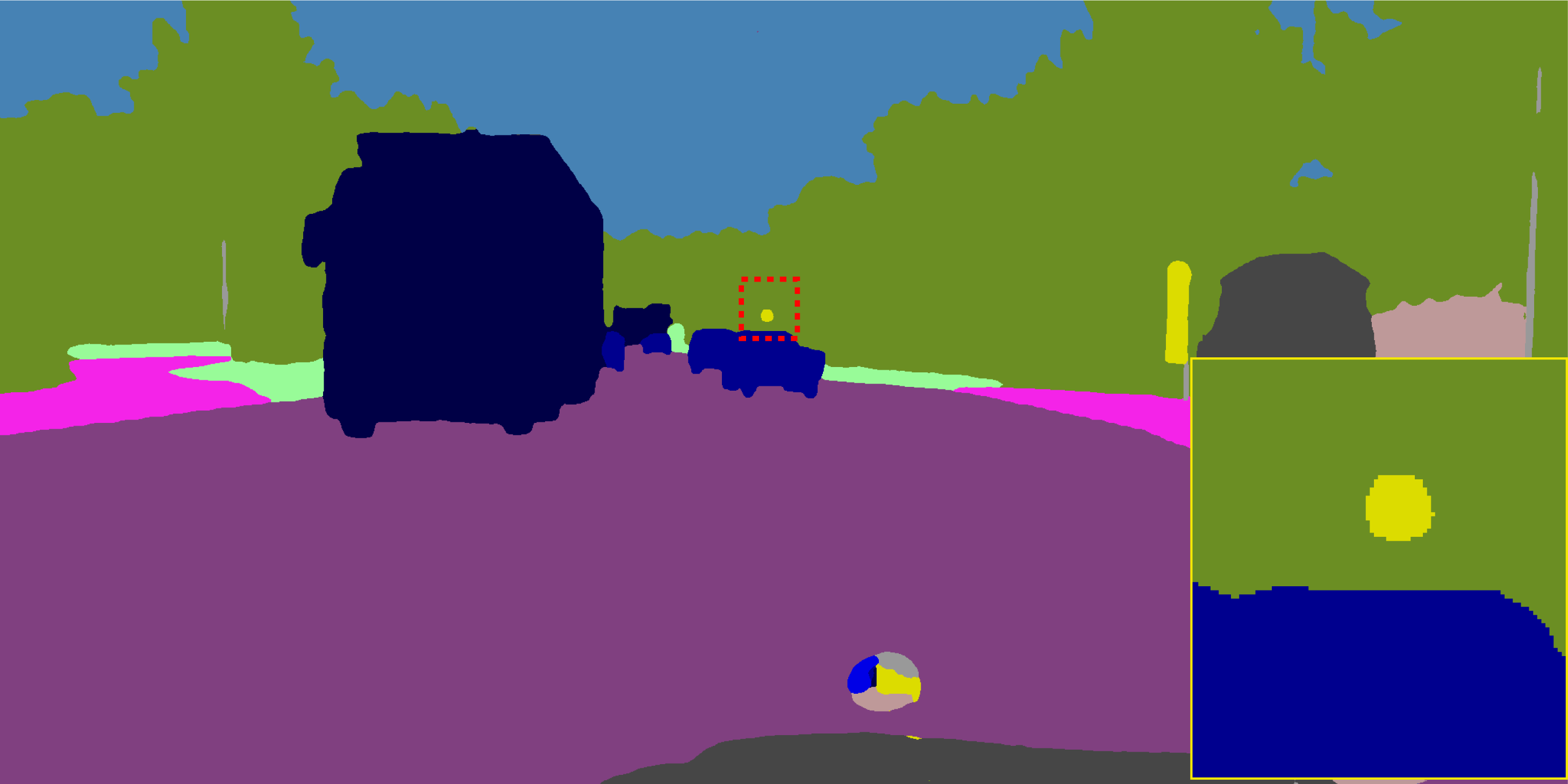}
	\end{subfigure}
	\begin{subfigure}[b]{0.19\textwidth}
		\centering
		\includegraphics[width=\textwidth]{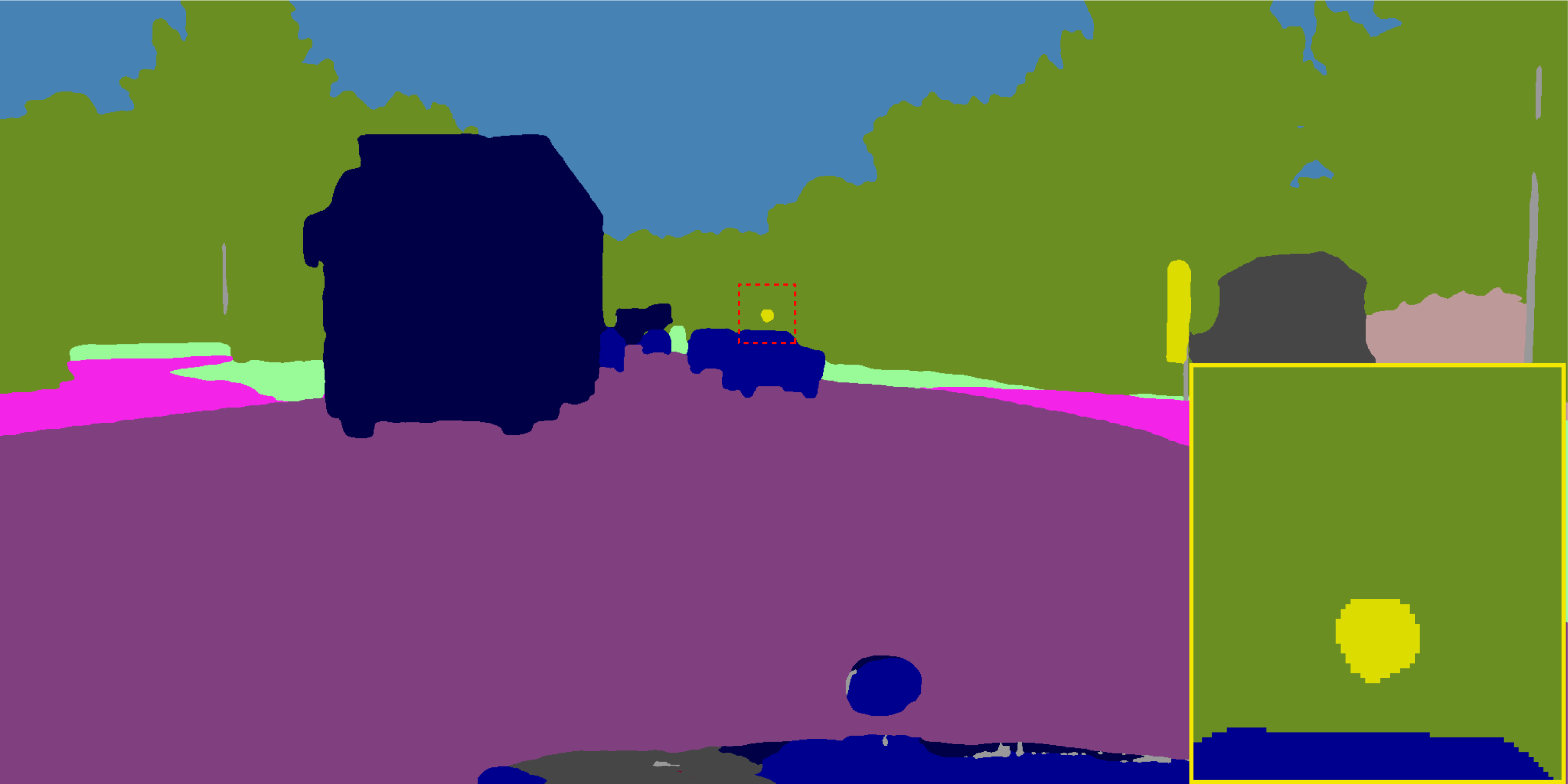}
	\end{subfigure}
	\begin{subfigure}[b]{0.19\textwidth}
		\centering
		\includegraphics[width=\textwidth]{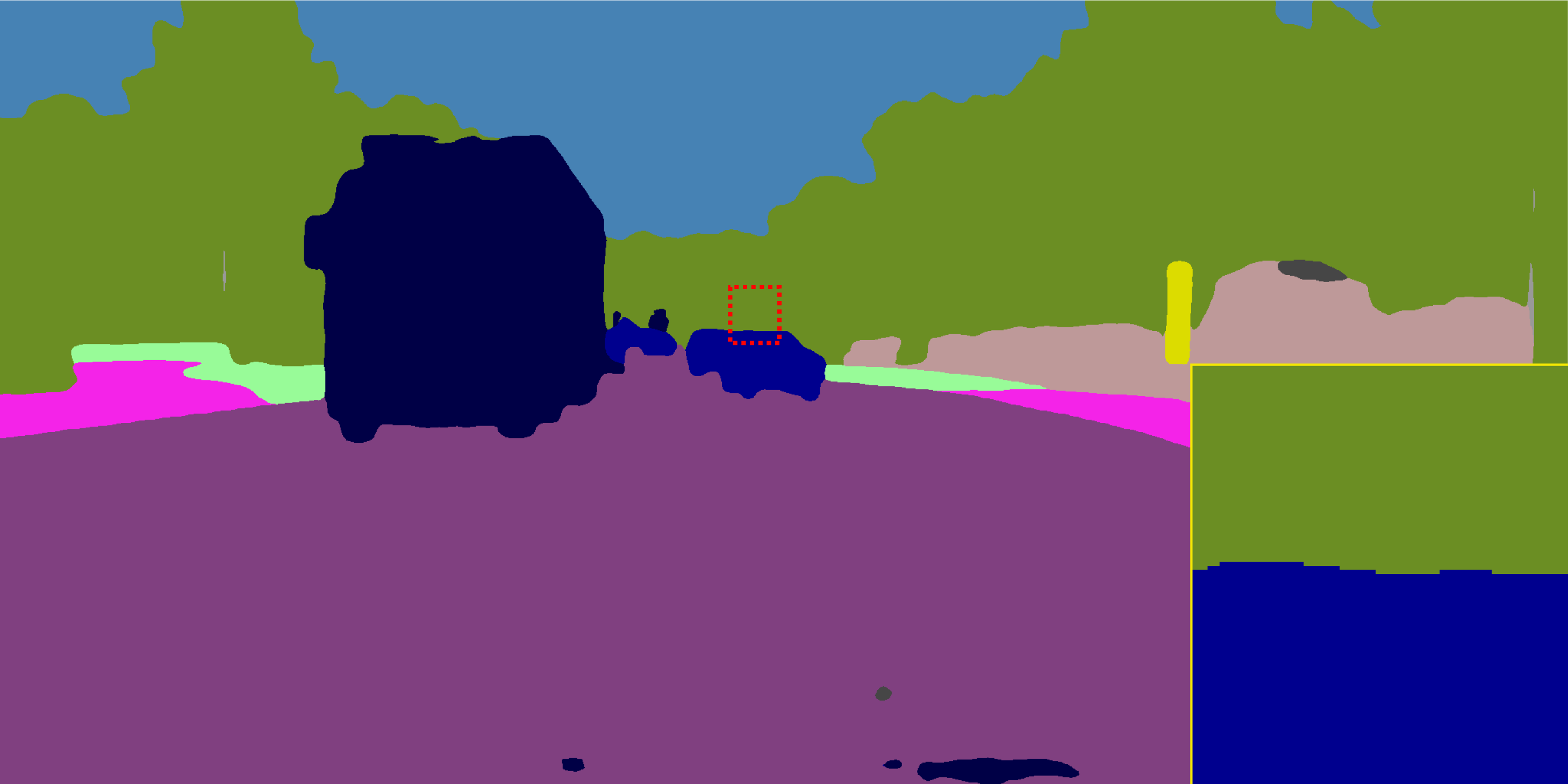}
	\end{subfigure}
	\begin{subfigure}[b]{0.19\textwidth}
		\centering
		\includegraphics[width=\textwidth]{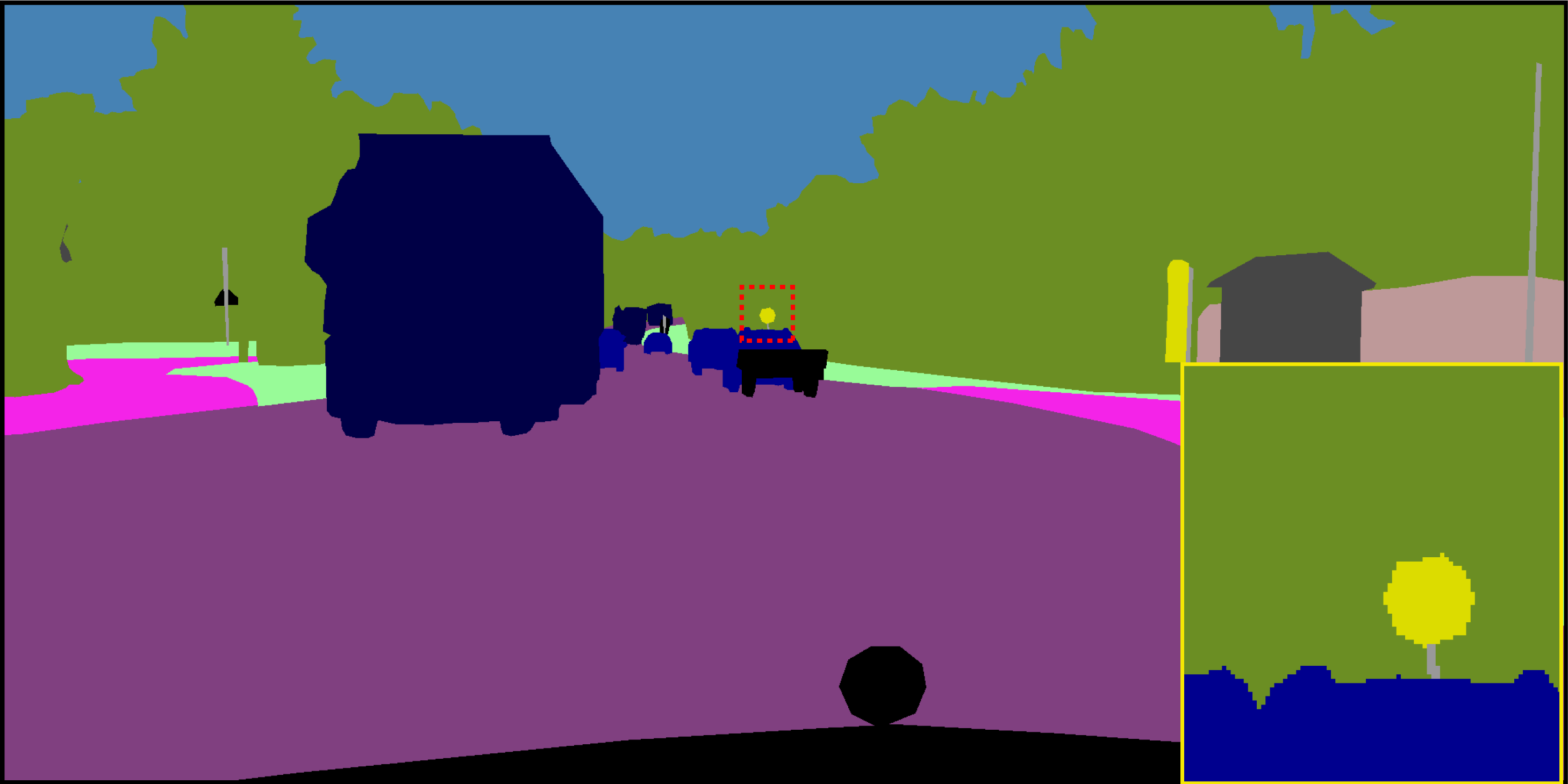}
	\end{subfigure}
	\centering
	\begin{subfigure}[b]{0.19\textwidth}
		\centering
		\includegraphics[width=\textwidth]{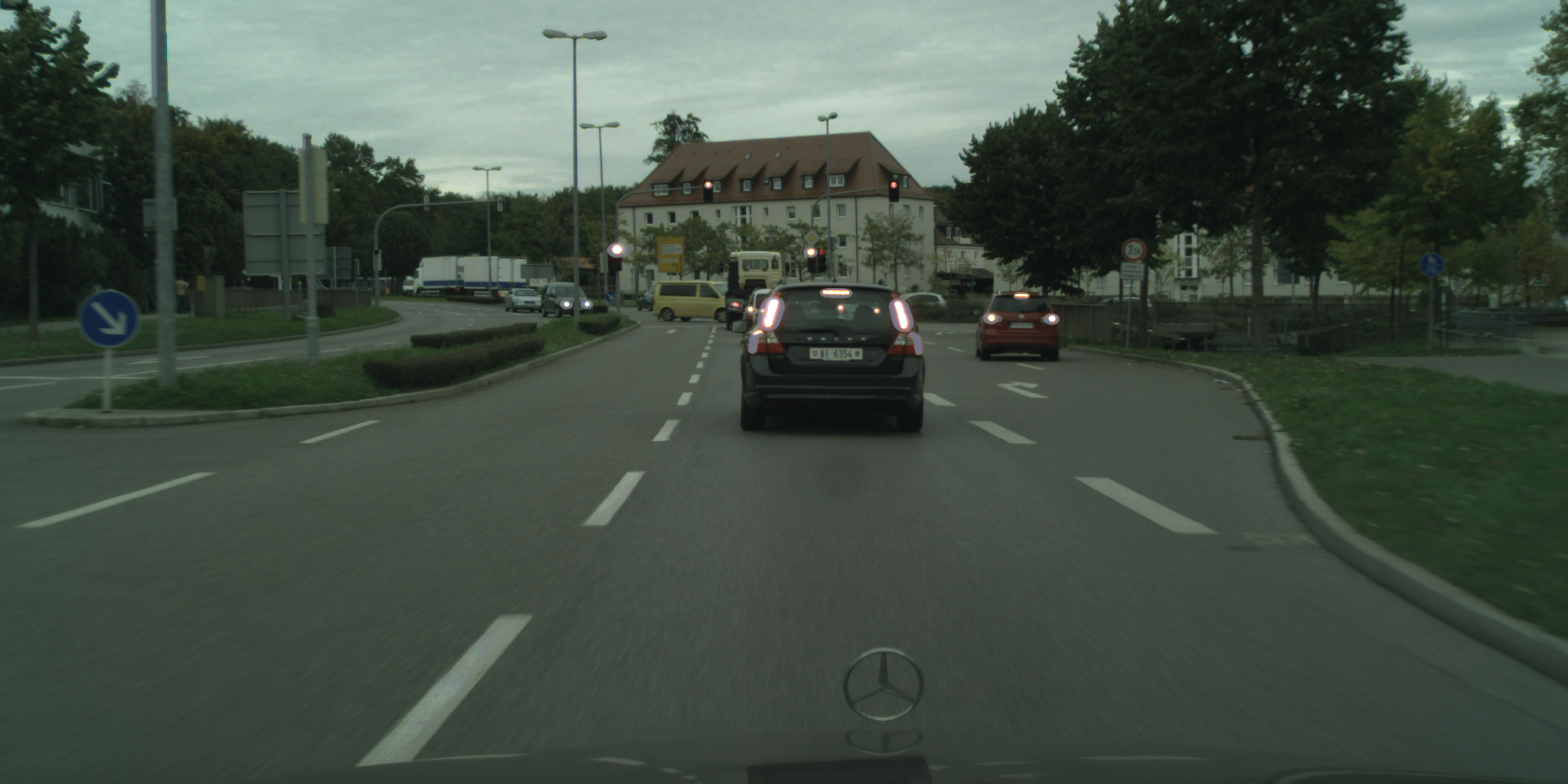}
	\end{subfigure}
	\begin{subfigure}[b]{0.19\textwidth}
		\centering
		\includegraphics[width=\textwidth]{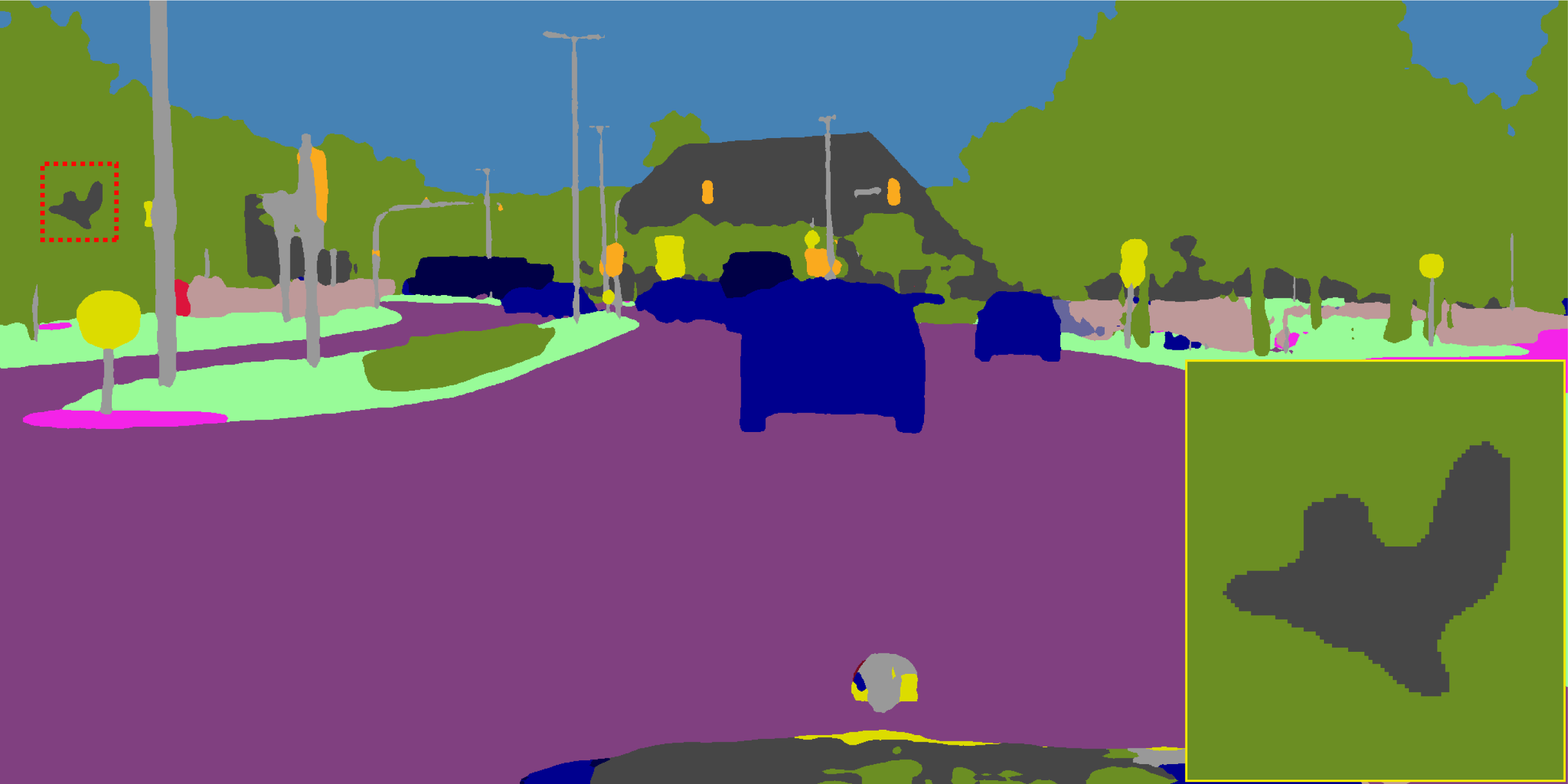}
	\end{subfigure}
	\begin{subfigure}[b]{0.19\textwidth}
		\centering
		\includegraphics[width=\textwidth]{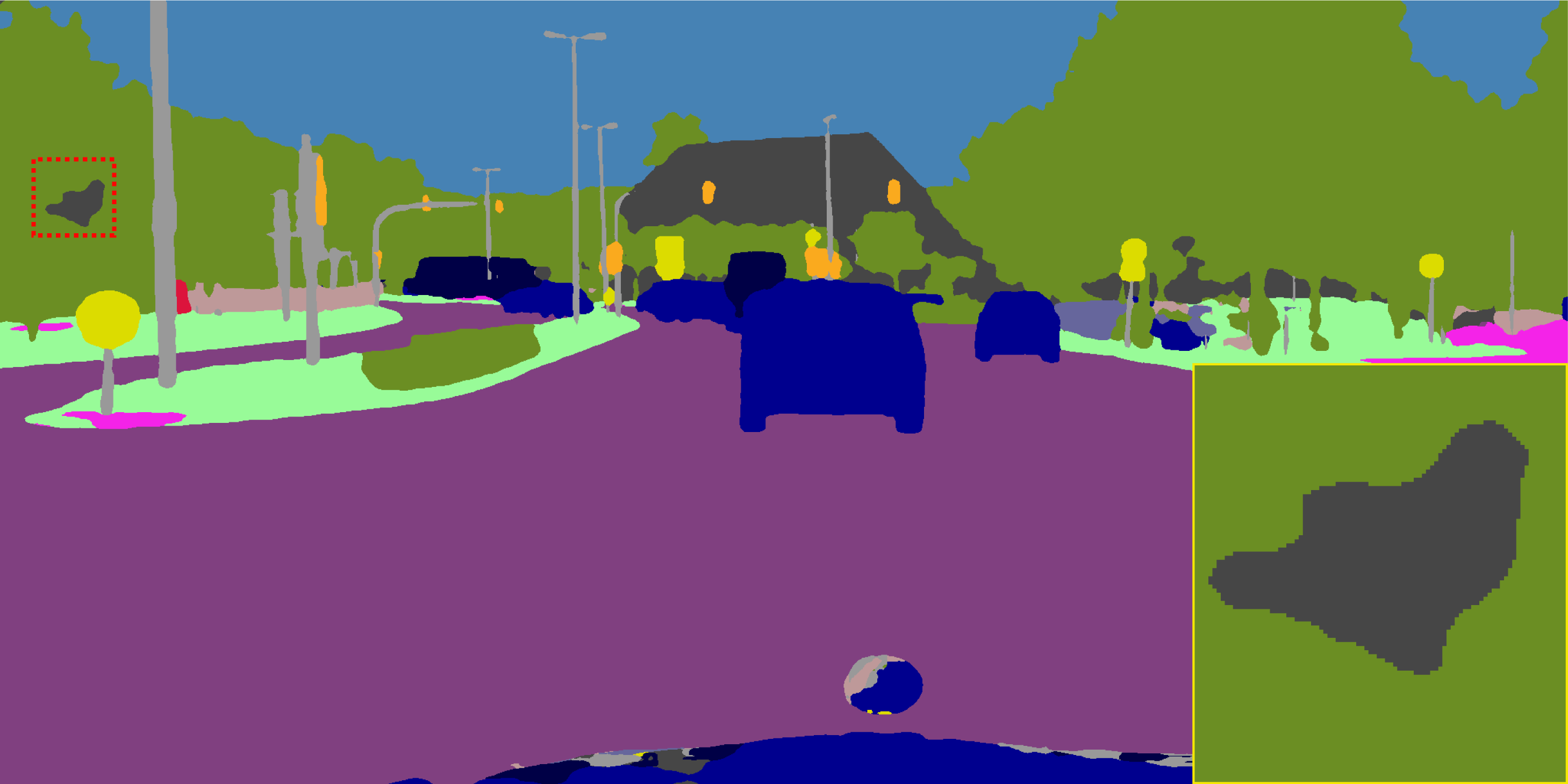}
	\end{subfigure}
	\begin{subfigure}[b]{0.19\textwidth}
		\centering
		\includegraphics[width=\textwidth]{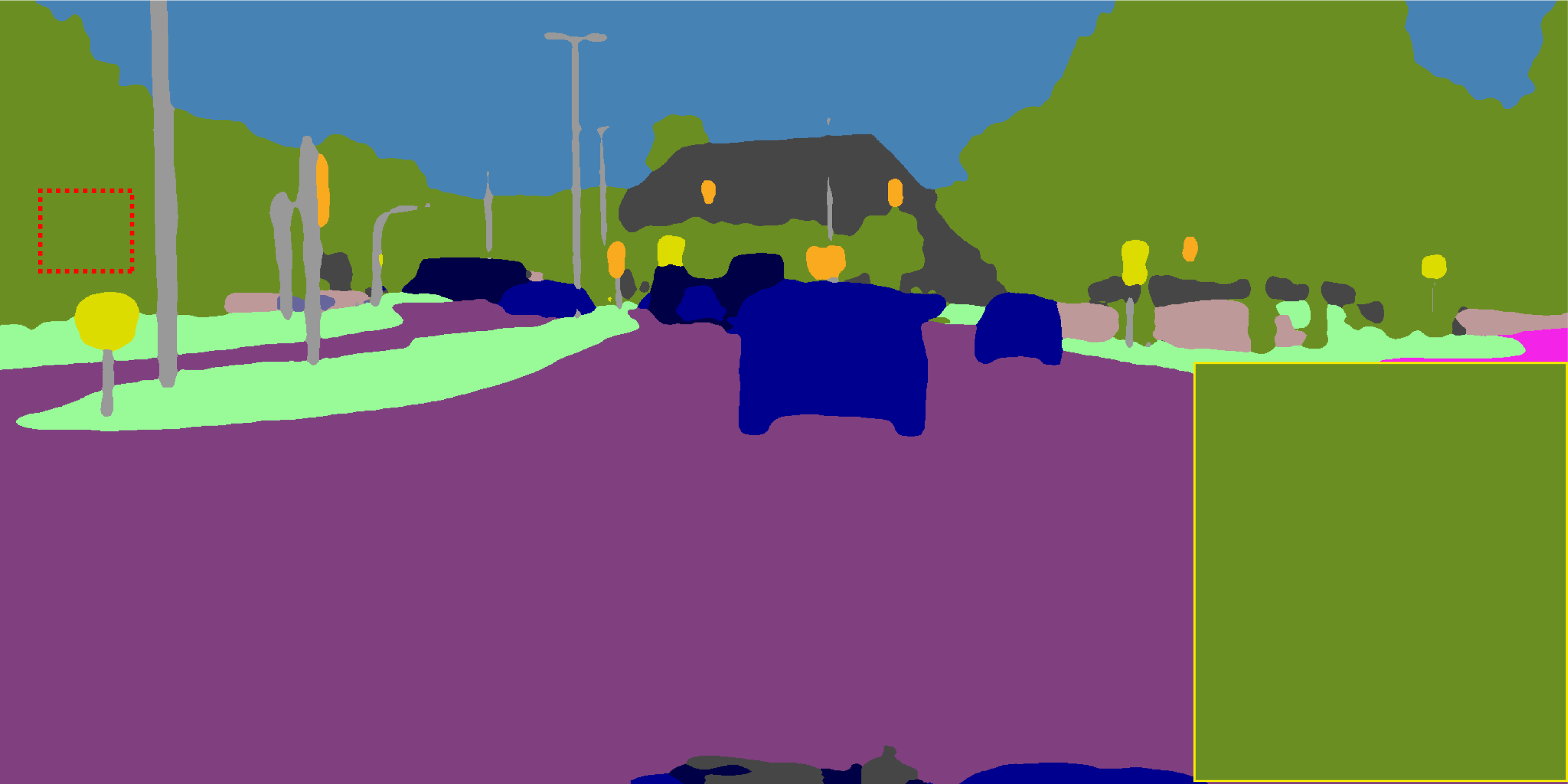}
	\end{subfigure}
	\begin{subfigure}[b]{0.19\textwidth}
		\centering
		\includegraphics[width=\textwidth]{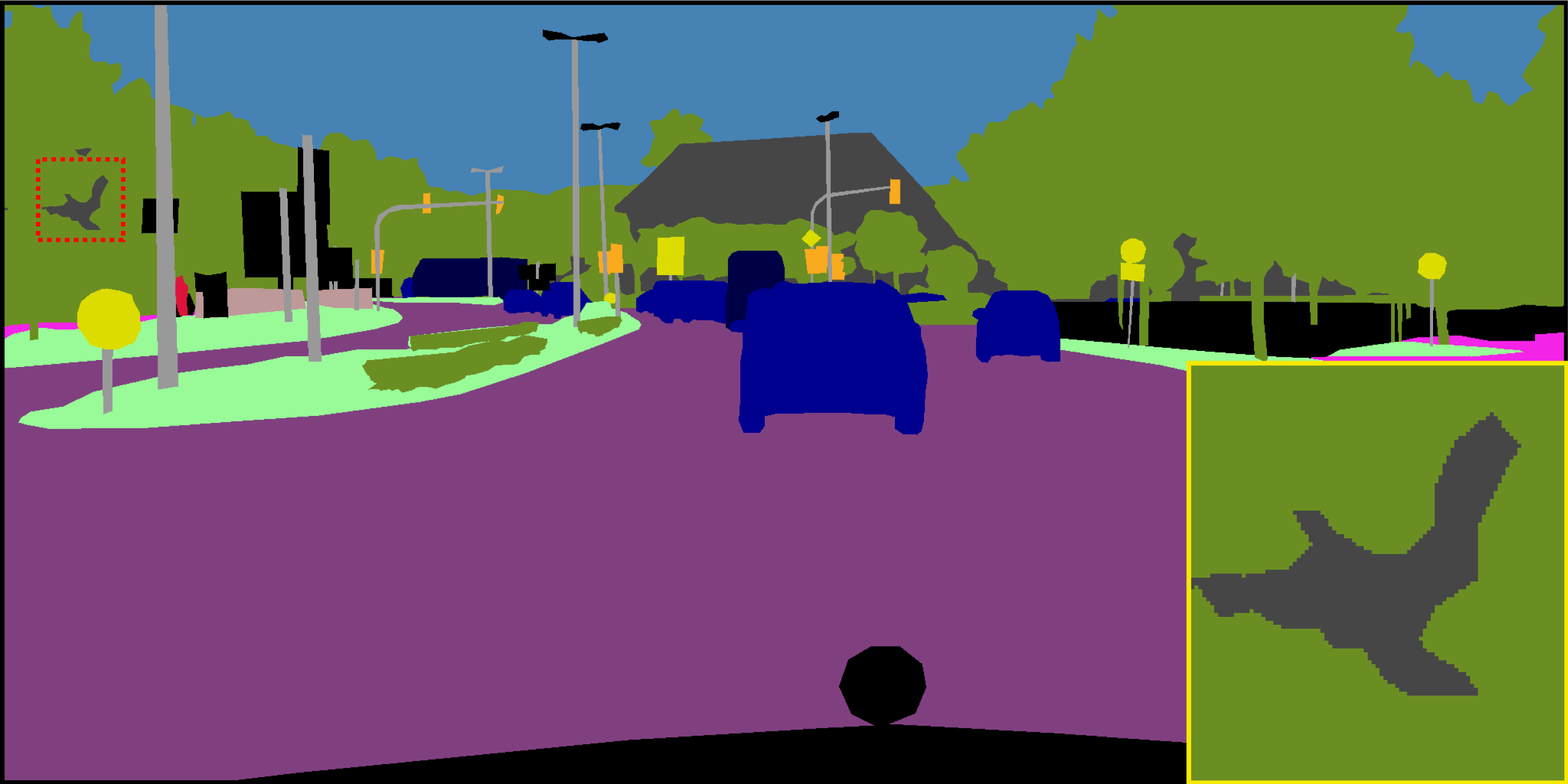}
	\end{subfigure}
	\begin{subfigure}[b]{0.19\textwidth}
		\centering
		\includegraphics[width=\textwidth]{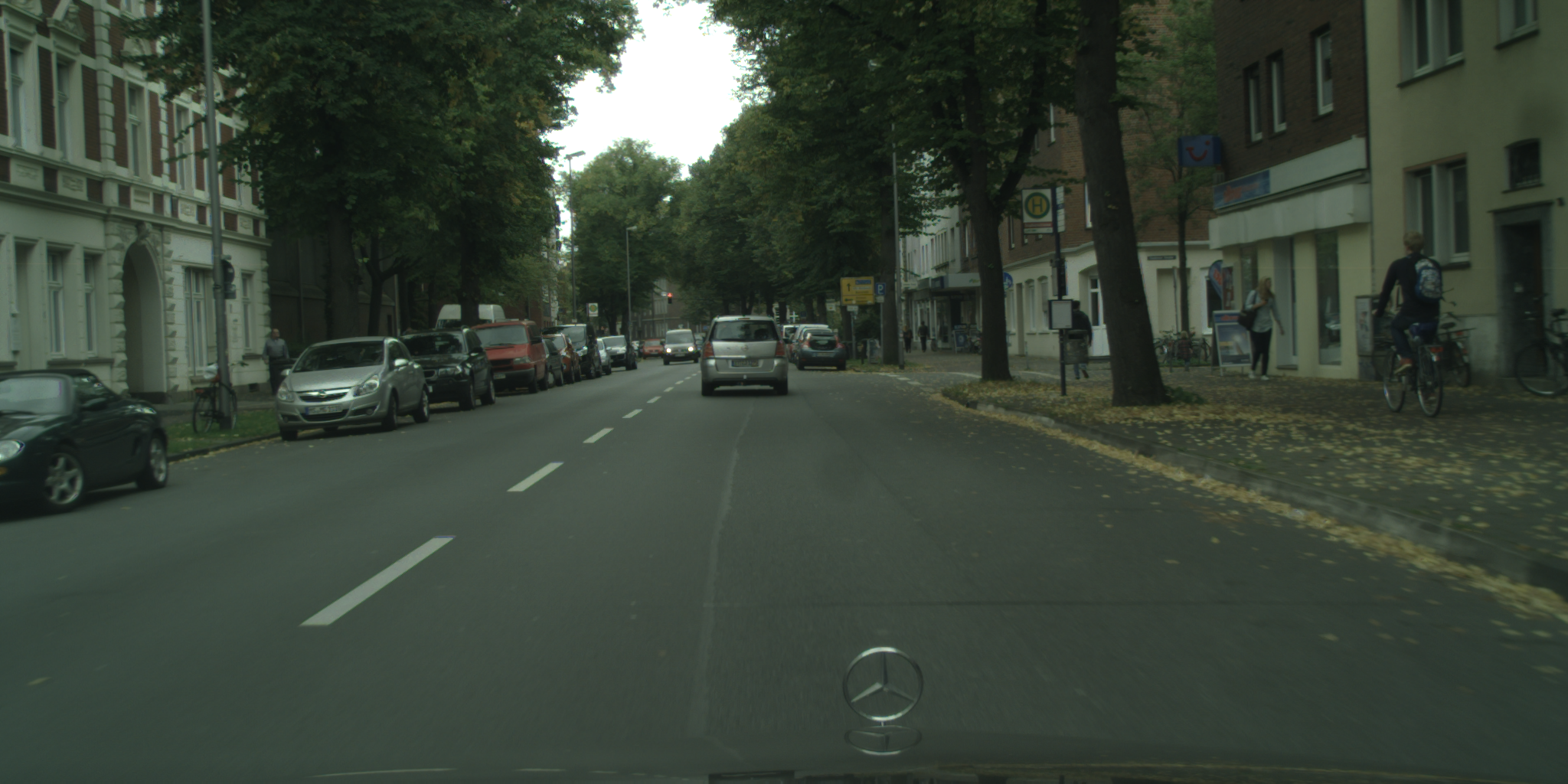}
	\end{subfigure}
	\begin{subfigure}[b]{0.19\textwidth}
		\centering
		\includegraphics[width=\textwidth]{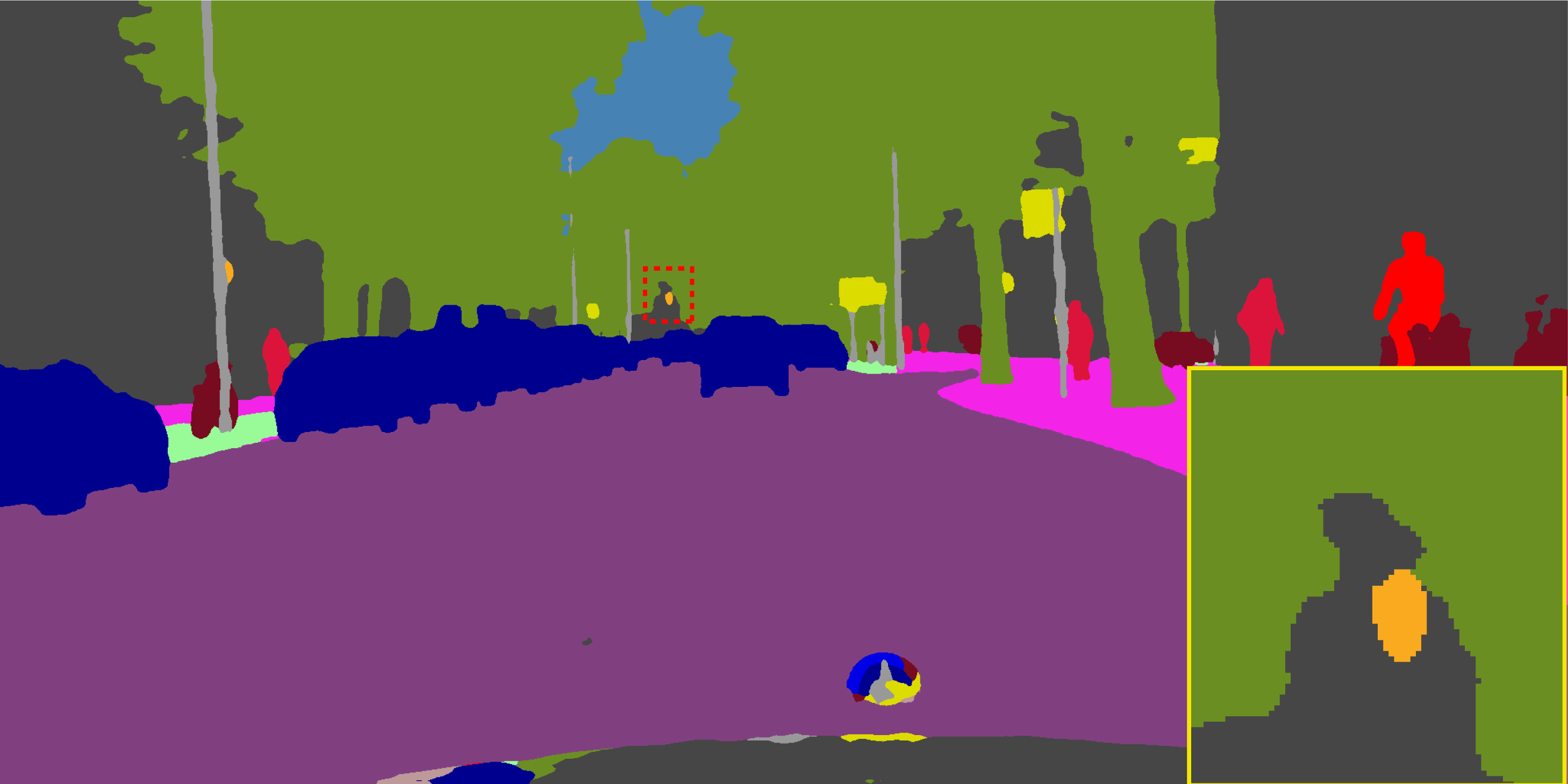}
	\end{subfigure}
	\begin{subfigure}[b]{0.19\textwidth}
		\centering
		\includegraphics[width=\textwidth]{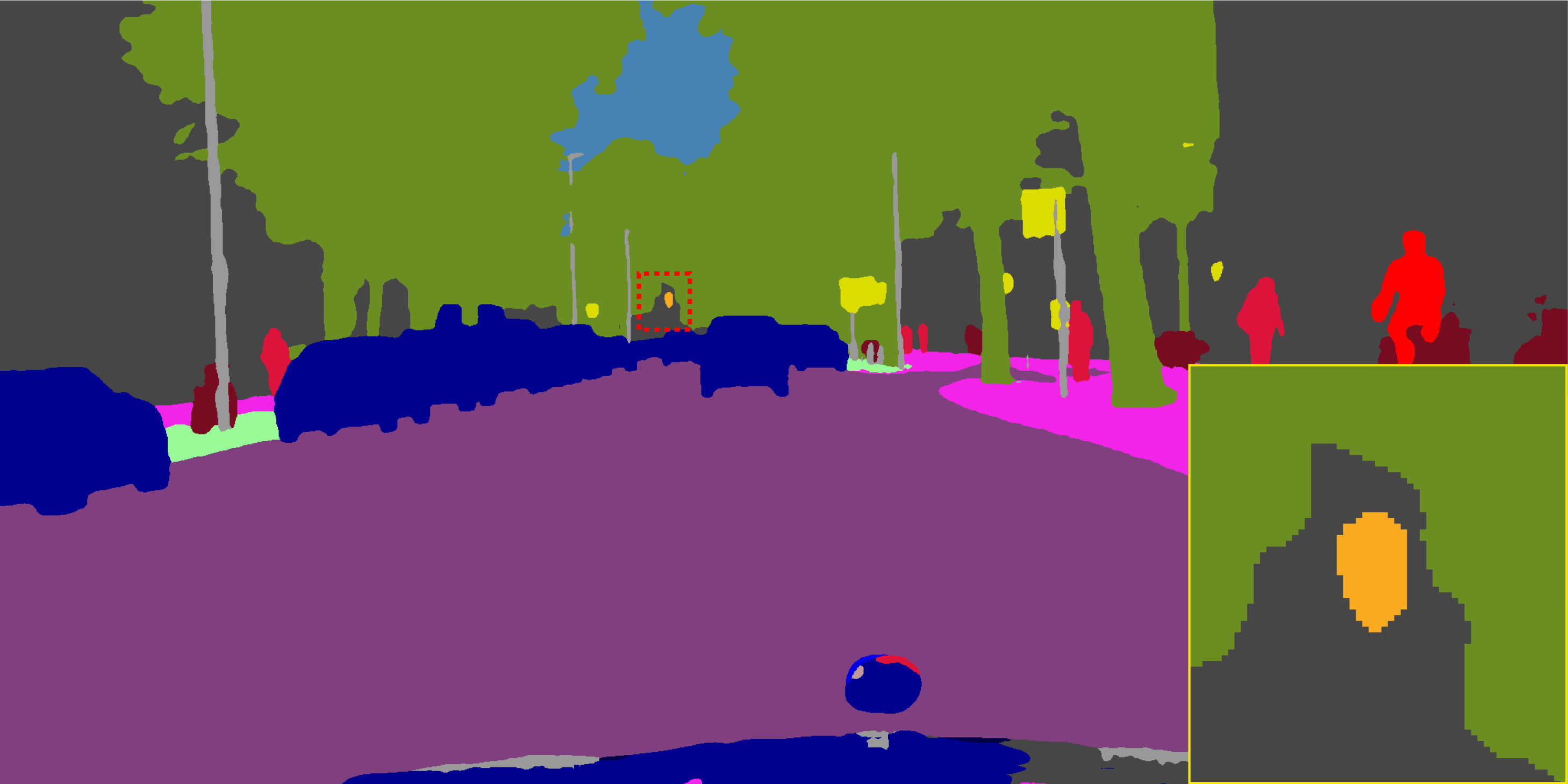}
	\end{subfigure}
	\begin{subfigure}[b]{0.19\textwidth}
		\centering
		\includegraphics[width=\textwidth]{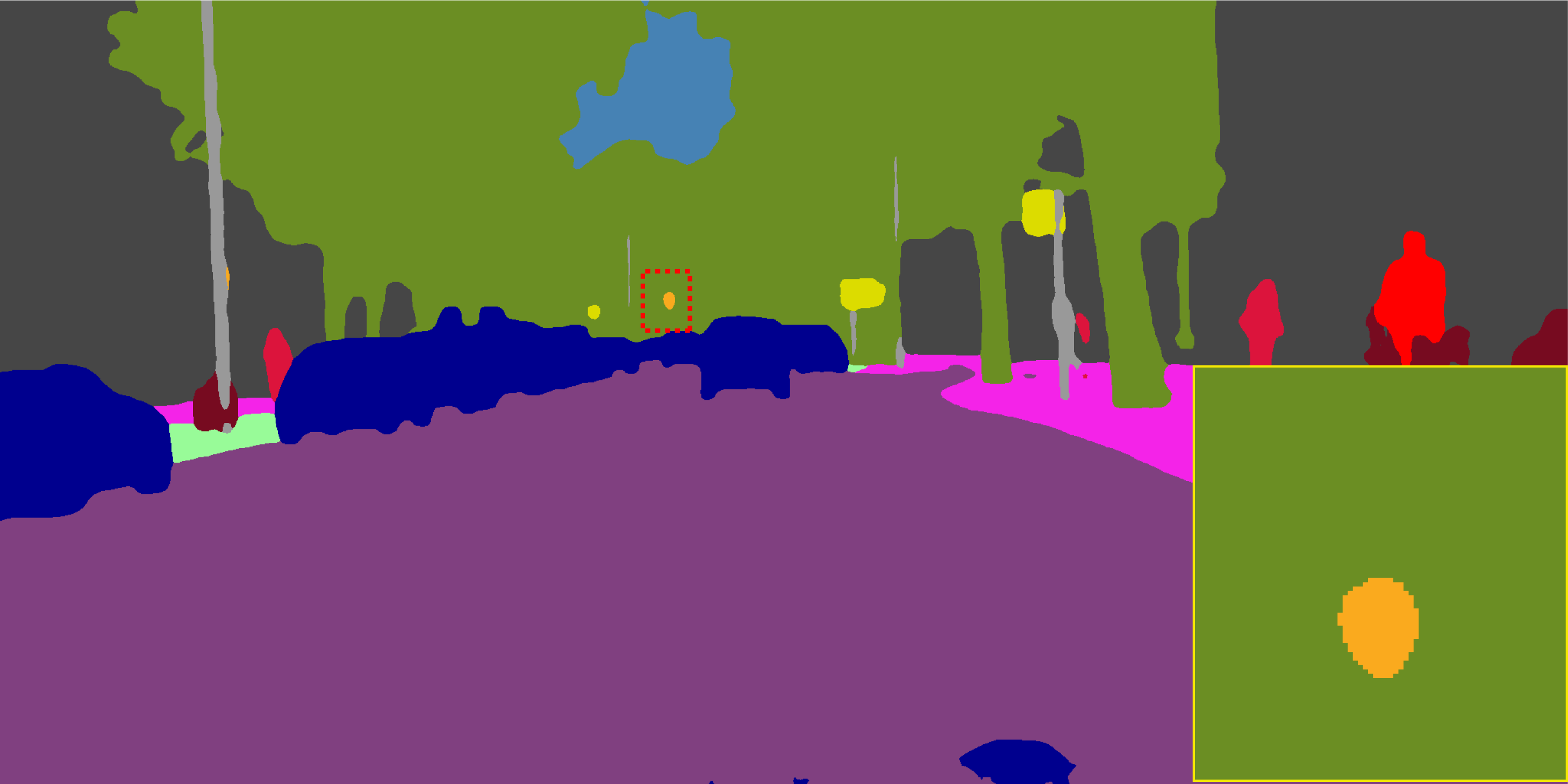}
	\end{subfigure}
	\centering
	\begin{subfigure}[b]{0.19\textwidth}
		\centering
		\includegraphics[width=\textwidth]{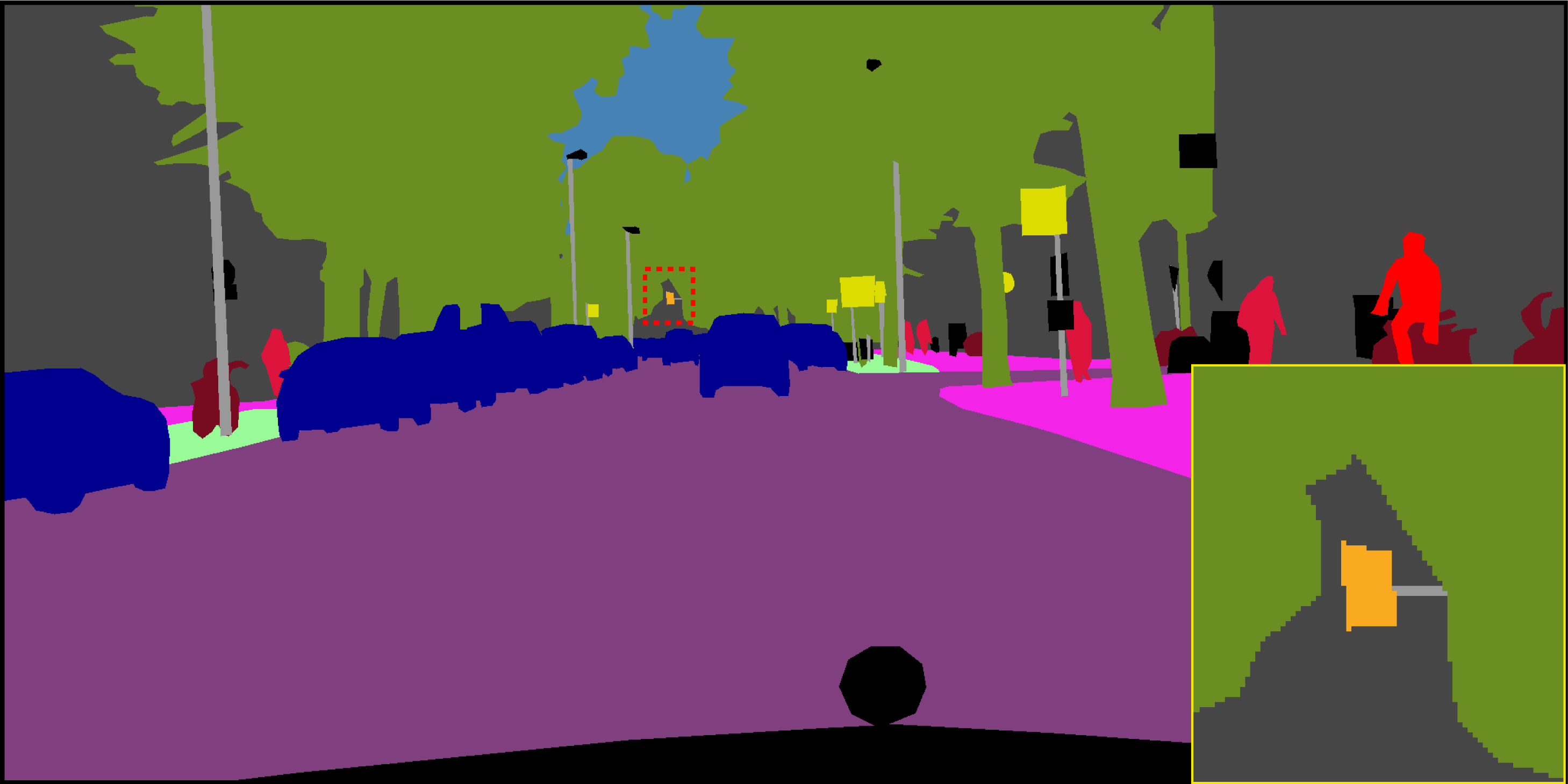}
	\end{subfigure}
	\centering
	\begin{subfigure}[b]{0.19\textwidth}
		\centering
		\includegraphics[width=\textwidth]{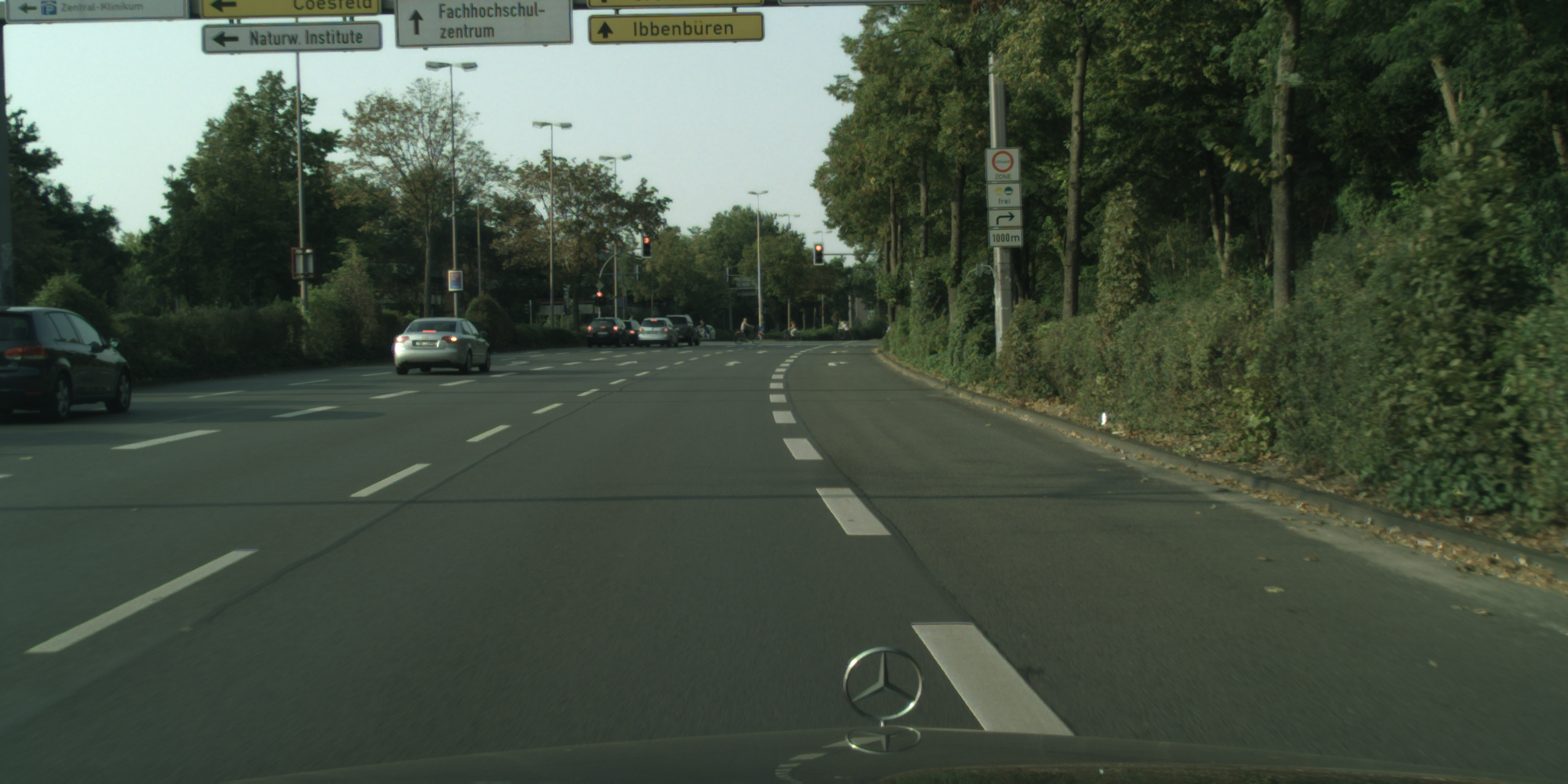}
		\caption{Original}
	\end{subfigure}
	\begin{subfigure}[b]{0.19\textwidth}
		\centering
		\includegraphics[width=\textwidth]{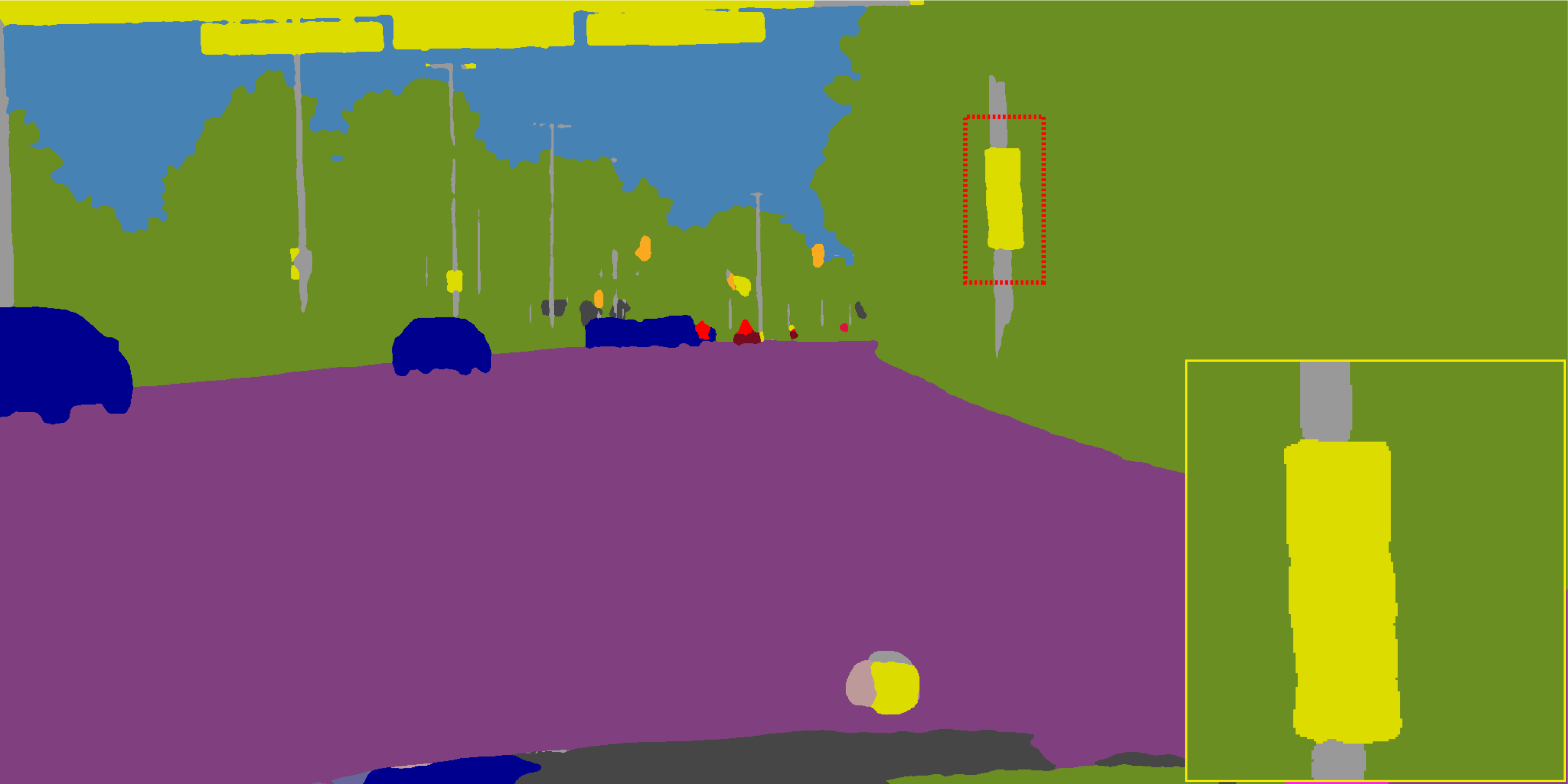}
		\caption{S\textsuperscript{2}-FPN18}
	\end{subfigure}
	\begin{subfigure}[b]{0.19\textwidth}
		\centering
		\includegraphics[width=\textwidth]{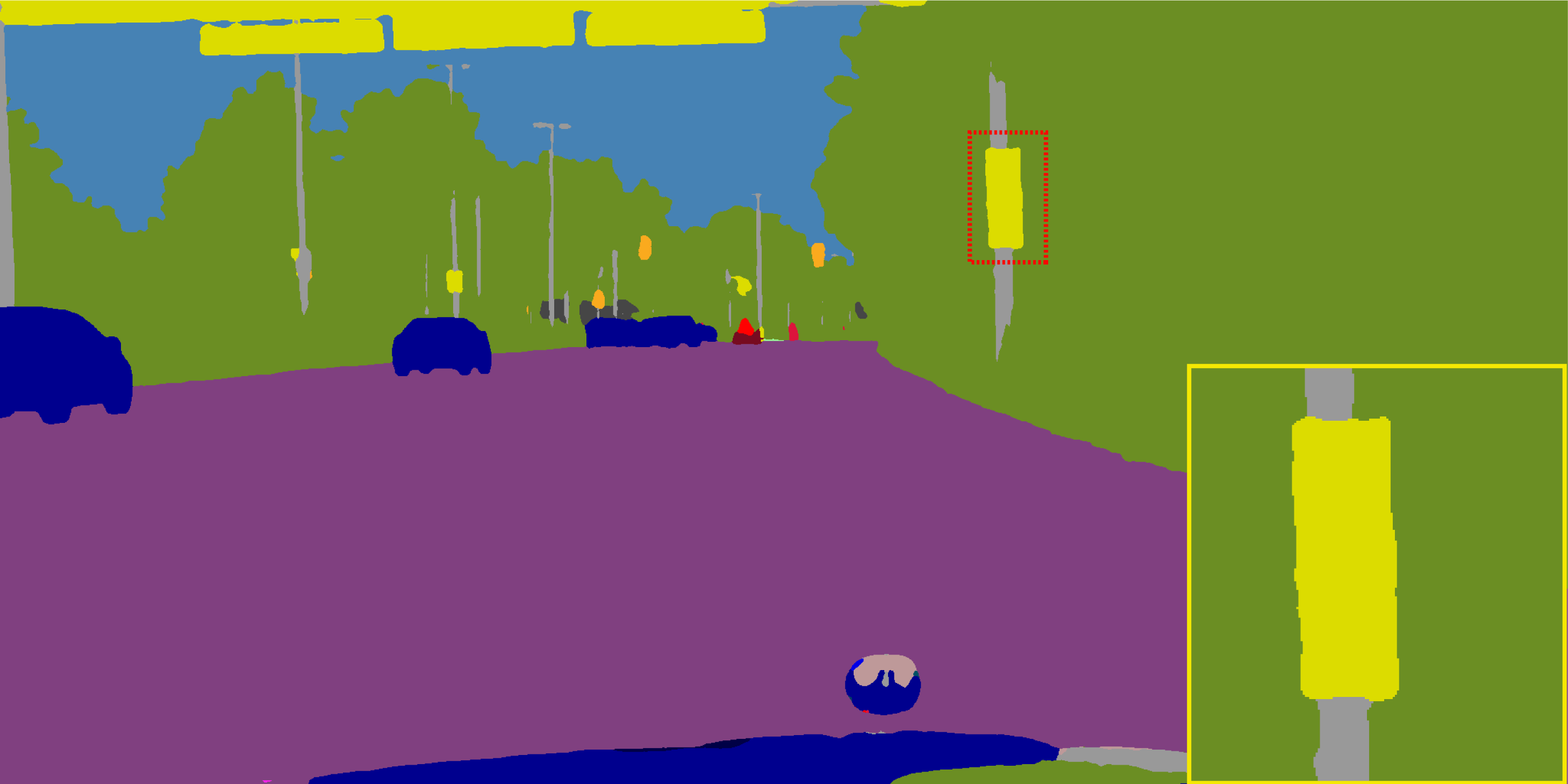}
		\caption{S\textsuperscript{2}-FPN34}
	\end{subfigure}
	\begin{subfigure}[b]{0.19\textwidth}
		\centering
		\includegraphics[width=\textwidth]{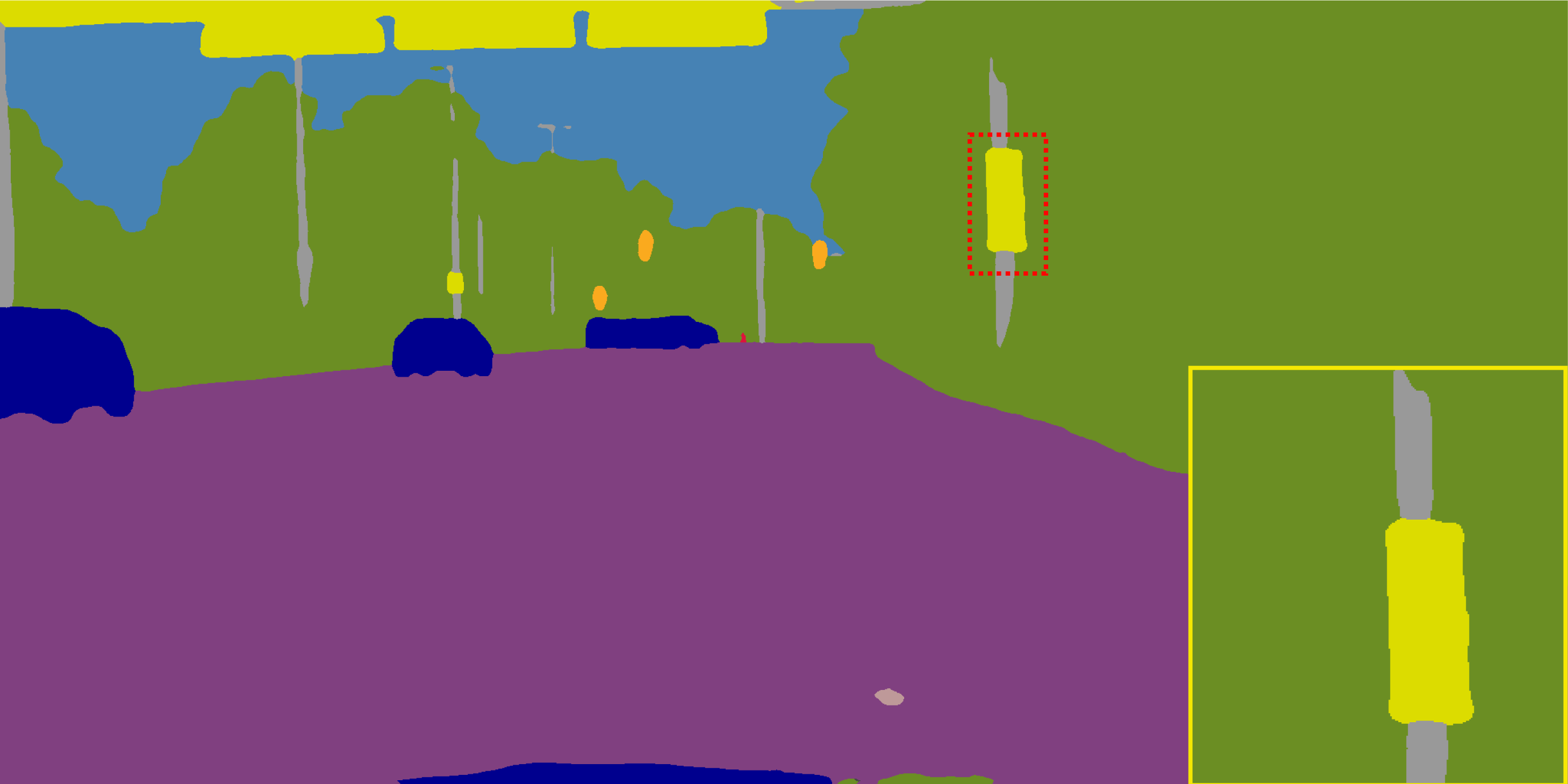}
		\caption{S\textsuperscript{2}-FPN34M}
	\end{subfigure}
	\centering
	\begin{subfigure}[b]{0.19\textwidth}
		\centering
		\includegraphics[width=\textwidth]{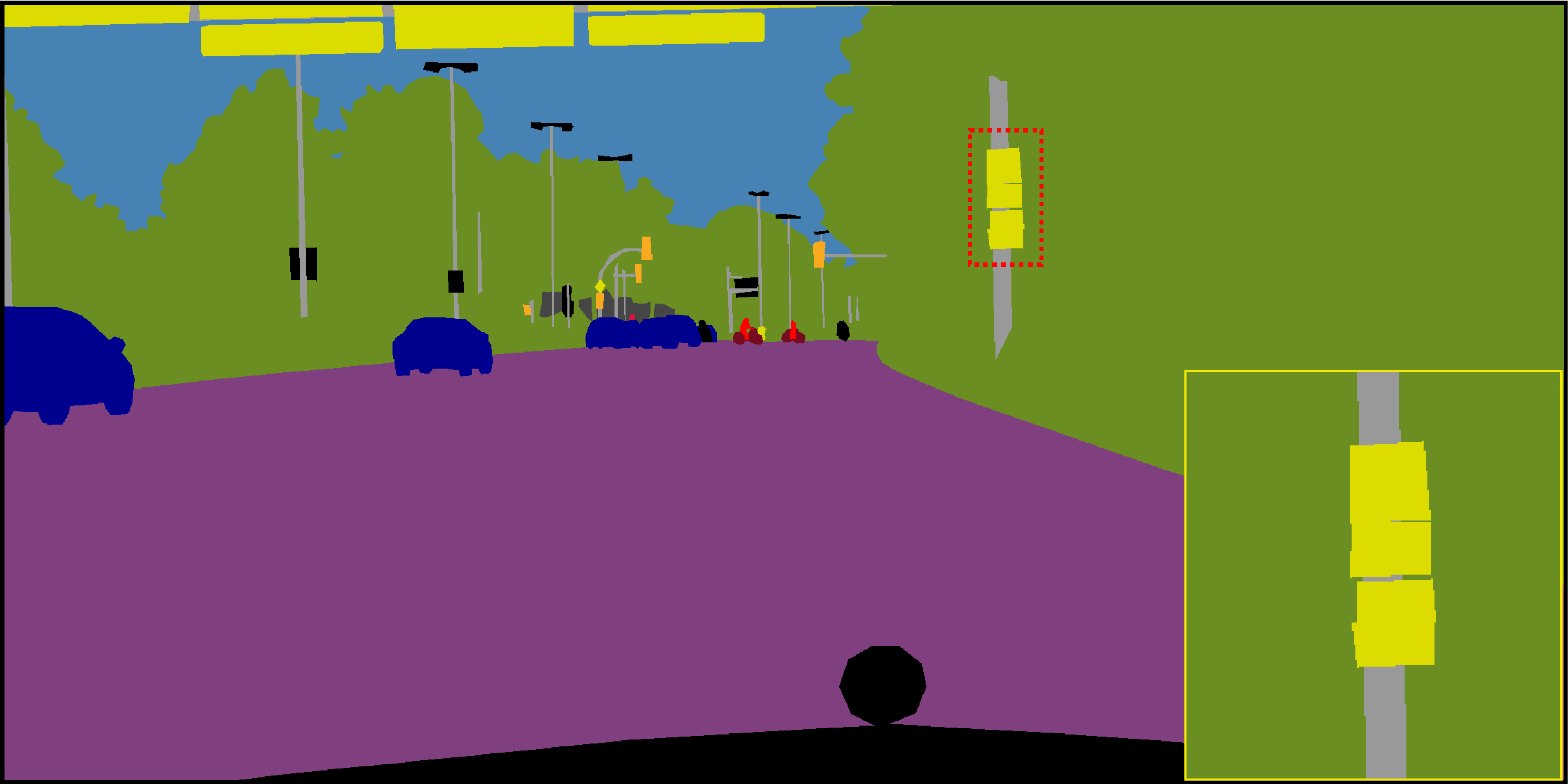}
		\caption{Ground Truth}
	\end{subfigure}
	\caption{Visual results of our method S\textsuperscript{2}-FPN on Cityscapes dataset.}
	\label{cityscapes_fig}
\end{figure*}
\begin{table}
	\caption{\MakeUppercase{Comparison Corssentropy and OHEM loss on Cityscapes Validation dataset}}
	\label{compare_of_ohem}
	\centering
	\begin{adjustbox}{width=0.9\linewidth}
		\small
		\begin{tabular}{c|c|c|c|c}
			\hline
			Method&Params&Flops&FPS&mIoU\\ 
			\hline\hline
			CE&17.5&29.1&87&75.6\\
			OHEM&17.5&29.1&87&76.4\\
			\hline
		\end{tabular}
	\end{adjustbox}
\end{table}

\textit{Online Bootstrapping:} Camvid and Cityscapes datasets contain easy and hard pixels. One efficient way to train these hard pixels is to automate their selection process during training. Following the previous works[], we utilize OHEM or online hard example mining for hard training pixels. The examples with a probability less than the threshold are considered hard training pixels. In our experiments, k depends on the image crop size and the threshold is set to 0.7. Table \ref{compare_of_ohem} shows that using OHEM increases the performance by 0.9 \% mIoU over Cross-entropy on Cityscapes validation dataset. 

\textbf{Ablation study with different baselines:} We further detailed the performance comparison of three settings for our proposed S\textsuperscript{2}-FPN model on the Cityscapes validation dataset. Specifically, we train the model using the following baseline, ResNet18, ResNet34 and a modified version of ResNet34 as shown in Table \ref{compare_sspnet_settings}. S\textsuperscript{2}-FPN obtains accuracy of 76.4\% mIoU and 87.3 FPS, which has a better inference speed than using ResNet34 variants with minor inaccuracy. Using the modified ResNet34 (S\textsuperscript{2}-FPN34M) achieve the highest accuracy with 0.4 \% mIoU increase over S\textsuperscript{2}-FPN34 with the same number of parameters, but with very high floating points. S\textsuperscript{2}-FPN18 and S\textsuperscript{2}-FPN34 seem to obtain a better accuracy/speed trade-offs.

\begin{table}
	\caption{\MakeUppercase{Ablation study for the proposed modules on the	Cityscapes validation set,where ResNet18 architecture serves as the baseline. We show the effectiveness of GFM: Global Fusion Module, AFP: Attention Pyramid Fusion , SSAM: Scale-aware Strip Attention, and supervision}}
	\label{cityscape_ablation_result}
	\begin{center}
		\begin{adjustbox}{width=\linewidth}
			\small
			\begin{tabu}{l|l|l|l|l|l|l|l|l}
				\hline
	            Backbone&GFU&APF&Supervision&SSAM&Params&Flops&FPS&mIoU(\%)\\\hline \hline
				\checkmark&&&&&11.2&19&187&65.7\\
				\checkmark& \checkmark&&&&11.5&19.7&156.9&71.9\\
				\checkmark& \checkmark & \checkmark&&&17.8&29.2&93.8&75.6\\
				\checkmark& \checkmark& \checkmark& \checkmark&&17.8&29.2&88&75.9\\
				\checkmark& \checkmark& \checkmark&\checkmark& \checkmark&17.8&29.1&87&76.4\\
				\hline
			\end{tabu}
		\end{adjustbox}
	\end{center}
\end{table}

\textbf{Effectiveness of Attention Pyramid Fusion (APF) module:}. To verify the effectiveness of the attention pyramid fusion module, we compare it with the conventional feature pyramid network. As shown in Table \ref{compare_feature_pyramid_network}, the attention feature pyramid module achieves 73.9\% mIoU, which is 4.8 \% higher than the feature pyramid network. The result shows that channel reweighing and contextual information encoding are effectively incorporated in the proposed attention pyramid fusion.

\begin{table}
	\caption{\MakeUppercase{Comparison between Feature Pyramid Network (FPN) and Attention Pyramid Fusion (APF) on the Cityscapes Validation dataset}}
	\label{compare_feature_pyramid_network}
	\centering
	\begin{adjustbox}{width=0.85\linewidth}
		\small
		\begin{tabular}{c|c|c|c|c}
			\hline
			Method&Params&Flops&FPS&mIoU\\ 
			\hline\hline
			baseline&11.2&19&187&65.7\\
			FPN&11.4&20.9&152&69.1\\
			AFPN&17.5&28.6&94.3&73.9\\
			\hline
		\end{tabular}
	\end{adjustbox}
\end{table}

\subsection{Comparisons with State-of-the-Art Methods}
In Table \ref{miou_cityscapes}, we present the comparisons between our S\textsuperscript{2}-FPN and the state-of-the-arts real-time semantic segmentation methods. Our method is tested on a single GTX  1080Ti GPU with image resolution of 512 × 1024. We tested the speed without any accelerating strategy and used only fine-data to train the model. As stated in Table \ref{miou_cityscapes}, our proposed network setting get 76.2\% mIoU with 102 FPS by using ResNet18 as the backbone network. As can be observed, the accuracy is better than all other methods except for STD-Seg75, which has 0.2 \% mIoU better than our small model. The inference speed is significantly faster, even though we tested it using GTX1080 ti, this accuracy and inference speed prove that even with light backbones our approach still obtains better performance than other approaches. Besides, S\textsuperscript{2}-FPN with ResNet34 base model obtains 85 FPS inference speed with 77.4 \% mIoU accuracy,
which is state of the art on speed/accuracy trade-offs on the Cityscapes benchmark and exceed the best model STD-Seg75 by a margin of 0.6 \% mIoU. Finally, we have modified the ResNet34 backbone by changing the stride in stage 2 from 2 to 1 and run the model (we name it S\textsuperscript{2}-FPN34M), this modification increases the accuracy by about 0.4 \% mIoU but reduced the speed significantly to 30.5 FPS. We give a detail of per-class accuracy of our three configurations Table \ref{per_class_cityscapes}.

\begin{figure*}
	\centering
	\begin{subfigure}[b]{0.19\textwidth}
		\includegraphics[width=\textwidth]{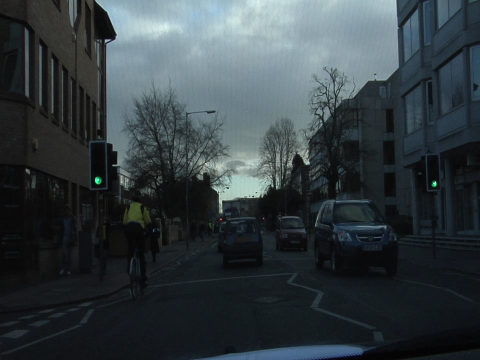}
	\end{subfigure}
	\begin{subfigure}[b]{0.19\textwidth}
	\includegraphics[width=\textwidth]{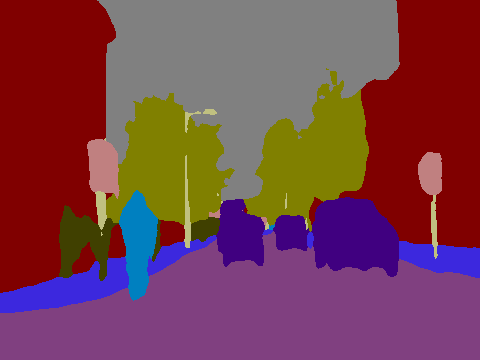}
	\end{subfigure}
	\begin{subfigure}[b]{0.19\textwidth}
		\includegraphics[width=\textwidth]{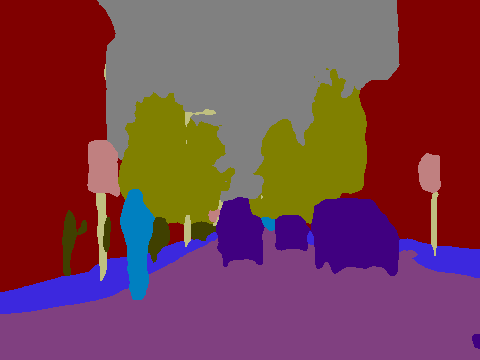}
	\end{subfigure}
	\begin{subfigure}[b]{0.19\textwidth}
		\includegraphics[width=\textwidth]{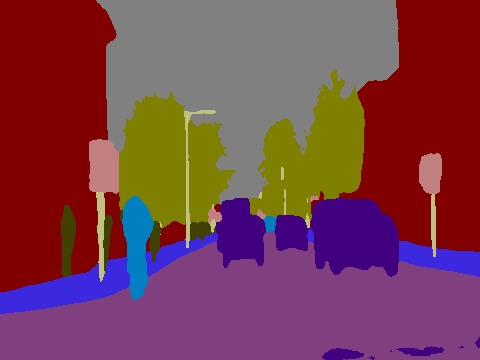}
	\end{subfigure}
	\begin{subfigure}[b]{0.19\textwidth}
	\includegraphics[width=\textwidth]{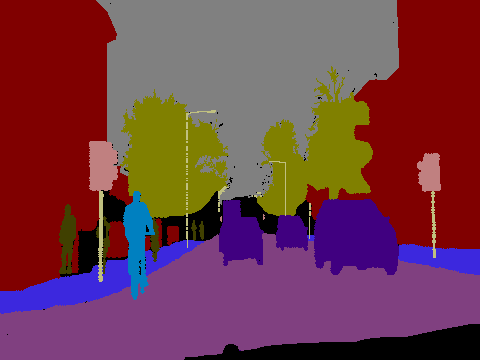}
    \end{subfigure}
	\begin{subfigure}[b]{0.19\textwidth}
		\centering
		\includegraphics[width=\textwidth]{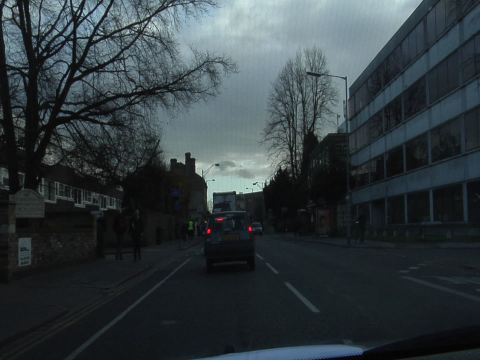}
	\end{subfigure}
	\begin{subfigure}[b]{0.19\textwidth}
	\centering
	\includegraphics[width=\textwidth]{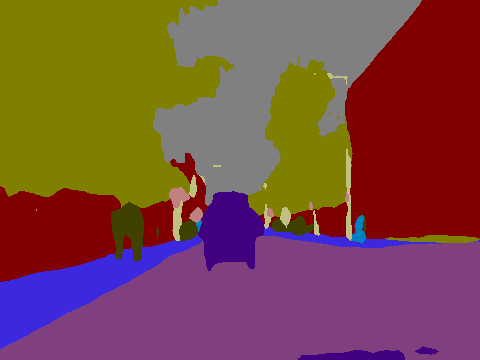}
    \end{subfigure}
	\begin{subfigure}[b]{0.19\textwidth}
		\centering
		\includegraphics[width=\textwidth]{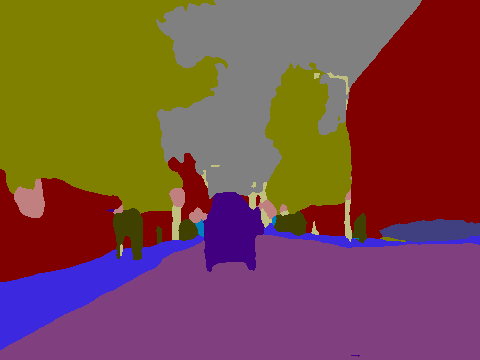}
	\end{subfigure}
	\begin{subfigure}[b]{0.19\textwidth}
		\centering
		\includegraphics[width=\textwidth]{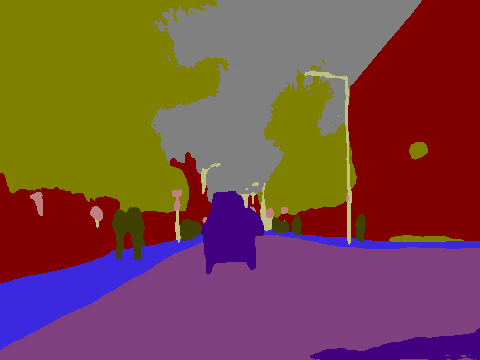}
	\end{subfigure}
	\begin{subfigure}[b]{0.19\textwidth}
	\centering
	\includegraphics[width=\textwidth]{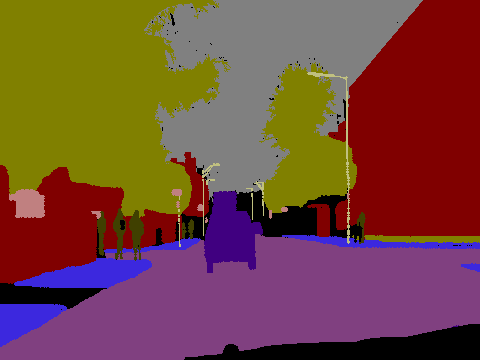}
	\end{subfigure}
	\centering
	\begin{subfigure}[b]{0.19\textwidth}
		\centering
		\includegraphics[width=\textwidth]{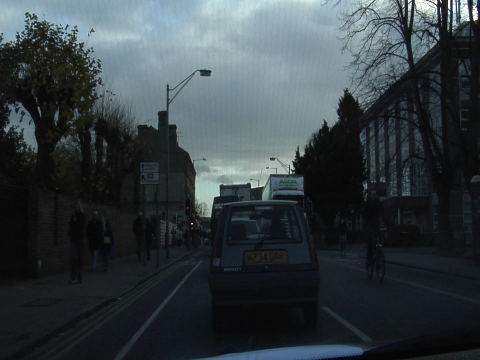}
	\end{subfigure}
\begin{subfigure}[b]{0.19\textwidth}
	\centering
	\includegraphics[width=\textwidth]{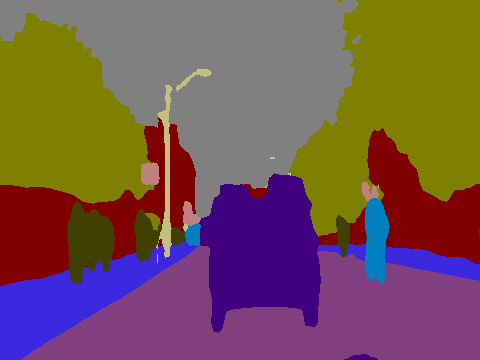}
\end{subfigure}
	\begin{subfigure}[b]{0.19\textwidth}
		\centering
		\includegraphics[width=\textwidth]{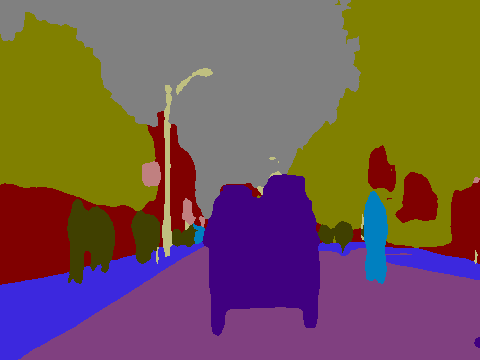}
	\end{subfigure}
	\begin{subfigure}[b]{0.19\textwidth}
		\centering
		\includegraphics[width=\textwidth]{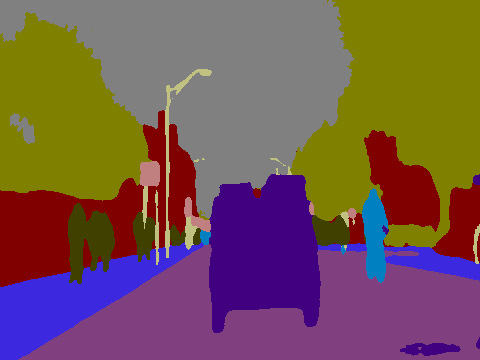}
	\end{subfigure}
	\begin{subfigure}[b]{0.19\textwidth}
	\centering
	\includegraphics[width=\textwidth]{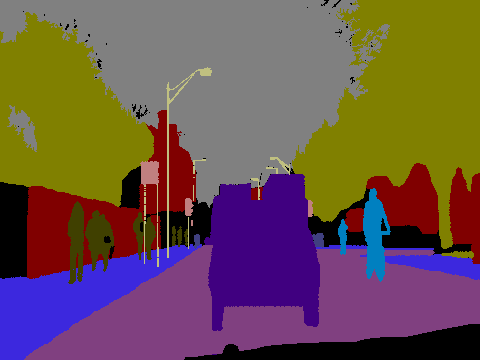}
     \end{subfigure}
	\centering
	\begin{subfigure}[b]{0.19\textwidth}
		\centering
		\includegraphics[width=\textwidth]{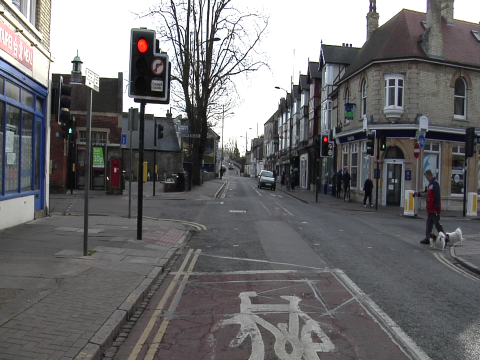}
	\end{subfigure}
	\begin{subfigure}[b]{0.19\textwidth}
	\centering
	\includegraphics[width=\textwidth]{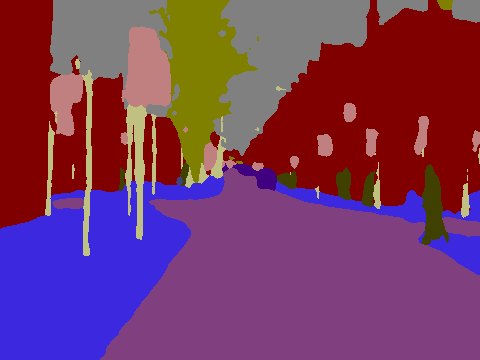}
\end{subfigure}
	\begin{subfigure}[b]{0.19\textwidth}
		\centering
		\includegraphics[width=\textwidth]{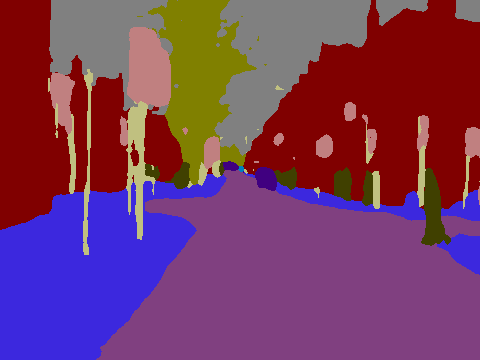}
	\end{subfigure}
	\begin{subfigure}[b]{0.19\textwidth}
		\centering
		\includegraphics[width=\textwidth]{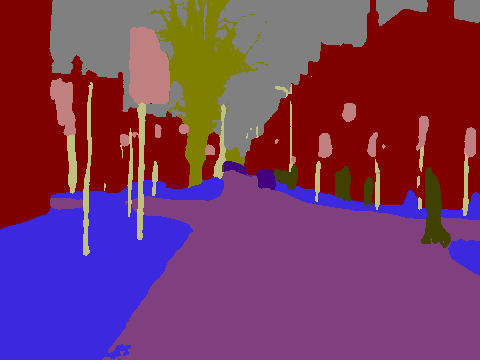}
	\end{subfigure}
	\begin{subfigure}[b]{0.19\textwidth}
	\centering
	\includegraphics[width=\textwidth]{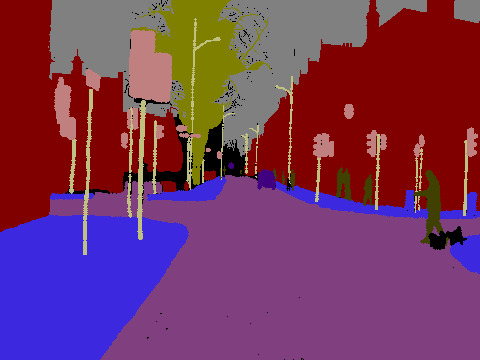}
	\end{subfigure}
	\begin{subfigure}[b]{0.19\textwidth}
		\centering
		\includegraphics[width=\textwidth]{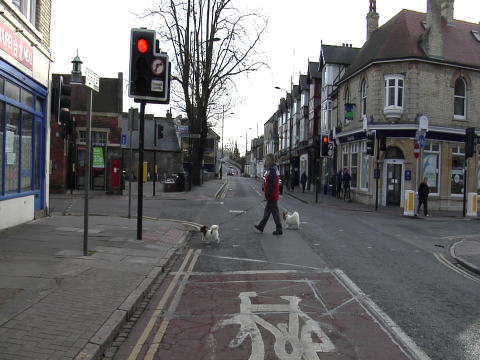}
	\end{subfigure}
	\begin{subfigure}[b]{0.19\textwidth}
	\centering
	\includegraphics[width=\textwidth]{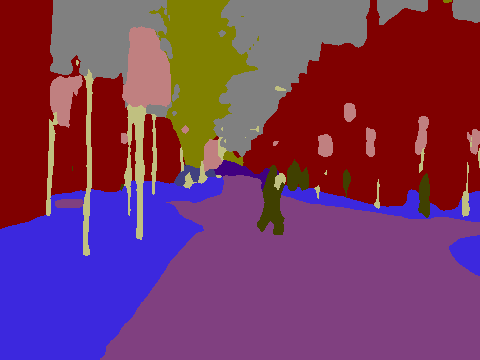}
\end{subfigure}
	\begin{subfigure}[b]{0.19\textwidth}
		\centering
		\includegraphics[width=\textwidth]{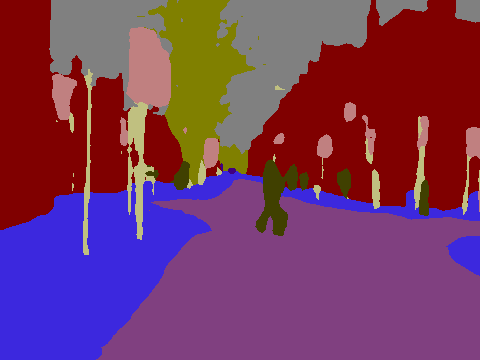}
	\end{subfigure}
	\begin{subfigure}[b]{0.19\textwidth}
		\centering
		\includegraphics[width=\textwidth]{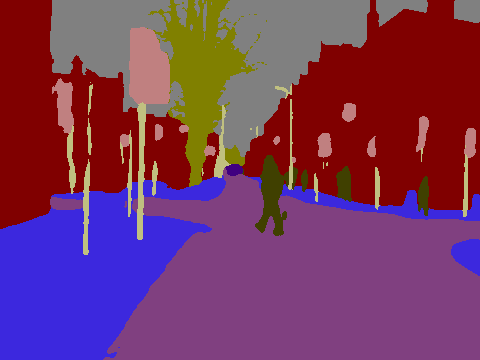}
	\end{subfigure}
	\centering
	\begin{subfigure}[b]{0.19\textwidth}
		\centering
		\includegraphics[width=\textwidth]{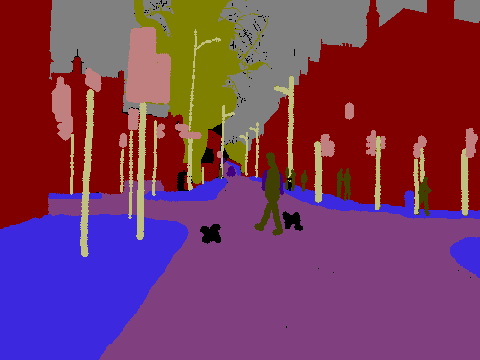}
	\end{subfigure}
	\centering
	\begin{subfigure}[b]{0.19\textwidth}
		\centering
		\includegraphics[width=\textwidth]{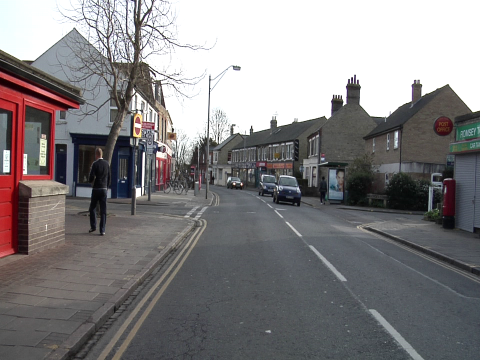}
		\caption{Original}
	\end{subfigure}
	\begin{subfigure}[b]{0.19\textwidth}
	\centering
	\includegraphics[width=\textwidth]{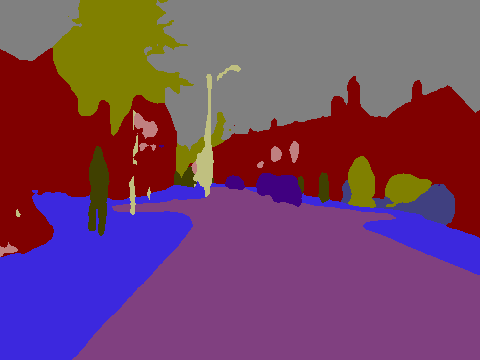}
	\caption{S\textsuperscript{2}-FPN18}
\end{subfigure}
	\begin{subfigure}[b]{0.19\textwidth}
		\centering
		\includegraphics[width=\textwidth]{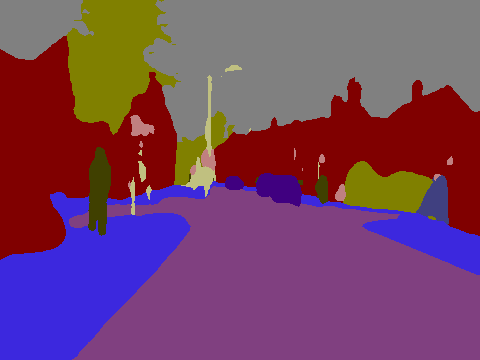}
		\caption{S\textsuperscript{2}-FPN34}
	\end{subfigure}
	\begin{subfigure}[b]{0.19\textwidth}
		\centering
		\includegraphics[width=\textwidth]{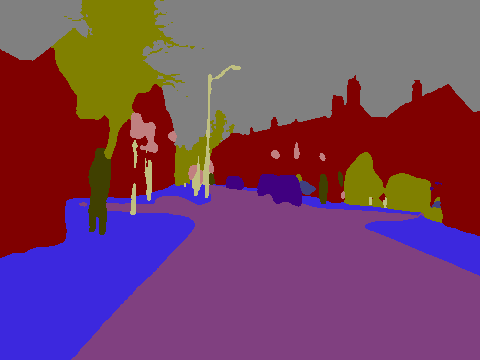}
		\caption{S\textsuperscript{2}-FPN34M}
	\end{subfigure}
	\centering
	\begin{subfigure}[b]{0.19\textwidth}
		\centering
		\includegraphics[width=\textwidth]{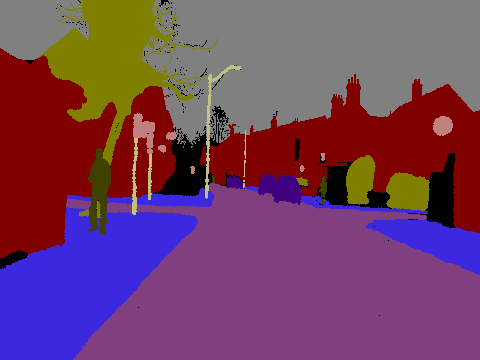}
		\caption{Ground Truth}
	\end{subfigure}
	\caption{Visual results of our method S\textsuperscript{2}-FPN on Camvid dataset.}
	\label{camvid_fig}
\end{figure*}
\subsection{Results on Camvid Dataset}
\label{subsec:camvid_result}
As illustrated in Table \ref{camvid_s}, we evaluate the segmentation accuracy of our proposed method on the Camvid dataset. We trained with an input image size of 360x480, with no use of external data. The training images used to train the model and validation images to evaluate our models while the testing samples are used to get the result for comparison with other methods. Overall, our model has better accuracy. Our S\textsuperscript{2}-FPN model with ResNet18 as a backbone achieves a 69.6 \% mIoU on the test set, while S\textsuperscript{2}-FPN34 obtains accuracy of 71.0 \% mIoU, which is a significant trade-off between speed and accuracy. The last model configuration S\textsuperscript{2}-FPN34M which replaces the stride at stage 2 from 2 to 1 achieves 74.2\% mIoU . This further shows the validity and effectiveness of our proposed method. 
\begin{table*}
	\centering
	\caption{\MakeUppercase{Individual category results on Camvid test set in terms of mIoU for 11 classes}}
	\label{camvid_s}
	\vspace{1ex}
	\begin{center}
		\begin{adjustbox}{width=0.95\textwidth}
		\small
	\begin{tabu}{l|l|l|l|l|l|l|l|l|l|l|l|l}
		\hline
		\bf{Method}&\bf{Bui}&\bf{Tree}&\bf{Sky}&\bf{Car}&\bf{Sig}&\bf{Roa}&\bf{Ped}&\bf{Fen}&\bf{Pol}&\bf{Side}&\bf{Bic}&\bf{mIoU}\\
		\hline
		\hline
		SegNet\cite{badrinarayanan2017segnet}&88.8&87.3&92.4&82.1&20.5&97.2&57.1&49.3&27.5&84.4&30.7&55.6\\
		ENet\cite{paszke2016enet}&74.7&77.8&95.1&82.4&51.0&95.1&67.2&51.7&35.4&86.7&34.1&51.3\\
		BiSeNet1\cite{yu2018bisenet}&82.2&74.4&91.9&80.8&42.8&93.3&53.8&49.7&25.4&77.3&50.0&65.6\\
		BiSeNet2\cite{yu2018bisenet}&83.0&75.8&92.0&83.7&46.5&94.6&58.8&	53.6&31.9&81.4&54.0&68.7\\
		NDNet45-FCN8-LF\cite{yang2020small}&85.5&84.6&94.8&82.6&39.2&97.4&60.1&37.3&17.6&86.8&53.7&57.5\\
		LBN-AA\cite{dong2020real}&83.2&70.5&92.5&81.7&51.6&93.0&55.6&53.2&36.3&82.1&47.9&68.0\\
		AGLNet\cite{zhou2020aglnet}&82.6&76.1&91.0&87.0&45.3&95.4&61.5&39.5&39.0&83.1&62.7&69.4\\
		BiSeNetV2\slash BiSeNetV2L\cite{yu2021bisenet}&-&-&-&-&-&-&-&-&-&-&-&72.4\slash 73.2\\
		\hline
		S\textsuperscript{2}-FPN18&83.&77.2&91.8&88.9&48.2&95.7&56.4&43.4&32.4&84.8&62.5&69.6\\
		S\textsuperscript{2}-FPN34&85.3&77.4&91.7&91.2&49.6&95.7&59.1&46.8&33.2&85.4&66.5&71.0\\
		S\textsuperscript{2}-FPN34M&86.0&78.8&92.6&92.2&56.2&96.0&67.1&47.3&42.1&86.8&70.7&74.2\\
		\hline
	\end{tabu}}
\end{adjustbox}
	\end{center}
\end{table*}  
\begin{table}
	\caption{\MakeUppercase{Results of the Model on Camvid Dataset.$*$ indicates the models pre-trained on Cityscapes}}
	\label{camvid_wit_sota}
	\begin{center}
		\begin{adjustbox}{width=0.48\textwidth}
		\small
		\begin{tabu}{l|c|c|c|c}
			\hline
			\bf{Method}&\bf{Year}&\bf{Params M}&\bf{Speed (FPS)}&\bf{mIoU\%}\\
			\hline\hline
			Deeplab\cite{chen2017deeplab}&2017&262.1&4.9&61.6\\
			PSPNet \cite{zhao2017pyramid}&2017& 250.8&5.4&69.1\\
			\hline
			SegNet \cite{badrinarayanan2017segnet}&2015&29.5&46&55.6\\
			ENet \cite{paszke2016enet}&2016&0.36&-&61.3\\
			DFANet-A \cite{li2019dfanet}&2019&7.8&120&64.7\\
			DFANet-B \cite{li2019dfanet}&2019&4.8&160&59.3\\
			BiSeNet1 \cite{yu2018bisenet}&2018&5.8&175&65.7\\
			BiSeNet2 \cite{yu2018bisenet}&2018&49.0&116.3&68.7\\
			ICNet \cite{zhao2018icnet}&2018&26.5&27.8&67.1\\
			DABNet \cite{li2019dabnet}&2019&0.76&&66.4\\
			AGLNet \cite{zhou2020aglnet}&2020&1.12&90.1&69.4\\
			NDNet45-FCN8-LF \cite{yang2020small}&2020&1.1&-&57.5\\
			LBN-AA \cite{dong2020real}&2020&6.2&39.3&68.0\\
			BiSeNetV2/BiSeNetV2L \cite{yu2021bisenet}&2021&-&-&72.4/73.2\\
			STDC1-Seg75 \cite{fan2021rethinking}&2021&8.4&197.6&73.0\\
			STDC2-Seg75 \cite{fan2021rethinking}&2021&12.5&152.2&73.9\\
			\hline 
			S\textsuperscript{2}-FPN18 &&17.8&124.2&69.5\\
			S\textsuperscript{2}-FPN34&&27.9 &107.2&71.0\\
			S\textsuperscript{2}-FPN34M&&27.9&55.5&74.2\\\hline
		\end{tabu}
	\end{adjustbox}
	\end{center}
\end{table}
\section{Conclusion}\label{sec13}
In this work, we designed architecture for semantic segmentation that achieves a better trade-off accuracy/speed and presents Scale-aware Strip Attention S\textsuperscript{2}-FPN for scene parsing in real-time. First, we introduce Scale-aware Strip Attention and channel attention modules to model the long-range dependencies. Specifically, by using scale-aware and vertical strip operations, our network drastically reduced the computation cost of the attention mechanism. Furthermore, we propose the Attention Pyramid Fusion module, which enables and facilitates acquiring contextual information by utilizing the attention mechanism. The ablation studies show the effectiveness of scale-ware strip attention and attention pyramid fusion modules. Experimental results demonstrate that S\textsuperscript{2}-FPN achieves accuracy/speed trade-offs on Camvid and Cityscapes.

\bibliographystyle{IEEEtran}
\bibliography{refs}
\end{document}